\PassOptionsToPackage{table}{xcolor}
\documentclass[runningheads]{llncs}

\usepackage{eccv}
\usepackage{eccvabbrv}

\usepackage{graphicx}
\usepackage{booktabs}

\usepackage[accsupp]{axessibility}  % Improves PDF readability for those with disabilities.

\usepackage[pagebackref,breaklinks,colorlinks]{hyperref}
\usepackage{orcidlink}

\usepackage{amsmath,amsfonts,bm}

\def\eqref#1{equation~\ref{#1}}
\def\floor#1{\lfloor #1 \rfloor}
\def\1{\bm{1}}

\DeclareMathAlphabet{\mathsfit}{\encodingdefault}{\sfdefault}{m}{sl}
\SetMathAlphabet{\mathsfit}{bold}{\encodingdefault}{\sfdefault}{bx}{n}

\usepackage{url}            % simple URL typesetting
\usepackage{booktabs}       % professional-quality tables

\usepackage{nicefrac}       % compact symbols for 1/2, etc.
\usepackage{microtype}      % microtypography
\usepackage{comment}
\usepackage{times}
\usepackage{epsfig}
\usepackage{graphicx}
\usepackage{subcaption}
\usepackage{amsmath}
\usepackage{amssymb}
\usepackage{multirow}
\usepackage{tikz} %for drawing trees
\usepackage[vlined,ruled]{algorithm2e}
\usepackage{amsfonts}       % blackboard math symbols

\newcommand{\red}[1]{\textcolor{red}{#1}}
\newcommand{\blue}[1]{\textcolor{blue}{#1}}

\newcommand{\black}[1]{\textcolor{black}{#1}}

\newcommand{\SupervisedColor}[1]{\textcolor[HTML]{6B76F9}{#1}}

\newcommand{\SWSLColor}[1]{\textcolor[HTML]{73B778}{#1}}
\newcommand{\HumanColor}[1]{\textcolor[HTML]{ff0000}{#1}}
\newcommand{\ViTColor}[1]{\textcolor[HTML]{DA73D5}{#1}}
\newcommand{\AdversarialColor}[1]{\textcolor[HTML]{DCD97C}{#1}}
\newcommand{\ClipColor}[1]{\textcolor[HTML]{5fe8e8}{#1}}

\definecolor{olivegreen}{cmyk}{.6,.4,0.8,0}
\definecolor{brickred}{rgb}{0.8, 0.25, 0.33}
\definecolor{green(pigment)}{rgb}{0.0, 0.65, 0.31}

\newcommand{\redb}[1]{\red{\textbf{#1}}}
\newcommand{\method}{VCR-Bench}

\newcommand{\SoTA}{SoTA}
\newcommand{\robustness}{VCR}
\newcommand{\robustnesssymbol}{\mathcal{R}}
\newcommand{\imagenetc}{\textsc{ImageNet-C}}

\newcommand{\RN}[1]{%
  \textup{\uppercase\expandafter{\romannumeral#1}}%
}

\newcommand{\gitlink}{https://github.com/HuakunShen/reliabilitycli}

\title{Assessing Visually-Continuous Corruption Robustness of Neural Networks Relative to Human Performance}

\begin{document}

%\makeatletter
%\def\thmheadbrackets#1#2#3{%
%  \thmname{#1}\thmnumber{\@ifnotempty{#1}{ }\@upn{#2}}%
%  \thmnote{ {\the\thm@notefont[#3]}}}
%\makeatother

%\newtheoremstyle{defbrakets}% Name
%  {}% space above
%  {}% space below
%  {\normalfont}% body font
%  {}% indent
%  {\bfseries}% head font
%  {.}% punctuation after head
%  { }% space after head (has to be space or dimension!)
%  {\thmheadbrackets{#1}{#2}{#3}}% head spec
%\theoremstyle{defbrakets}

%\theoremstyle{definition}
%\newtheorem{mydef}{Definition}
%\newtheorem{mysubdef}{Definition}[definition]

\titlerunning{Assessing Visually-Continuous Corruption Robustness}

% TODO FINAL: Replace with your author list. 
% Include the authors' OCRID for the camera-ready version, if at all possible.
\author{Huakun Shen\inst{1} \and
Boyue Caroline Hu\inst{1} \and
Krzysztof Czarnecki\inst{2} \and Lina Marsso\inst{1} \and Marsha Chechik\inst{1}}

% TODO FINAL: Replace with an abbreviated list of authors.
\authorrunning{H.~Shen et al.}
% First names are abbreviated in the running head.
% If there are more than two authors, 'et al.' is used.

% TODO FINAL: Replace with your institution list.
\institute{University of Toronto, Ontario, Canada \email{\{huakun.shen,lina.marsso\}@utoronto.ca \\ \{boyue,chechik\}@cs.toronto.edu} \and
University of Waterloo, Ontario, Canada\\
\email{kczarnec@gsd.uwaterloo.ca}}

\maketitle

\begin{abstract}
%\blue{Page limit: 9 pages of technical content plus additional pages solely for references}

%\red{Motivation: "better" comparison with humans for better understanding. To do this, we defined, and conducted experiments with people and some models. Our findings are ... }
%\red{NOTE: goal here is to show that we have this new idea of checking robustness and we conducted experiments to show this is promising, and in the meantime we provide tools to benchmark with our results. This can promote multiple directions of future research.}
%\red{the gap is bigger than we think because existing work does not check continuous changes, in this paper, we revisit corruption robustness and conduct a study...}
%\red{Page limit: 14 + references, deadline: Feb 29}
% robustness compared to humans
While Neural Networks (NNs) have surpassed human accuracy in image classification on ImageNet, they often lack robustness against 
%when faced with 
image corruption, i.e., corruption robustness.  Yet such robustness is seemingly effortless for human perception.
%To better understand NN robustness relative to humans, 
In this paper, we propose \emph{visually-continuous corruption robustness} (VCR) -- an extension of corruption robustness to allow assessing it over the wide and continuous range of changes that correspond to the human perceptive quality (i.e.,  from the original image to the full distortion of all perceived visual information),  along with two novel human-aware metrics for NN evaluation.
To compare VCR of NNs with human perception, we conducted extensive experiments on 14 commonly used image corruptions with 7,718 human participants and state-of-the-art robust NN models with different training objectives (e.g., standard, adversarial, corruption robustness), different architectures (e.g., convolution NNs, vision transformers), and different amounts of training data augmentation. 

% outcome: more space for improvement, similar transformations
% correspond to the new things we did when revisting: continous range allows us to see gaps undetected by existing benchmark; human aware metrics showed us to quantify this gap; extend to changes that affect humans: 1) how human compare with ML with big change; 2) similar changes affecting human, how they affect ML.
Our study showed that: 1) %for a parameterized image corruption, solely 
assessing robustness %through selected parameter values can lead to biased outcomes---a concern that the VCR effectively addresses (
against continuous corruption can reveal insufficient robustness undetected by existing benchmarks; as a result, 2) the gap between NN and human robustness is larger than previously known; and finally,
%There is still unclosed gap between NN and human robustness. 
3) some image corruptions have a similar impact on human perception, offering opportunities for more cost-effective robustness assessments.
Our validation set with 14 image corruptions, human robustness data, and the evaluation code is provided as a toolbox and a benchmark\footnote{\label{git}\gitlink}.

%In safety-critical domains such as autonomous driving, it is crucial to check robustness of Deep Neural Network (DNN) models against possible visual changes that can be encountered during deployment. In this paper, we introduce a new concept, \emph{visual-change robustness} (VCR), defined as the average-case DNN performance against a wide and continuous range of realistic visual changes. Additionally, since humans are currently the most robust decision-makers in complex, real-world scenarios, we propose a method for evaluating DNN \robustness{} relative to humans using two new human-aware evaluation metrics. We also propose a method to reduce the cost of measuring human robustness by identifying transformations with similar visual effects and reusing human performance data. Through evaluation with state-of-the-art (\SoTA) robust image-classification DNNs, we show that our method allows us to detect cases of insufficient robustness (compared to humans) that cannot be detected with the \SoTA{} benchmark. 
\end{abstract}

\section{Introduction}

%Neural Network (NN) models have been deployed in safety-critical systems, such as autonomous vehicles, that are subject to corruptions in their deployment environment (Hu et al., 2022a), including varying brightness and weather conditions. 

For Neural Networks (NN), achieving robustness against possible corruption (i.e., corruption robustness) that can be encountered during deployment
%such corruptions 
is essential for the application of NN models in safety-critical domains~\cite{hendrycks2019robustness}. Since NN models in these domains automate tasks typically performed by humans, it is necessary to compare the model's robustness with that of humans. %We have introduced two novel metrics that allow for a quantitative comparison between the model and humans. (contribution does not go here)

\vskip 0.1in
\noindent
\textbf{Human VS NN robustness.}
Corruption robustness measures the average-case performance of an NN or humans on a set of image corruption functions~\cite{hendrycks2019robustness}. 
Existing studies, including
out-of-distribution anomalies~\cite{Hendrycks2016}, benchmarking~\cite{hendrycks2019robustness,Hendrycks2019Natural}, and comparison with humans~\cite{hu-et-al-22,GeirhosNMTBWB21}, generally evaluate robustness against a pre-selected, fixed set of transformation parameter values that represent varying degrees of image corruption.
However, parameter values cannot accurately represent the degree to which human perception is affected by image corruptions. For instance, using the same parameter to brighten an already bright image will make the objects harder to see but will have the opposite effect on a dark image~\cite{hu-et-al-22}. Additionally, humans can perceive and generalize across a wide and continuous spectrum of visual corruptions from subtle to completely distorted~\cite{Geirhos2018GeneralisationIH,VIF}. Relying solely on preset parameter values for test sets could lead to incomplete coverage of the full range of visual corruptions, resulting in potentially biased evaluation that cannot accurately represent NN robustness compared with humans. 

\vskip 0.1in
\noindent
\textbf{Contributions and Outlook.}
To address the above problem, we propose a new concept called \emph{visually-continuous corruption robustness} (VCR). This concept focuses on the robustness of neural networks (NN) against a continuous range of image corruption levels. Additionally, we introduce two novel human-aware NN evaluation metrics (HMRI and MRSI) to assess NN robustness in comparison to human performance. We conducted extensive experiments with 7,718 human participants on the Mechanical Turk platform on 14 commonly used image transformations from three different sources\footnote{The number is comparable with 15 corruptions included in \imagenetc.}. Comparing NN and human VCR with our metrics, we found that a significant robustness gap between NNs and humans still exists: no model can fully match human performance throughout the entire continuous range in terms of both accuracy and prediction consistency, and few models can exceed humans by only a small margin in specific levels of corruption.
Furthermore, our experiments yield insightful findings about the robustness of human and state-of-the-art (SoTA) NNs concerning accuracy, degrees of visual corruption, and consistency of classification, which can contribute towards the development of NNs that match or surpass human perception. We also discovered classes of corruption transformations 
for which humans showed similar robustness 
%that affect human robustness similarly, 
(e.g., different types of noise), while NNs reacted differently. Recognizing these classes can contribute to reducing the cost of measuring human robustness and elucidating the differences between humans and computational models. 
To foster future research, we open-sourced all human data as a comprehensive benchmark along with a Python code that enables test set generation, testing, and retraining.

%Using these two metrics, we found that despite some models showcasing super-human accuracy, a significant robustness gap still exists between NNs and humans.

%To do this, we conducted extensive experiments with 7,718 human participants on the Mechanical Turk platform on 14 commonly used image transformations. To foster future research, we open-sourced all human data as a comprehensive benchmark along with a Python toolbox that enables test set generation, testing, and retraining. Our experiments yield insightful findings about human and state-of-the-art (SoTA) NN robustness concerning accuracy, degrees of visual corruption, and consistency of classification, which can contribute towards NNs that better represent human perception. Additionally, we discovered classes of corruption transformations that affect human robustness similarly, (e.g., Shot Noise and Gaussian Noise in Fig. 1), while NNs react differently. Recognizing these classes can contribute to reducing the cost of measuring human robustness and elucidating the differences between humans and computational models.
\section{Related Work}
\label{sec:related}
%In this section, we summarize related work in robustness related to adversarial robustness, benchmarks and comparisons between humans and DNNs. 
%\red{See more related work in sup}
%\red{TODO: add Partial success in closing the gap between human and machine vision}
% comparing with humans
%\noindent
%\textbf{Comparison between Humans and DNNs.}
We briefly review related work on the comparison of human and NN robustness, adversarial robustness, robustness benchmarks and improving robustness.%; a more extensive review of related work on robustness can be found in the appendix.

\vskip 0.1in
\noindent
\textbf{Human VS NN Robustness.} Prior studies have used human performance to study the existing differences between humans and neural networks~\cite{Firestone26562,ZhangIESW18}, to study invariant transformations~\cite{kheradpisheh2016deep}, to compare recognition accuracy~\cite{HoPhuoc2018CIFAR10TC, STALLKAMP2012323}, to compare robustness against image transformations~\cite{Geirhos2018GeneralisationIH,GeirhosNMTBWB21}, or to specify expected model behaviour~\cite{hu-et-al-22}. %\red{comparison with Partial success in closing the gap between human and machine vision, also comparison written in def before}
The main difference between our study and existing work, specifically, the most recent study by~\cite{GeirhosNMTBWB21}, is three-fold: 1) we are the first to quantify robustness across the full continuous visual corruption range\black{, thus revealing previous undetected robustness gap}; 2) our experiments for obtaining human performance are designed to include more participants for measuring the \emph{average} human robustness\black{, resulting in more generalizable results and reduced influence of outliers}; 3) we identified visually similar transformations for humans but not NNs\black{, potentially reducing experiment costs}. %\blue{As a result, we showed that VCR revealed a larger robustness gap between NNs and humans considering continuous changes unable to be picked up by~\cite{GeirhosNMTBWB21}.}
\vskip 0.1in
\noindent
\textbf{Robustness Benchmarks.} Several robustness benchmarks have been developed. Hendrycks et al. built the \imagenetc{} and \textsc{-P} benchmarks for checking NN model classification robustness against common corruptions and perturbations on \textsc{ImageNet} images~\cite{hendrycks2019robustness}.
%, and this 
They have inspired other benchmarks for different corruption functions, datasets, and tasks~\cite{KarYAZ22, ChattopadhyayHM21, KamannR21, michaelis2019dragon, MintunKX21,Sun2022, YiYLTK21}. However, these benchmarks generate images by applying corruption functions with only five pre-selected values per parameter. \textsc{ImageNet-CCC}~\cite{press2023rdumb} is the only prior work targeting a more continuous range of corruptions, by using 20 pre-selected values per parameter. It does not check the coverage in terms of the visual effects on the images, which we do with an Image Quality Assessment (IQA) metric Visual Information Fedility (VIF)~\cite{VIF}. Further, this work focuses on continuous changes over time for benchmarking test-time adaptation, which is different from a general robustness benchmark, and the dataset has not been released as the time of writing.
In contrast to all these previous works, our method randomly and uniformly samples parameter values to cover the full range of visual change that a corruption function can achieve, which is modeled and assessed for coverage using an IQA metric. Finally, our work compares robustness of NNs with humans.

%adversarial benchmarks: robustbench
%imagenet 3dcc
%imagenet c
%problem: distribution of parameters (coverage), no comparison to humans.

\vskip 0.1in
\noindent
\textbf{Adversarial Robustness.} Adversarial robustness measures the worst-case performance on images with added `small' distortions or perturbations tailored to confuse a classifier~\cite{hendrycks2019robustness}. However, changes that can be encountered in the real-world situations are often of a much bigger range~\cite{KarYAZ22}. Thus, in this paper, we focus on \emph{average-case performance} over a \emph{realistic} range of changes. 

\vskip 0.1in
\noindent
\textbf{Improving Robustness.} Numerous methods for improving model robustness have been proposed, e.g., data augmentation with corrupted data~\cite{GeirhosJMZBBW20, Lopes2019, MadryMSTV18, Rusak2020}, texture changes~\cite{GeirhosRMBWB19, HendrycksBMKWDD21}, image compositions~\cite{YunHCOYC19, ZhangCDL18} and corruption functions~\cite{YinLSCG19, Hendrycks_AugMix}.  All of these have different abilities to generalize to unseen data~\cite{KarYAZ22}. 
While not our primary focus, we demonstrate that NN robustness compared to humans can be improved through data augmentation and fine-tuning with our generated images for VCR.

%Although improving robustness is not the main focus of this paper, we show that NN robustness relative to humans can be improved through data augmentation and fine-tuning with our generated images for VCR.% (see Sec.~\ref{sec:train}). 

%\begin{itemize}
%    \item Other benchmark papers (human vs non-human)
%    \item robustness evaluation paper (adversarial vs bigger range of changes)
%    \item What is the limitation of them?
%    \begin{itemize}
%        \item Not comparing to human performance.
%        \item High performance with iid assumption (testing and training data come from the same distribution)
%        \begin{itemize}
%            \item poor performance otherwise
%            \item Distribution Shift
%        \end{itemize}
%    \end{itemize}
    
%\end{itemize}

%\section{Methods: VCR, testing, metrics, crowdsourcing, NN models}
\section{Visually-Continuous Corruption Robustness (VCR)}
\label{sec:method}
%\blue{goal here is to separate intuition with definition so it's easier to understand}
To study NN robustness against a wide and continuous spectrum of visual changes, we first define VCR and then describe our method for generating test sets. To study VCR of NNs in relation to humans, we also present the human-aware metrics.%, followed by human robustness data and NN models used in the study.

%1. better represent human opinions + the range of "reasonable" changes
%\noindent
\subsection{Visually-Continuous Corruption Robustness (VCR) Definition}
%\subsubsection{Visual-corruption Robustness (VCR).}
%\blue{Corruption robustness measures the average-case performance of an NN on corruptions~\cite{hendrycks2019robustness} and existing benchmarks for corruption robustness~\cite{KarYAZ22, ChattopadhyayHM21, KamannR21, michaelis2019dragon, MintunKX21,Sun2022, YiYLTK21,GeirhosNMTBWB21} use test datasets obtained by transforming original images with a pre-selected list of parameter values for parameterized image corruption transformations. A key difference between corruption robustness and VCR is that rather than focusing on the parameter domain, VCR is defined relative to the impact of image corruptions on human perception.
%This is important because parameter values cannot accurately represent corruption as perceived by humans. For instance, using the same parameter to brighten an already bright image will make the objects harder to see; but will have the opposite effect on a dark image~\cite{hu-et-al-22}. }
A key difference between corruption robustness and VCR is that the latter
is defined relative to the \emph{visual impact} of image corruption on human perception, 
rather than the transformation parameter domain.
To quantify visual corruption, VCR uses the Image Quality Assessment (IQA) metric Visual Information Fidelity (\textit{VIF})~\cite{VIF, VIF_python}. VIF measures the perceived quality of a corrupted image $x'$ compared to its original form $x$ by measuring the visual information unaffected by the corruption. Thus, we define the \emph{change} in the perceived quality caused by the corruption as $\Delta_v(x, x') = max(0, 1-\textit{VIF}(x, x'))$. See Appendix.~\ref{sec:app_vcr} for more detail on $\Delta_v$. With $\Delta_v$, whose value ranges from 0 and 1, we can consider VCR against the wide, finite, and continuous spectrum of visual corruptions ranging from no degradation to visual quality (i.e., the original image) ($\Delta_v=0$) to the full distortion of all visual information ($\Delta_v=1$).  %For example, the minimal blurring in Fig.~\ref{fig:a} caused a minimal change in visual quality of $\Delta_v = 0.02$, whereas the more pronounced blurring in Fig.~\ref{fig:b} corresponds to $\Delta_v = 0.94$. 

\noindent
\underline{Limitation:} VCR is limited to image corruption that is applicable to the chosen IQA metric, thus by using VIF, VCR is limited to only pixel-level corruption.
%
%only applicable to  
%
%By using VIF, VCR is limited to pixel-level corruption only.
%By applying VIF, VCR is only applicable to pixel-level corruptions that can be measured with VIF
%Since VIF is used, VCR is only applicable to pixel-level corruptions that can be measured with VIF. 
Further research is needed for metrics suitable for other types of corruption (e.g., geometric).

%returns a value between 0 and 1, with 0 indicating no degradation to visual quality (i.e., original image) and 1 indicating that all visual information has been corrupted Using $\Delta_v$ in VCR allows us to consider corruption robustness relative to a wide, finite, and continuous spectrum of degradation of human perceived quality ranging from the original image to the full distortion of all visual information. 

%2. not affected by distribution of different levels of corruptions in the tests
%3. also represent preservation
%\red{TODO: Update definition for more clarity + introduce preservation + comment about distribution}

For VCR, we consider a classifier NN $f: X \rightarrow Y$ trained on samples of a distribution of input images $P_X$, a ground-truth labeling function $f^*$, and a parameterized image corruption function $T_X$ with a parameter domain $C$. We wish to consider the robustness of $f$ against images with all degrees of visual corruption \emph{uniformly} ranging from $\Delta_v = 0$ to $\Delta_v =1$.\footnote{Note that distributions other than uniform can be used based on the application. For example, one may wish to favour robustness against heavy snow conditions for NNs deployed in arctic areas.} Therefore, given a value $v \in [0,1]$, we define $P(x,x'|v)$ as the \emph{joint distribution} of original images ($x$) and corresponding corrupted images ($x' =T_X(x, c)$, $c\in C$) with $\Delta_v(x, x')=v$.
VCR is defined in the presence of a robustness property $\gamma$ that $f$ should satisfy in the presence of $T_X$: 
%A generic form of VCR $\robustnesssymbol$ for a given robustness property $\gamma$ that $f$ should satisfy in the presence of $T_X$ is: 
\begin{align}
 \robustnesssymbol_{\gamma} &= \mathbb{E}_{v \sim \textit{Uniform(0,1)}}(P_{x, x'\sim P(x,x'|v)}(\gamma)).
\end{align}

\begin{figure}[t]
   \centering
   \includegraphics[width=\linewidth]{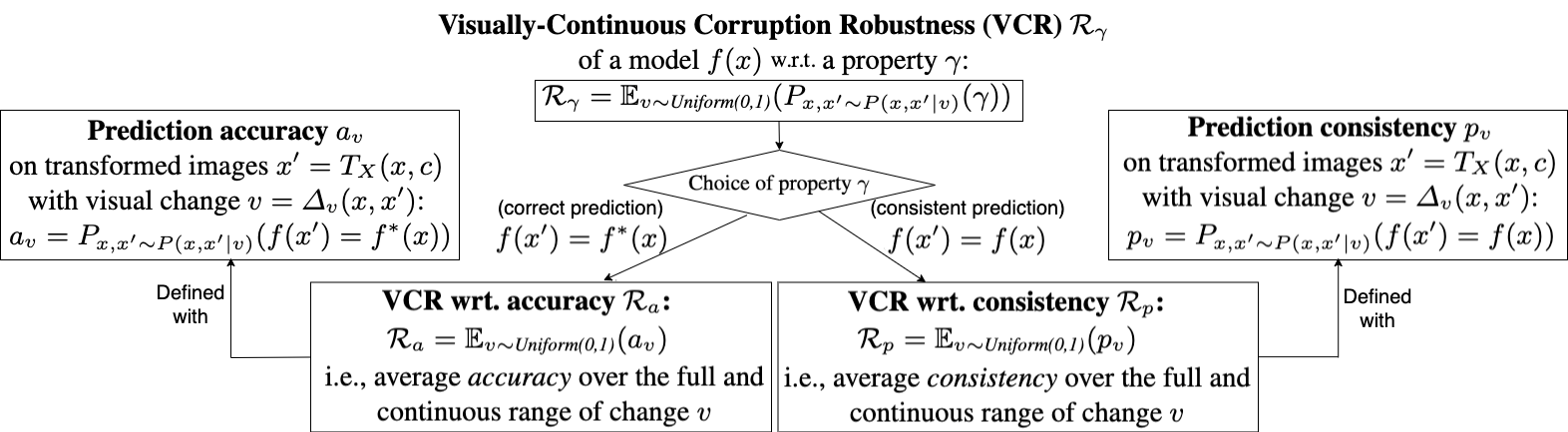}
    \caption{\small  Summary of VCR definitions with respect to accuracy and consistency. }
    \label{fig:vcr}
    \vspace{-0.25in}
\end{figure}

\noindent
In this paper, we instantiate VCR with two existing robustness properties (see Fig.~\ref{fig:vcr}). The first one is \emph{accuracy ($a$)}, requiring that the prediction on corrupted images should be correct, i.e., $f(x') = f^*(x)$. It is also used in the existing definition of corruption robustness~\cite{hendrycks2019robustness}. Thus, 
\begin{align}
\robustnesssymbol_a &= \mathbb{E}_{v \sim \textit{Uniform(0,1)}}(P_{x, x'\sim P(x,x'|v)} (f(x') = f^*(x))).
\end{align}
\noindent
The second property is \emph{prediction consistency ($p$)}, requiring consistent predictions before and after corruptions, i.e., $f(x') = f(x)$~\cite{hu-et-al-22}. It is applicable when ground truth is not available, which is common during deployment. Thus, 
\begin{align}
\robustnesssymbol_p &= \mathbb{E}_{v \sim \textit{Uniform(0,1)}}(P_{x, x'\sim P(x,x'|v)} (f(x') = f(x))).
\end{align}

\vskip 0.1in
\noindent
{\bf Summary of VCR Definitions}. Fig.~\ref{fig:vcr} gives a visual summary of the VCR metrics, starting with the general definition $\robustnesssymbol_\gamma$ at the top, and instantiating it for accuracy as $\robustnesssymbol_a$ and consistency as $\robustnesssymbol_p$. Each of them is simply the average accuracy or prediction consistency, respectively, over the full and continuous range of visual change.

\begin{comment}
    
\begin{definition}[\textbf{VCR ($\robustnesssymbol$)}]\label{def:visual-change-robustness}
Let a classifier $f$, a distribution of input images $P_X$, a ground-truth labeling function $f^*$, a transformation $T_X$ with parameter domain $C$,  % and parameter distribution $P_C$
and the visual change measure $\Delta_v$ be given. 
Given a $\Delta_v$ value $v \in [0,1]$, the \emph{joint distribution of original and transformed images with $v$} is $P(x,x'|v) = \frac{1}{P(v)}\int_C P(x’,v|x,c)P_X(x)P(c) dc$, where $x' =T_X(x, c)$, $v =\Delta_v(x, x')$ and $P(c)$ is uniform. 

\red{first general with a property, then we say we instantiate with accuracy and prediction similarity}

\begin{subdefinition}[\textbf{VCR w.r.t. accuracy ($\robustnesssymbol_a$)}]\label{def:visual-change-robustness-acc}
Visual-change robustness of a classifier $f$ for accuracy is %$\mathbb{E}_{v \sim Uniform(0,1)}(P_{x, x_i \sim P_{TX, v}} (f(x_i) = f^*(x)))$. 
$\robustnesssymbol_a = \mathbb{E}_{v \sim \textit{Uniform(0,1)}}(P_{x, x'\sim P(x,x'|v)} (f(x') = f^*(x)))$.
\end{subdefinition}

\begin{subdefinition}[\textbf{VCR w.r.t. prediction similarity ($\robustnesssymbol_p$)}]\label{def:visual-change-robustness-pred-sim}
Visual-change robustness of a classifier $f$ for prediction similarity is %$\mathbb{E}_{v \sim Uniform(0,1)}(P_{x, x_i \sim P_{TX, v}} (f(x_i) = f(x)))$. 
$\robustnesssymbol_p = \mathbb{E}_{v \sim \textit{Uniform(0,1)}}(P_{x, x'\sim P(x,x'|v)} (f(x') = f(x)))$.
\end{subdefinition}
\label{def:robustness}
\end{definition}
\end{comment}

\subsection{Testing VCR} 
%\subsubsection{Testing VCR.}
\robustness{} of a subject (a human or an NN) is measured by first generating a test set through sampling and then estimating it using the sampled data. 
%To measure \robustness{}, our method \method{} generates a validation set (Step~\RN{1} of Fig.~\ref{fig:intro-paper-design-flowchart}) for a given transformation $T_X$. 
The test set is generated by sampling images and applying corruption to obtain $P(x,x’|v)$ for different $\Delta_v$ values $v$. We sample $x\sim P_X$ and $c \sim \textit{Uniform}(C)$, and obtain $x'=T_X(x, c)$ and $v= \Delta_v(x, x')$, resulting in samples $(x, x', c, v)$. % We divide a continuous parameter domain $C$ into $M$ random levels for sampling.
Then, we divide them into groups of $(x,x’,c)$, each with the same $v$ value. %Thus, the resulting groups are samples from $P(x,x’,c|v)$.
Next, by dropping $c$, we obtain groups of $(x,x’)$ with the same $v$, which are samples from $P(x,x’|v)$. Note that this procedure requires only sufficient data in each group but not uniformity, i.e., $v \sim \textit{Uniform}(0,1)$ is not required. The varying size of each group, i.e., the non-uniformity of $v$ distribution, will not distort VCR estimates, but only impact the estimate uncertainty at a given $v$. Further, interpolation in the next step helps address any missing points. 

%This specific design of VCR estimation removes the possibility of biased results due to a biased distribution of test data.
%\blue{Note that $P(x,x’|v)$ is individually sampled and estimated for each distinct $v$. Consequently, to estimate VCR, one only needs sufficient data in each $v$ bin to measure performance for every $P(x,x’|v)$. 
%The distribution of $\Delta_v$ values in all test data, i.e., the varying number of images per $\Delta_v$ bin, by definition would not affect VCR estimation. This specific design of VCR removes the possibility of biased results due to a biased distribution of test data.
%dependency of robustness results on the distribution of data sampled.
%removes the limitation of existing robustness evaluation, which is results rely heavily on the distribution of data sampled. 
%}

With the test set, we estimate the performance w.r.t. the property $\gamma$ for each $v$. For each $v$ in the test data, we compute the \emph{rate} of accurate predictions $f(x') = f^*(x)$ to estimate accuracy, i.e., $a_v = P_{x, x'\sim P(x,x'|v)} (f(x') = f^*(x))$ [resp. consistent predictions $f(x') = f(x)$ to estimate consistency, i.e., $p_v = P_{x, x'\sim P(x,x'|v)} (f(x') = f(x))$]. Then by plotting $(v, a_v)$ and $(v,p_v)$ and applying monotonic smoothing splines~\cite{koenker-94} to reduce randomness and outliers, we obtain smoothed spline curves $s_{a}$ and $s_p$, respectively. The curves $s_{\gamma}$ (namely, $s_{a}$ and $s_p$) describe how the performance w.r.t. the robustness property $\gamma$ (namely, $a$ and $p$) decreases as the visual corruption in images increases. Finally, 
we estimate $\robustnesssymbol_a = \mathbb{E}_{v \sim \textit{Uniform(0,1)}}(a_v)$ [resp. $\robustnesssymbol_p =\mathbb{E}_{v \sim \textit{Uniform(0,1)}}(p_v)$] as the area under the spline curve, i.e., $\hat{\robustnesssymbol}_a = A_a = \int_0^{1} s_a(v) dv$ [resp. $\hat{\robustnesssymbol}_p =A_p = \int_0^{1} s_p(v) dv$]. See Alg.~\ref{alg:vcr-est} in the Appendix for the pseudo-code of VCR estimation.

\subsection{Human-Aware Metrics for VCR}
%\subsubsection{Human-aware Metrics.}
A commonly used metric for  measuring corruption robustness is the \emph{Corruption Error (CE)}~\cite{hendrycks2019robustness}---the top-1 classification error rate on the corrupted images, normalized by the error rate of a baseline model. CE can be used to compare an NN with humans if the baseline model is set to be humans. However, CE is not able to determine whether an NN can exceed humans, and NN models could potentially have super-human accuracy for particular types of perturbations or in some $\Delta_v$ ranges.
%as shown in blue in Fig.~\ref{fig:metric-human-ml-difference}. 
Therefore, inspired by CE, we propose two new human-aware metrics, \emph{Human-Relative Model Robustness Index} (\emph{HMRI}) that measures NN \robustness{} relative to human \robustness{}; and \emph{Model Robustness Superiority Index} (\emph{MRSI}) that measures how much an NN exceeds human \robustness{}.
%incapable of indicating whether a DNN can exceed human.}
%DNN models could potentially have super-human accuracy for particular types of perturbations or in some $\Delta_v$ ranges. 
%Thus we also want to measure how much a DNN can \emph{exceed} human performance. As shown in  Fig.~\ref{fig:metric-human-ml-difference}, $A_{m>h}$, highlighted in blue, represents the area where DNN performance exceeds a human; and $A_{m}-A_{m>h}$, highlighted in green (equal to $A_{h}-A_{h>m}$), represents human performance that the DNN can reach. \red{here is motivation, move up}
%\red{we give the two metrics, then we say plugging in the accuracy and prediction definition we get ...}
%

\begin{figure}[t]
    \centering
    \begin{subfigure}[b]{0.34\linewidth}
        \includegraphics[width=1.2\linewidth]{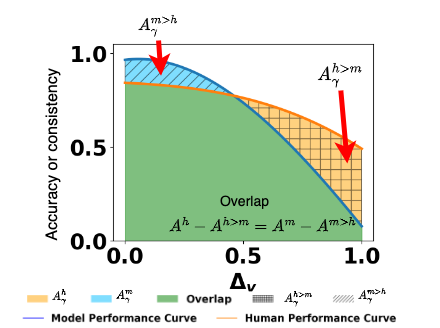}
    %\vspace*{-15mm}
    \caption{Visualization of auxiliary metrics for model vs. human performance.}
    \label{fig:aux}
    \end{subfigure}
    \begin{subfigure}[b]{0.65\linewidth}
    \centering
    \scalebox{0.7}{
        \begin{tabular}{|c|} \hline 
         \textbf{Auxiliary metric} (cf. Fig.~\ref{fig:aux})\\ \hline 
         \shortstack{\textbf{VCR of humans} wrt. a property $\gamma$,\\ estimated as an area under performance curve $A^h_\gamma$: \\ $\hat{\robustnesssymbol}_\gamma^h=A^h_\gamma=\int_0^{1} \ s^h_\gamma(v) dv$ }\\ \hline 
         \shortstack{\textbf{VCR of a model} $f(x)$ wrt. a property $\gamma$,\\ estimated as an area under performance curve $A^m_\gamma$:  \\$\hat{\robustnesssymbol}_\gamma^m=A^m_\gamma=\int_0^{1} s^m_\gamma(v) dv$} \\ \hline 
         \shortstack{\textbf{VCR lead of humans over a model} $f(x)$ wrt. a property $\gamma$,\\ estimated as a difference area $A^{h>m}_\gamma$: \\ $\hat{\robustnesssymbol}_\gamma^{h>m}=A^{h>m}_\gamma$ $=\int_0^{1} \max(0,s^h_\gamma(v)-s^m_\gamma(v)) dv$} \\ \hline 
         \shortstack{\textbf{VCR lead of a model} $f(x)$ \textbf{over humans} wrt. a property $\gamma$,\\estimated as a difference area $A^{m>h}_\gamma$: \\ $\hat{\robustnesssymbol}_\gamma^{m>h}=A^{m>h}_\gamma$ $=\int_0^{1} \max(0,s^m_\gamma(v)-s^h_\gamma(v)) dv$} \\ \hline
    \end{tabular}}
    \caption{Summary of auxiliary metrics for defining \emph{HMRI} and \emph{MRSI}.}
    \label{tab:aux}
    \end{subfigure}
    
    %\vspace{0.05in}
    %\begin{subfigure}[b]{\linewidth}
    %\centering
    %\scalebox{0.7}{
    %        \begin{tabular}{|c|} \hline 
    %     \textbf{Human-model comparison metric}\\ \hline 
    %     \textbf{Human-Relative Model Robustness Index (HMRI)} of a model $f(x)$ wrt. a property $\gamma$, i.e., $\textit{HMRI}_{\gamma}=\frac{A^h_{\gamma}-A^{h>m}_\gamma}{A^h_{\gamma}} =1-\frac{A^{h>m}_{\gamma}}{A^h_{\gamma}}$ \\ \hline 
    %     \textbf{Model Robustness Superiority Index (MRSI)} of a model $f(x)$ wrt. a property $\gamma$, i.e.,  $\textit{MRSI}_{\gamma}=\frac{A^{m>h}_{\gamma}}{A^m_{\gamma}}$ \\ \hline
    %\end{tabular}}
    %\caption{\emph{HMRI} and \emph{MRSI} definitions (using auxiliary metrics from Tab.~\ref{tab:aux}).}
    %\label{tab:indices}
    %\end{subfigure}
    \vspace{-0.1in}
    \caption{\small Auxiliary VCR metrics to compute \emph{HMRI} and \emph{MSRI}.}
    \label{fig:all_aux}
    \vspace{-0.25in}
\end{figure}

\vskip 0.1in
\noindent
{\bf Auxiliary VCR metrics to compute \emph{HMRI} and \emph{MRSI} }. 
%These metrics 
\emph{HMRI} and \emph{MRSI} take as inputs the estimated spline curves for humans ($s^h_{\gamma}$) and for NN ($s^m_{\gamma}$). %both the estimated spline curve for humans, $s^h_{\gamma}$, and for NN, $s^m_{\gamma}$, 
We denote areas under these curves as $A^h_{\gamma}$ and $A^m_{\gamma}$, respectively (see Fig.~\ref{fig:aux}). 
%A mock example of $s^h_{a}$ and $s^m_{a}$ is shown in Fig.~\ref{fig:metric-human-ml-difference}.
To compare NN model and human performance, VCR w.r.t. prediction consistency or accuracy is estimated using Alg.~\ref{alg:vcr-est} using both model and human performance data, as illustrated by the yellow ($A^h_{\gamma}$) and blue ($A^m_{\gamma}$) areas Fig.~\ref{fig:aux}, respectively. Both the blue and yellow areas also include the green area representing their overlap. Additionally, the VCR lead of humans over a model $A^{h>m}_{\gamma}$, the girded area in Fig.~\ref{fig:aux}, and the VCR lead of a model over humans $A^{m>h}_{\gamma}$, the striped area in Fig.~\ref{fig:aux}, are estimated. The definitions of these four auxiliary metrics are summarized in Tab.~\ref{tab:aux}, and they are used to define \emph{HMRI} and \emph{MRSI}.%, as shown in  Tab.~\ref{tab:indices}.
%: (i) HMRI, which characterizes the human lead in VCR over the model; and (ii) MRSI, which characterizes the model lead in VCR over humans. Their definitions are summarized in Tab.~\ref{tab:indices}.

%Using human robustness data collected in Step 2 (Fig.~\ref{fig:intro-paper-design-flowchart})  and DNN predictions, we first obtain the following spline curves: for $\robustnesssymbol_a$, 1) $s^h_a$ for humans and 2) $s^m_a$ for the DNN, see Fig.~\ref{fig:metric-human-better-ml}-\ref{fig:metric-human-ml-difference};  similarly for $\robustnesssymbol_p$, 1) $s^h_p$ for humans and 2) $s^m_p$ for the DNN. 

%We then use two new
%In this section, we propose our new 
%human-aware evaluation metrics  \emph{Human-Relative Model Robustness Index} (\emph{HMRI}) and  \emph{Model Robustness Superiority Index} (\emph{MRSI}). %that takes human performance into account and measures how DNN robustness is proportional to human performance.
% Definition starts here
\begin{definition}[\textbf{Human-Relative Model Robustness Index (\textit{HMRI})}]
%Checks how much human performance DNN models can achieve.\\
Given $s_{\gamma}^h$ and $s_{\gamma}^m$, let $A_\gamma^{h>m} = \int_0^{1}(s^h_{\gamma}(v)-s^m_{\gamma}(v))^+ dv$ denote the average (accuracy or preservation) performance lead of humans over a model across the visual change range, where the performance lead is defined as the positive part of performance difference, i.e., $(s^h_{\gamma}(v)-s^m_{\gamma}(v))^+=max(0,s^h_{\gamma}(v)-s^m_{\gamma}(v))$. \textit{HMRI}, which quantifies the extent to which a DNN can replicate human performance, is defined as $\frac{A^h_{\gamma}-A^{h>m}_\gamma}{A^h_{\gamma}}=1-\frac{A^{h>m}_\gamma}{A^h_{\gamma}}$.
\label{def:metric-HMRI}
\end{definition}

%\textit{HMRI} takes both DNN and human performance into the equation. As shown in Fig.~\ref{fig:metric-human-better-ml} and Fig.~\ref{fig:metric-human-ml-difference},
%A mock example with $s_{a}^h$ and $s_{p}^m$ is shown in Fig.~\ref{fig:metric-human-ml-difference}. $A_{h>m}$, highlighted \blue{with grid shadow}, represents the area where human performance is better than DNN; and $A_{h}-A_{h>m}$, \blue{not shadowed but }highlighted in green, represents %the \blue{percentage} the area where the DNN can be comparable to human performance.
%\blue{how much human performance the DNN can be comparable with. }%of human performance that the DNN can reach. Thus, 
%\textit{HMRI} is calculated as the area of the green region over the entire $A_h$, which is $\frac{A_{h}-A_{h>m}}{A_h}$. 

\noindent
The \textit{HMRI} value ranges from $[0,1]$; a higher \textit{HMRI} indicates a NN model closer to human \robustness{}, and $\textit{HMRI}=1$ signifies that $s^m$ is the same as or completely above $s^h$ in the entire $\Delta_v$ domain, i.e., the NN is at least as robust as an average human (see Fig.~\ref{fig:aux}).%  {\bf MC:  until now we used NNs}

\begin{definition}[\textbf{Model Robustness Superiority Index (\textit{MRSI})}]
%Checks how much DNN can exceed human.\\
Given $s_{\gamma}^h$ and $s_{\gamma}^m$, let $A^{m>h}_\gamma = \int_0^{1}(s_{\gamma}^m(v)-s_{\gamma}^h(v))^+ dv$ denote the average performance lead of a model over a human across the visual change range.  \textit{MRSI}, which quantifies the extent to which a DNN model can surpass human performance, is defined as $\frac{A^{m>h}_\gamma}{A^m_\gamma}$.
\label{def:metric-MRSI}
\end{definition}

% In Figure \ref{fig:metric-human-ml-difference}, HMRI is represented as the ratio of the green region to the combined green and yellow region. Similarly, MRSI is depicted as the ratio of the blue region to the combined blue and yellow region.
% \red{update to match the new $M_{m>h}$.}
% \red{It is possible that ...}
%also can be interpreted as the area under the ML performance curve $m$ such that DNN does not exceed humans. 
%In Fig.~\ref{fig:metric-human-ml-difference}, $A_{m>h}$, highlighted with line shadow {\bf MC:  what are line and grid shadows?}, represents the area where the NN performs  better than humans.
%Thus, \textit{MRSI} is calculated as the green area over the entire $A_m$, which is $\frac{A_{m}-A_{m>h}}{A_m}$. %Intuitively, $P_{h-m}$ measures how much DNN models can outperform humans. 

\noindent
The \textit{MSRI} value ranges from $[0,1)$, with the higher value indicating better performance than humans. $\textit{MSRI}=0$ means that the given NN model performs worse than or equal to humans in the entire $\Delta_v$ domain. A positive \textit{MSRI} value indicates that the given NN model performs better than humans at least in some ranges of $\Delta_v$ (see Fig.~\ref{fig:aux}).

%{\bf MC:  again, DNN vs NN}
Comparing humans and NNs with \textit{HMRI} and \textit{MRSI}  yields three possible scenarios: (1)  humans' performance fully exceeds NN's, i.e., $0<\textit{HMRI}<1$ and $\textit{MRSI}=0$; %, see Fig.~\ref{fig:metric-human-better-ml}; 
(2)  NN's performance fully exceeds humans', i.e., $\textit{HMRI}=1$ and $\textit{MRSI} > 0$
%, see Fig.~\ref{fig:metric-ml-better-huamn}
; and (3)  humans' performance is better than NN's in some $\Delta_v$ intervals and worse in others, i.e., $\textit{HMRI} < 1$ and $\textit{MRSI} > 0$.%, see Fig.~\ref{fig:metric-human-ml-difference}.

\noindent
\section{Experiments}
\label{sec:experiments}

In this section, we describe experiments that check the VCR of NN models against human performance.

\vskip 0.1in
\noindent
\textbf{NN models.} 
%\subsection{Models Included in our Study}
Tab.~\ref{tab:models} summarizes the models included in our study. We have selected a wide range of architectures (different CNN and transformer architectures) and training methods (supervised, adversarial, semi-weakly, and self-supervised), including dinov2\_giant~\cite{oquab2023dinov2}, which is on the top of the \textsc{ImageNet-C} leaderboard as of time of writing. In total, we studied 11 \SupervisedColor{standard supervised models}, 4 \AdversarialColor{adversarial learning models}, 2 \SWSLColor{SWSL models}, 1 \ClipColor{CLIP} (clip-vit-base-patch32) model and 3 \ViTColor{ViT models}.
%In total, we studied 11 \SupervisedColor{standard supervised models}: \textsc{NoisyMix}, \textsc{NoisyMix\_new}~\cite{NoisyMix}, \textsc{SIN}, \textsc{SIN\_IN}, \textsc{SIN\_IN\_IN}, \textsc{HMany}, \textsc{HAugMix}~\cite{Hendrycks_AugMix}, \textsc{Standard\_R50}~\cite{pytorch}, \textsc{AlexNet}~\cite{KrizhevskySH12}; 4 \AdversarialColor{adversarial learning models}: \textsc{Do\_50\_2\_Linf}~\cite{SalmanIEKM20}, \textsc{Liu-Swin-L}, \textsc{Liu-ConvNeXt-L}~\cite{liu2023comprehensive}, \textsc{Singh-ConvNeXt-L-ConvStem}~\cite{singh2023revisiting}; 2 \SWSLColor{SWSL models}: \textsc{swsl\_resnet18}, \textsc{swsl\_resnext101\_32x16d}~\cite{DBLP:journals/corr/abs-1905-00546}; 3 \ViTColor{ViT models}: \textsc{Tian\_DeiT-S}, \textsc{Tian\_DeiT-B}~\cite{tian2022deeper}, \textsc{dinov2\_giant}~\cite{oquab2023dinov2}; and 1 \ClipColor{CLIP (clip-vit-base-patch32) model}~\cite{radford2021learning}.
For CLIP, we used the prompt ``a picture of (ImageNet class)'' while tokenizing the labels.

\begin{table}[t!]
\centering
%\vspace{-0.1in}
\scalebox{0.6}{
    \begin{tabular}{|lll|lll|}  
    \hline
    \textbf{Model}                & \textbf{Architecture}  & \textbf{Training Method} & \textbf{Model}                & \textbf{Architecture}  & \textbf{Training Method} \\ 
    
    \hline
    
    \textsc{NoisyMix}~\cite{NoisyMix}           & ResNet-50             & \SupervisedColor{Supervised}             &
    \textsc{NoisyMix\_new}~\cite{NoisyMix}      & ResNet-50              & \SupervisedColor{Supervised}             \\
    \textsc{SIN}~\cite{GeirhosRMBWB19}               & ResNet-50              & \SupervisedColor{Supervised}              &
    \textsc{SIN\_IN}~\cite{GeirhosRMBWB19}           & ResNet-50              & \SupervisedColor{Supervised}         \\
    \textsc{SIN\_IN\_IN}~\cite{GeirhosRMBWB19}       & ResNet-50              & \SupervisedColor{Supervised}             &
    \textsc{HMany}~\cite{HendrycksBMKWDD21}              & ResNet-50              & \SupervisedColor{Supervised}            \\ 
    \textsc{HAugMix}~\cite{Hendrycks_AugMix}            & ResNet-50              & \SupervisedColor{Supervised}            &
    \textsc{Standard\_R50}~\cite{Resnet50}                  & ResNet-50              & \SupervisedColor{Supervised}        \\ 
    \textsc{AlexNet}~\cite{KrizhevskySH12}                        & AlexNet              & \SupervisedColor{Supervised}              &
    \textsc{Tian\_DeiT-S}~\cite{tian2022deeper}           & DeiT Small           & \SupervisedColor{Supervised} \ViTColor{ViT}                      \\ 
    \textsc{Tian\_DeiT-B}~\cite{tian2022deeper}           & DeiT Base           & \SupervisedColor{Supervised} \ViTColor{ViT}                   &
    \textsc{Do\_50\_2\_Linf}~\cite{SalmanIEKM20}      & WideResNet-50-2              & \AdversarialColor{Adversarial}           \\ 
    \textsc{Liu\_Swin-L}~\cite{liu2023comprehensive}     & Swin-L              & \AdversarialColor{Adversarial}       &
    \textsc{Liu\_ConvNeXt-L}~\cite{singh2023revisiting} & ConvNeXt-L              & \AdversarialColor{Adversarial}           \\
    \textsc{Singh\_ConvNeXt-L-ConvStem}~\cite{singh2023revisiting}  & ConvNeXt-L + ConvStem              & \AdversarialColor{Adversarial}          &
    \textsc{swsl\_resnet18}~\cite{DBLP:journals/corr/abs-1905-00546}                 & ResNet-18           & \SWSLColor{Semi-weakly sup.}                \\
    \textsc{swsl\_resnext101\_32x16d}~ \cite{DBLP:journals/corr/abs-1905-00546}       & ResNext-101           & \SWSLColor{Semi-weakly sup.}                & 
    \textsc{CLIP}~\cite{radford2021learning}                            & Clip          & \ClipColor{Supervised CLIP}                 \\ 
    \textsc{dinov2\_giant}~\cite{oquab2023dinov2}                            & ViT          & \ViTColor{Self-supervised ViT}          & & &         \\ 
    \hline
    \end{tabular}
    }
    \caption{\small Summary of the models included in our study.}
    \label{tab:models}
    \vspace{-0.25in}
\end{table}

\begin{figure}[t!]
\centering
\captionsetup{justification=centering}
\scalebox{0.8}{
    \begin{tabular}{c c c c c c}
        %\toprule
         %& & & & \\
    \parbox[t]{2mm}{\multirow{-7}{*}{\rotatebox[origin=c]{90}{noise}}} 
     &
     \begin{subfigure}[b]{0.20\textwidth}
         \centering
         \includegraphics[width=\textwidth]{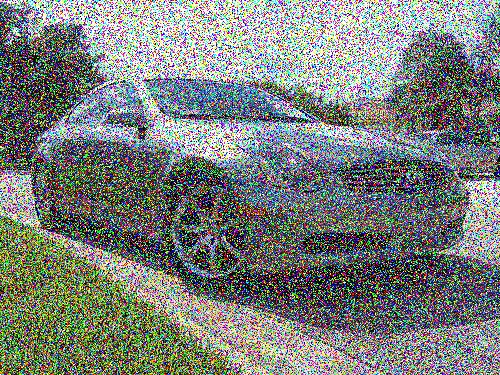}
         \caption{\scriptsize Impulse Noise \\(\imagenetc)}
         \label{fig:n1}
     \end{subfigure}
     &
     \begin{subfigure}[b]{0.20\textwidth}
         \centering
         \includegraphics[width=\textwidth]{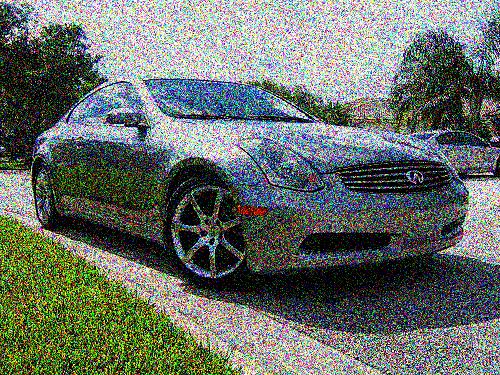}
         \caption{\scriptsize Shot Noise \\(\imagenetc)}
         \label{fig:n2}
     \end{subfigure}
     &
     \begin{subfigure}[b]{0.20\textwidth}
         \centering
         \includegraphics[width=\textwidth]{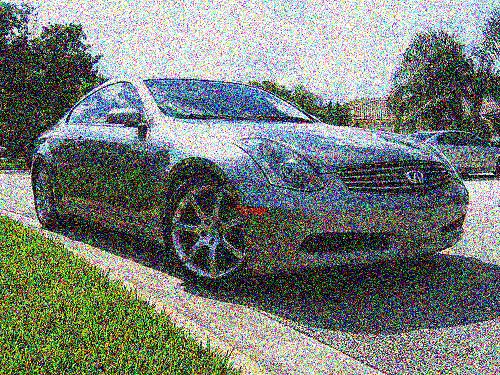}
         \caption{\scriptsize Gaussian Noise \\(\imagenetc)}
         \label{fig:n3}
     \end{subfigure}
     &
     \begin{subfigure}[b]{0.20\textwidth}
         \centering
         \includegraphics[width=\textwidth]{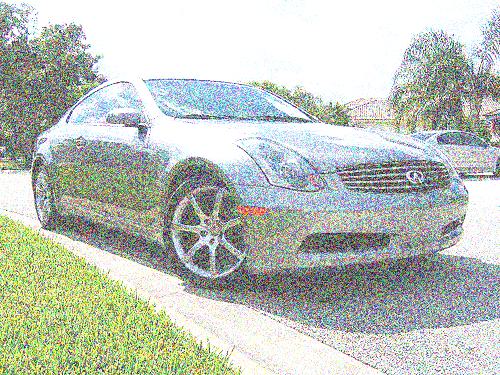}
         \caption{\scriptsize Uniform Noise \\(\cite{Geirhos2018GeneralisationIH})}
         \label{fig:n4}
     \end{subfigure}
     &
     \\
     \hline
     \parbox[t]{2mm}{\multirow{-7}{*}{\rotatebox[origin=c]{90}{blur}}} 
     &
     \begin{subfigure}[b]{0.20\textwidth}
         \centering
         \includegraphics[width=\textwidth]{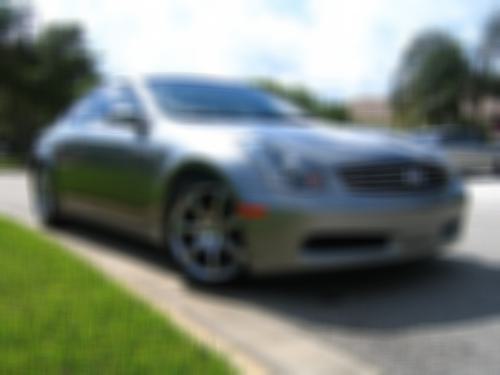}
         \caption{\scriptsize Blur \\(Albumentation)}
         \label{fig:b1}
     \end{subfigure}
     &
     \begin{subfigure}[b]{0.20\textwidth}
         \centering
         \includegraphics[width=\textwidth]{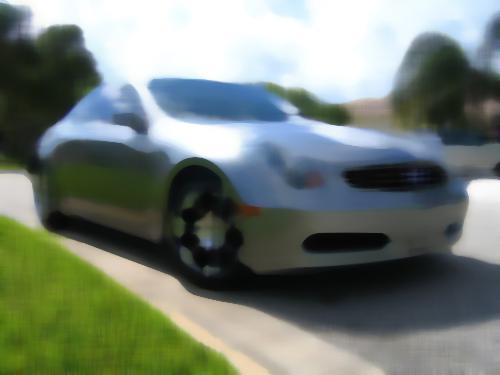}
         \caption{\scriptsize Median Blur \\(Albumentation)}
         \label{fig:b2}
     \end{subfigure}
     &
     \begin{subfigure}[b]{0.20\textwidth}
         \centering
         \includegraphics[width=\textwidth]{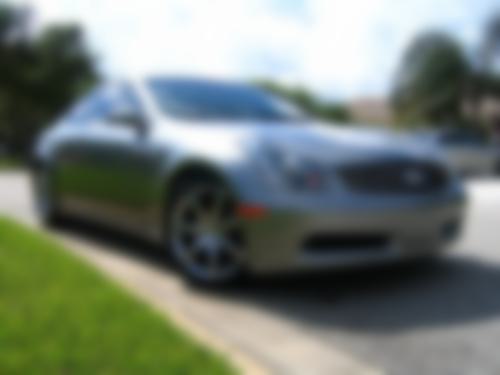}
         \caption{\scriptsize Glass Blur \\(\imagenetc)}
         \label{fig:b3}
     \end{subfigure}
     &
     \begin{subfigure}[b]{0.20\textwidth}
         \centering
         \includegraphics[width=\textwidth]{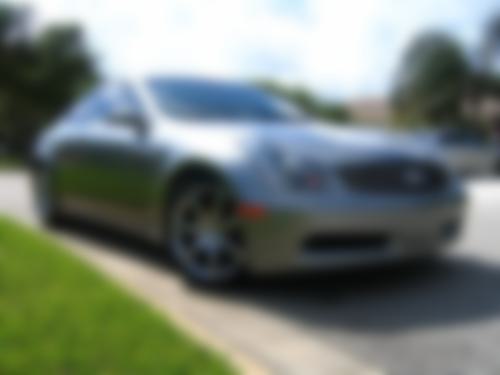}
         \caption{\scriptsize Gaussian Blur \\(\imagenetc)}
         \label{fig:b4}
     \end{subfigure}
     &
     \begin{subfigure}[b]{0.20\textwidth}
         \centering
         \includegraphics[width=\textwidth]{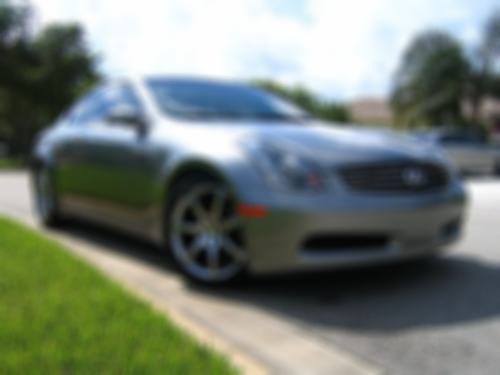}
         \caption{\scriptsize Defocus Blur \\(\imagenetc)}
         \label{fig:b5}
     \end{subfigure}\\
     \hline
     \parbox[t]{2mm}{\multirow{-7}{*}{\rotatebox[origin=c]{90}{others}}} 
     &
     \begin{subfigure}[b]{0.20\textwidth}
         \centering
         \includegraphics[width=\textwidth]{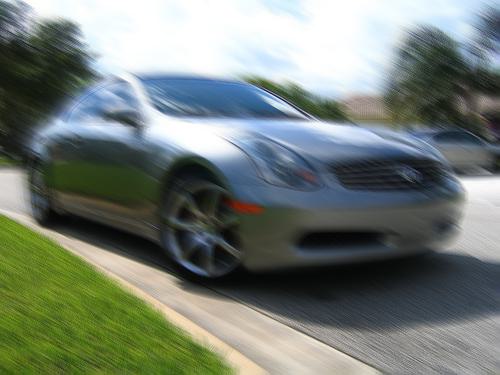}
         \caption{\scriptsize Motion Blur \\(\imagenetc)}
         \label{fig:o1}
     \end{subfigure}
     &
     \begin{subfigure}[b]{0.20\textwidth}
         \centering
         \includegraphics[width=\textwidth]{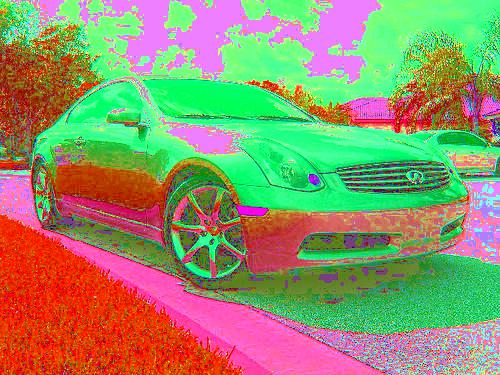}
         \caption{\scriptsize Hue Saturation Value \\(Albumentation)}
         \label{fig:o2}
     \end{subfigure}
     &
     \begin{subfigure}[b]{0.20\textwidth}
         \centering
         \includegraphics[width=\textwidth]{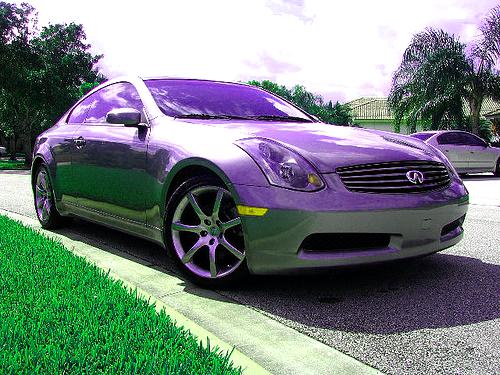}
         \caption{\scriptsize Color Jitter \\(Albumentation)}
         \label{fig:o3}
     \end{subfigure}
     &
     \begin{subfigure}[b]{0.20\textwidth}
         \centering
         \includegraphics[width=\textwidth]{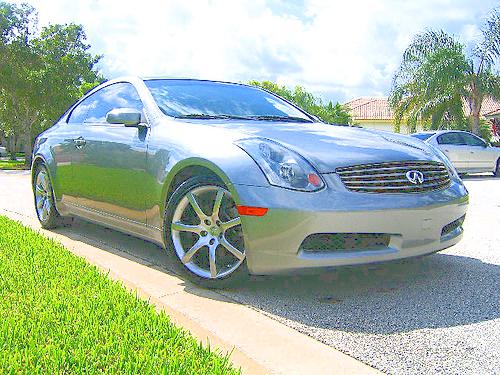}
         \caption{\scriptsize Brightness \\(\imagenetc)}
         \label{fig:o4}
     \end{subfigure}
     &
     \begin{subfigure}[b]{0.20\textwidth}
         \centering
         \includegraphics[width=\textwidth]{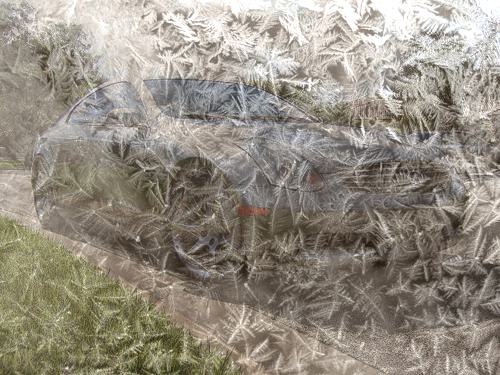}
         \caption{\scriptsize Frost \\(\imagenetc)}
         \label{fig:o5}
     \end{subfigure}\\
     \hline
    \end{tabular}
}
\vspace{-0.1in}
\caption{\small Image corruption functions. 
%discussed in the paper grouped by classes of similar corruption functions.
}
%The images are from Imagenet~\cite{ILSVRC2012} and the transformation is from Imagenet-c~\cite{hendrycks2019robustness}. }\
\label{fig:transf_example}
\vspace{-0.3in}
\end{figure}

\vskip 0.1in
\noindent
\textbf{Image Corruptions.}
%\subsubsection{Image Corruptions.}
As shown in Fig.~\ref{fig:transf_example}, we focus on studying \robustness{} of NNs in relation to humans regarding 14 commonly used image corruptions from three different sources: Shot Noise, Impulse Noise, Gaussian Noise, Glass Blur, Gaussian Blur, Defocus Blur, Motion Blur, Brightness and Frost from \imagenetc~\cite{hendrycks2019robustness}; Blur, Median Blur, Hue Saturation Value and Color Jitter from Albumentations~\cite{albumentation}; and Uniform Noise from~\cite{Geirhos2018GeneralisationIH}.%See the appendix for a visualization of these corruptions.

\vskip 0.1in
\noindent
\textbf{Crowdsourcing.} %description of the experiments. 
%\subsubsection{Crowdsourcing.}
%+ why crowdsourcing? because we want average case.
Given that VCR is focused on the average-case performance, we chose to use crowdsourcing for measuring human performance. This allowed us to involve a large number of participants for a more precise estimation of the average-case human performance.
The experiment is designed following \cite{hu-et-al-22} and~\cite{Geirhos2018GeneralisationIH}. 
The experiment procedure is a \emph{forced-choice image categorization task}: humans are presented with one image at a time, for 200 ms to limit the influence of recurrent processing, and asked to choose a correct category out of 16 entry-level class labels~\cite{Geirhos2018GeneralisationIH}. For NN models, 
%in order to obtain a choice from the same 16 categories,
the 1,000-class decision vector was mapped to the same 16 classes using the WordNet hierarchy~\cite{Geirhos2018GeneralisationIH}.
%The ILSRVR'12 dataset~\cite{ILSVRC2012} has 1,000 fine-grained classes; however, humans, when asked to label, are only able to categorize objects into entry-level classes (e.g., a dog rather than a German shepherd)~\cite{Geirhos2018GeneralisationIH}. Therefore, we provide humans with the 16 entry-level class labels mapped by Geirhos et al.~\cite{Geirhos2018GeneralisationIH}: airplane, bicycle, boat, car, chair, dog, keyboard, oven, bear, bird, bottle, cat, clock, elephant, knife, truck. 
The time to classify each image was set to ensure fairness in the comparison between humans and machines~\cite{Firestone26562}. Between images, we showed a noise mask to minimize feedback influence in the brain~\cite{Geirhos2018GeneralisationIH}. We included qualification tests and sanity checks aimed to filter out cases of participants misunderstanding the task and spammers~\cite{Papadopoulos-et-al-2017}, and only considered results from those participants who passed both tests. As a result, we had $7{,}718$ participants and obtained approximately (1) $70{,}000$ human predictions on images with different levels of visual corruptions; and (2) $50{,}000$ human predictions on original images as these can be repeated in experiments for different corruptions. The same original image, corrupted or not, was never shown again to the same participant. %See more detail in the supplementary materials.

\subsection{Experiment 1: Testing Robustness against Visual Corruption}
\label{sec:range}

%\red{This section will discuss the differences between checking with a few selected parameters and our full range. NOTE: this could be a short section merged with motivation of VCR}

%\red{experiments will be presented for:
%\begin{enumerate}
    %\item number of tests needed (does not increase cost)
    %\item Coverage difference  (can cover more levels of corruption) 
    %\item Difference in robustness evaluation: accuracy VS VCR values (show they can %result in biased results, we are independent of the distribution)
%\end{enumerate}
%stress that we are independent of sampling distribution while others are not. Distribution can affect evaluation results. Chose uniform sampling for the most fair representation of different corruption levels, but can be updated as desired.
%}

%\purple{figures needed:
%1. coverage table;
%2. figure 1 of partial, 1 for imagenet c accuracies, 1 for $R_a$, 1 for $R_p$ (mark colors and ones that are quite different)}

\imagenetc{} is the SoTA benchmark for corruption robustness. Rather than considering the continuous range of corruption like VCR, \imagenetc{} includes all \textsc{ImageNet} validation images corrupted using 5 pre-selected parameter values for each type of corruption~\cite{hendrycks2019robustness}. 
This section compares robustness measured with \imagenetc{} vs. VCR on all 9 \imagenetc{} corruption functions in our study. Due to the page limit, we include full results in the appendix.
%corruption functions that are shared between \imagenetc{} and our study. 
%{\bf MC:  I cannot parse this}
%We check \robustness{} of these 12 DNN models  against all 9 \imagenetc{} transformations with our generated benchmark validation set and our human-aware metrics \textit{HMRI} and \textit{MRSI}. We then compare results with robust accuracy of the same models on \imagenetc{} images.
%and then compare the results with their corruption robustness evaluated with \imagenetc. 
%Since human \robustness{} is measured on 16 entry-level classes,
%\textit{HMRI} and \textit{MRSI} are calculated by mapping DNN labels to the same 16 classes for a fair human-DNN comparison.
%For our human-aware metrics \textit{HMRI} and \textit{MRSI}, for a more fair comparison between human and DNNs, we mapped DNN predictions into the 16 entry-level classes human on which \robustness{} is measured. 
 %Mapping to 16 classes would result in higher accuracy and metric values than considering all 1000 classes.  {\bf MC:  not enough info here} 
%Additionally, \imagenetc{} ranks DNNs with accuracy {\bf MC:  what does it mean?} on all transformed images for all transformations, while we consider each transformation separately.  
%Results are shown in Tbl.~\ref{tab:eval_results}. We present results for two transformations in this paper; the rest are in the supplementary. %For each transformation, we used test sequences of size $50,000$. 
%\red{TODO: Calculate CO2 Emission}\blue{\bf{[Discussed in supplementary]}}.

\begin{figure}[t!]
\centering
%\vspace{-0.2in}
\begin{minipage}{.45\textwidth}
\centering
  \scalebox{0.9}{
    \begin{tabular}{c c}
        %\toprule
         %& & & & \\
         
     \begin{subfigure}[b]{0.5\linewidth}
         \centering
         \includegraphics[width=\textwidth]{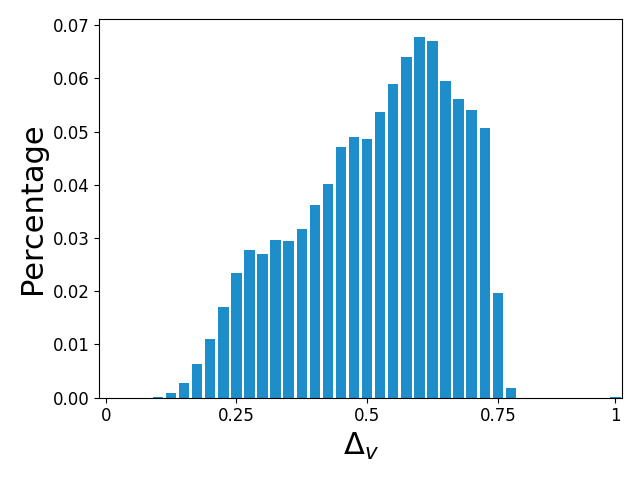}
         \caption{\scriptsize \imagenetc.}
         \label{fig:gauss_blur_c}
     \end{subfigure}
     &
     \begin{subfigure}[b]{0.5\linewidth}
         \centering
         \includegraphics[width=\textwidth]{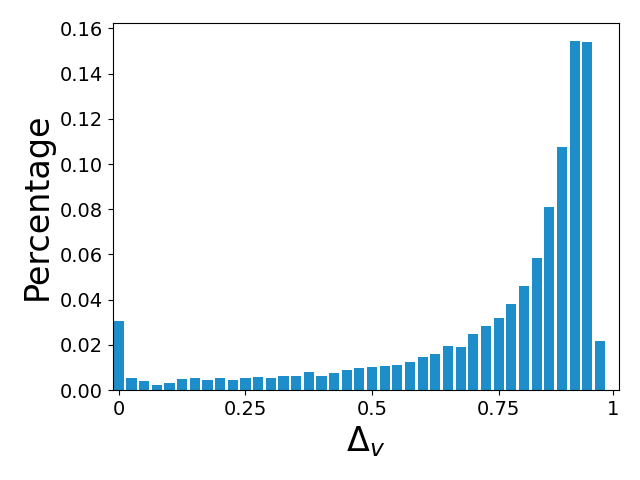}
         \caption{\scriptsize VCR Test Set.}
         \label{fig:gauss_blur_v}
     \end{subfigure}
    \end{tabular}
}
\vspace{-0.1in}
\caption{%Comparison of $\Delta_v$ distribution between \imagenetc{} and our \method test sets for Gaussian Blur. The figures are histograms, where x-axis is $\Delta_v$, y-axis is image count. 
\small Histograms showing $\Delta_v$ distribution between \imagenetc{} and our VCR test sets for Gaussian Blur. }
%The images are from Imagenet~\cite{ILSVRC2012} and the transformation is from Imagenet-c~\cite{hendrycks2019robustness}. }\
\label{fig:gauss_blur}
\vspace{-0.1in}
\end{minipage}%
\hspace{0.3cm}
\begin{minipage}{.45\textwidth}
  \centering
  \scalebox{0.6}{
    \begin{tabular}{lcc}
    \toprule
    \multirow{2}{*}{Corruption} & \multicolumn{2}{c}{Coverage} \\
    & \small \imagenetc{} & \small VCR Test Set  \\
    
    % &  ImageNetC Coverage &  Coverage \\
    \midrule
    Brightness &               0.590 &         1.000  \\
   
     Gaussian Blur &               0.564 &         0.974  \\
      Defocus Blur &               0.538 &         0.923 \\
             Shot Noise &               0.462 &         0.590 \\
             Frost &               0.436 &         1.000 \\
Gaussian Noise &               0.436 &         0.872 \\
     Impulse Noise &               0.385 &         0.641 \\
    
      Motion Blur  &               0.333 &         0.974 \\
        Glass Blur &               0.333 &         0.949 \\
    \bottomrule

%    {'gaussian_blur_imagenet16': 0.9743589743589743,
% 'shot_noise_imagenet16': 0.5897435897435898,
% 'glass_blur_imagenet16': 0.9487179487179487,
% 'brightness_imagenet16': 1.0,
% 'color_jitter_imagenet16': 1.0,
% 'frost_imagenet16': 1.0,
% 'impulse_noise_imagenet16': 0.6410256410256411,
% 'blur_wrap_imagenet16': 0.8974358974358975,
% 'gaussian_noise_imagenet16': 0.8717948717948718,
% 'hue_saturation_imagenet16': 0.9230769230769231,
% 'median_blur_imagenet16': 0.8717948717948718,
% 'defocus_blur_imagenet16': 0.9230769230769231,
% 'motion_blur_c_imagenet16': 0.9743589743589743,
% 'uniform_noise_imagenet16': 0.8717948717948718}
    \end{tabular}
    }
    \vspace{-0.1in}
    \captionof{table}{ \small $\Delta_v$ Coverage Comparison with \imagenetc{}.}
    \vspace{-0.1in}
    \label{tab:coverage_table}
\end{minipage}

\vspace{-0.1in}

\end{figure}

\vskip 0.1in
\noindent
\textbf{Visual Corruption in Test Sets.}
%\subsubsection{Visual Corruption in Test Sets.}
For each corruption, our tests generated for checking \robustness{} contain $50,000$ images, mirroring the size of the \textsc{ImageNet}~\cite{ILSVRC2012} validation set, while \imagenetc{} includes $5\times 50,000$ images. Due to the difference in how test sets are generated, we observe two major differences in the distributions of degrees of visual corruption: %(measured by $\Delta_v$ for each corruption type as shown in Fig.~\ref{fig:gauss_blur}): %,
%First we can see that 
%Because of the difference in parameter values used in \imagenetc{} and our benchmark, 
they have different coverage and peak at different values (e.g., Fig.~\ref{fig:gauss_blur}). 

% Because of the difference in the parameter values used, the $\Delta_v$ distributions between \imagenetc{} and our benchmark peak at different values. For example, in Fig.~\ref{fig:brightness_c} and Fig.~\ref{fig:brightness_v}, for Brightness  most \imagenetc{} images have $\Delta_v$ values between $0.4$ to $0.8$, while most \method{} images are between $0.6$ and $0.9$; a similar observation holds for Defocus Blur and Gaussian Blur. 

%The histogram plots may not discernibly show the actual coverage of the $\Delta_v$ ranges with low frequencies. To address this shortcoming,  
To quantitatively assess the actual coverage of $\Delta_v$ in the test sets, 
Tab.~\ref{tab:coverage_table} gives the coverage as a percentage of the full $\Delta_v$ range of $[0,1]$. To compute it, the distribution is divided into 40 bins with the same width. A bin is considered covered if it contains 20 or more images. The coverage is then determined by dividing the number of covered bins by the total number of bins (40).
We observed that \imagenetc{} exhibits a low coverage of $\Delta_v$ values. Specifically, as shown in Fig.~\ref{fig:gauss_blur} and Tab.~\ref{tab:coverage_table}, the distribution of \imagenetc{} in Gaussian blur has coverage of only  56.4\% focusing mainly on the center of the entire domain of $\Delta_v$ and missing coverage for low and high $\Delta_v$ values, \black{which can lead to biased evaluation. As we show in the appendix, the same can be observed for most \imagenetc{} corruption functions.}
On the other hand, our test sets provide coverage for almost the entire domain, with a coverage percentage of 97.4\%. This pattern holds true for other corruption functions as well---our test sets have consistently higher coverage than \imagenetc. 
% {\bf MC:  why is this important? \blue{addressed}}
%
%
% than all transformations shared with \imagenetc{}.
%
As for VCR, Shot Noise and Impulse Noise have relatively low coverage, because the level of noise these functions add is exponential to their parameters. As a result, uniform sampling of the parameter range $C$ fails to cover small $\Delta_v$ values.
%(TODO: I think we should also find another example where big $\Delta_v$ is not covered). 
%When using uniform sampling over $C$, reaching the full coverage of $\Delta_v$ would require a large amount of data. Note, however, that VCR is still computed over the full $\Delta_v$ range of $[0..1]$, and the lack of samples for low values of $\Delta_v$ has a limited impact on the VCR estimate. This is because we fit a monotonic spline over the entire range of $\Delta_v$ that is anchored with a known initial performance for $\Delta_v=0$, and this performance is preserved until the lower bound of the covered $\Delta_v$ range. A possible approach to obtain a more uniform sample of $\Delta_v$ would be to (1) fit a strictly monotonic spline into $(c,\Delta_v)$ values obtained from $c\sim \textit{Uniform}(C)$, (2) take $\Delta_v\sim \textit{Uniform}(0,1)$, and (3) map the latter to a new sample from $C$ using the inverted spline. This approach could be repeated iteratively to obtain a more uniform sample of $\Delta_v$. However, that would be computationally expensive while providing diminishing returns in terms of the VCR estimates.  
When using uniform sampling over $C$, reaching the full coverage of $\Delta_v$ would require a large amount of data. Note, however, Alg.~\ref{alg:vcr-est} still computes VCR over the full $\Delta_v$ range of $[0..1]$, and the lack of samples for low values of $\Delta_v$ has a limited impact on the VCR estimate. This is because we fit a monotonic spline that is anchored with a known initial performance for $\Delta_v=0$, as discussed in Appendix.~\ref{sec:app_cov}. %and this performance is assumed until the lower bound of the covered $\Delta_v$ range.

%Next, 
%We notice that \imagenetc{} images cannot cover all $\Delta_v$ values. Specifically, Fig.~\ref{fig:defocus_c} for Defocus Blur shows that \imagenetc{} validation set does not contain images with $\Delta_v$ greater than $0.8$ and less than $0.2$. The same can be observed for all transformations shown in Fig.~\ref{fig:coverage}. 
\noindent
\underline{Remark:}
%{\bf MC:  END.  Next sentence does not parse}
The reported accuracy of \imagenetc{} can be directly impacted both by a lack of coverage and by non-uniformity, as it is computed as the average accuracy of all corrupted images. In contrast, the shape of the $\Delta_v$ distribution in the test images does not impact VCR once sufficient coverage is achieved to estimate the spline curves $s_{\gamma}$.%, as already explained. 
%; i.e., non-uniform distribution of $\Delta_v$ in the test set would not affect \robustness{}. This is because the requirement for an accurate estimation of $\robustnesssymbol_{\gamma}$ is a sufficient number of points collected to estimate the spline curves $s_{\gamma}$, and at each point, as long as the test data is sufficient to measure performance, the distribution of $\Delta_v$ in the test data would not affect $\robustnesssymbol_{\gamma}$.

%When considering the full range of visual changes that a transformation can incur, the $\Delta_v$ distribution can directly impact the reported accuracy of \imagenetc{}, since the accuracy is computed as the average accuracy of all transformed images, therefore both lacking coverage and non-uniformity can lead to biased results.

%using \imagenetc{} can lead to biased results due to biased distribution and insufficient coverage.

%our metrics \textit{HMRI} and \textit{MRSI}.
%However, for \imagenetc{}, the $\Delta_v$ distribution directly impacts the reported accuracy, since the accuracy is computed as the average accuracy of all transformed images, therefore both lacking coverage and non-uniformity can lead to biased results.

\begin{figure}[t!]
    \centering
    %\vspace{-0.1in}
    \begin{subfigure}{0.36\textwidth}
        \includegraphics[width=0.85\textwidth]{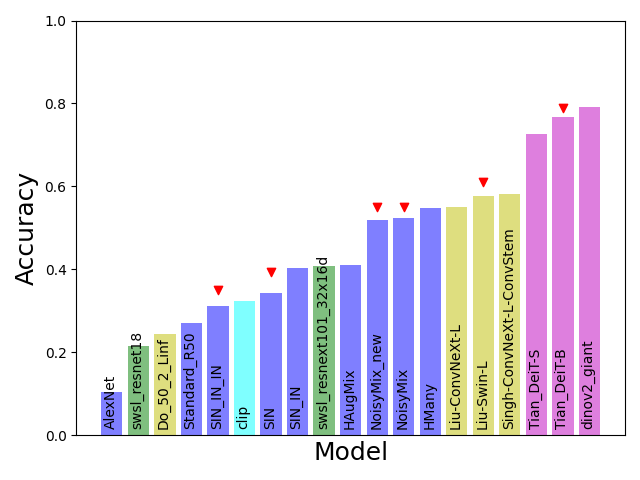}
        \caption{\scriptsize \imagenetc{} Gaussian Noise Accuracy.}
        \label{fig:imagenetc-gaussian-noise-acc-bar}
    \end{subfigure}
    \begin{subfigure}{0.30\textwidth}
        \includegraphics[width=\textwidth]{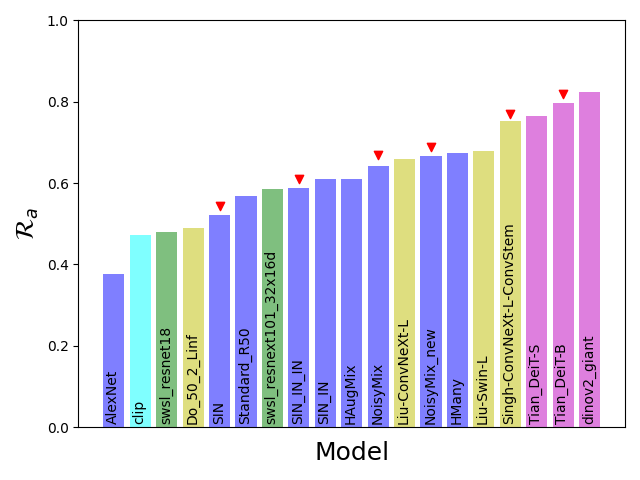}
        \caption{\scriptsize Gaussian Noise $\hat{\mathcal{R}}_a$.}
        \label{fig:vcr-gaussian-noise-acc-bar_in}
    \end{subfigure}
    \begin{subfigure}{0.30\textwidth}
        \includegraphics[width=\textwidth]{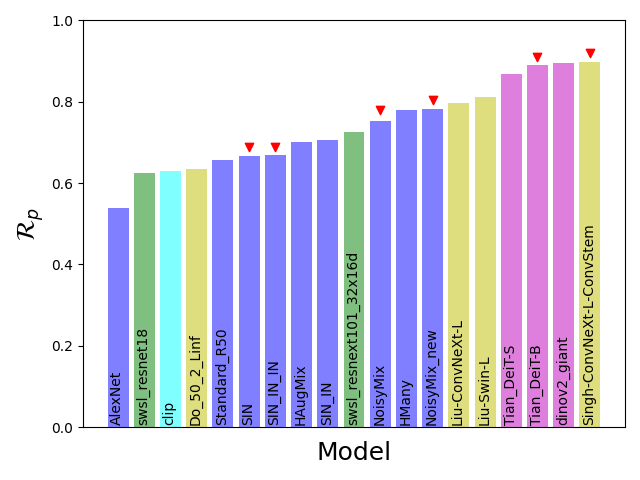}
        \caption{\scriptsize Gaussian Noise $\hat{\mathcal{R}}_p$.}
        \label{fig:vcr-gaussian-noise-pred-sim-bar_in}
    \end{subfigure}
    \vspace{-0.1in}
    \caption{Comparison between \textsc{ImageNet-C} and VCR with Gaussian Noise. Models discussed in the text are marked by a red triangle.}
    \label{fig:compare-imagenetc-vcr}

    \vspace{-0.2in}
\end{figure}

\vskip 0.1in
\noindent
\textbf{Robustness Evaluation Results.}
%\subsubsection{Robustness Evaluation Results.}
Next, we compare robustness evaluation results obtained with \imagenetc{} and \robustness{} test sets. Consider results for Gaussian Noise in Fig.~\ref{fig:compare-imagenetc-vcr}. %\purple{figure needed: figure 1 of partial, 1 for imagenet c accuracies, 1 for $\mathcal{R}_a$, 1 for $\mathcal{R}_p$ (mark colors and ones that are quite different} 
% As highlighted in red in Tbl.~\ref{tab:eval_results}, 
\textsc{NoisyMix} and \textsc{NoisyMix\_new} have
almost the same robust accuracy on \imagenetc{},  but \textsc{NoisyMix\_new} has higher 
%has the highest \imagenetc{} robustness accuracy, but one of the lowest 
$\hat{\robustnesssymbol}_a$; similarly, SIN has higher \imagenetc{} robust accuracy but lower $\hat{\robustnesssymbol}_a$ than SIN\_IN\_IN.
% Shot noise is removed here because this will require more space to put another row of figures
% If we consider transformation Shot Noise, \imagenetc{} ranks Do\_50\_2\_Linf~\cite{SalmanIEKM20} second to last,  but its $\hat{\robustnesssymbol}^a$ is around the median. 
%AugMix~\cite{hendrycks2020augmix} and NoisyMix\_new~\cite{NoisyMix} have the same \imagenetc{} robust accuracy but considering a wider range of visual change, AugMix~\cite{hendrycks2020augmix} is more robust with a higher $\hat{\robustnesssymbol}^a$ value.
%AugMix~\cite{hendrycks2020augmix} is ranked in the middle with \imagenetc{} robustness accuracy but it has the highest $\hat{\robustnesssymbol}^a$ among all the DNNs. 
This is due to the almost complete lack of coverage for $\Delta_v<0.5$ for Gaussian Noise in {\imagenetc} (see Tab.~\ref{tab:coverage_table}), which can lead to biased evaluation results (i.e., biased towards $\Delta_v\geq 0.5$).
%This is because checking robustness on only a portion of the full range of changes, as in \imagenetc, can lead to biased results. 
Checking VCR allows us to detect such biases. 

% Second, in addition to accuracy, our benchmark can also be used to check whether the DNN can preserve its predictions after transformations, i.e., its prediction similarity, which can give us additional information about DNN robustness. 
% From Fig.~\ref{fig:vcr-gaussian-noise-pred-sim-bar} and Fig.~\ref{fig:vcr-gaussian-noise-acc-bar} we can see that \purple{\textbf{Singh2023}} has a lower $\hat{\mathcal{R}}^a$ than \purple{Tian\_DeiT-B} but a higher $\hat{\mathcal{R}}^p$. This suggests that event though

%Secondly, 
In addition to accuracy, VCR can also %be used to 
check whether the NN can preserve its predictions after corruption, i.e., the prediction consistency property $p$, giving additional information about NN robustness. 
% As highlighted in green in Tbl.~\ref{tab:eval_results} for the transformation Shot Noise, the model 
%NoisyMix\_new~\cite{NoisyMix} has a higher $\hat{\robustnesssymbol}^a$ than NoisyMix~\cite{NoisyMix} but a lower $\hat{\robustnesssymbol}^p$; similarity,
From Figs.~\ref{fig:vcr-gaussian-noise-acc-bar_in},~\ref{fig:vcr-gaussian-noise-pred-sim-bar_in}, we can see that the model \textsc{Tian\_Deit\_B} has a higher $\hat{\robustnesssymbol}_a$ than \textsc{Singh-ConvNeXt-L-ConvStem} but a lower $\hat{\robustnesssymbol}_p$. This suggests that even though \textsc{Tian\_Deit\_B} has better accuracy for corrupted images, it labels the same image with different labels before and after the corruption. Since ground truth can be hard to obtain during deployment, having low prediction consistency indicates issues with model stability and could raise concerns about when to trust the model prediction. Results for the remaining corruptions are in Appendix.~\ref{sec:app_extra}.

\noindent
\underline{Summary:} It is essential to test robustness before deploying NNs into an environment with a wide and continuous range of visual corruptions. Our results confirmed that \textbf{testing robustness in this range using a fixed and pre-selected number of parameter values can lead to undetected robustness issues}, which can be avoided by checking VCR. Additionally, \textbf{accuracy cannot accurately represent model stability when facing corruptions}, which can be addressed by testing $\robustnesssymbol_p$. 

\subsection{Experiment 2: VCR of DNNs Compared with Humans}
%\red{TODO: add little arrow to indicate model name in plots}
\label{sec:humans}

In this experiment, we use our new human-aware metrics, \textit{HMRI} and \textit{MRSI}, and the data from the human experiment data to compare VCR of the studied models against human performance.% as a baseline.
%We briefly show the results for Gaussian Noise and Shot Noise; the results for the remaining corruptions are in the supplementary.  
%In this paper, we also introduced human-aware metrics for comparing model \robustness{} with humans. Since humans are naturally robust to a wide range of changes, using human performance as a baseline can give information about the trustworthiness of NNs. 

\begin{figure}[bt!]
%\vspace{-0.1in}
    \centering
    \begin{tabular}{p{0.32\textwidth} p{0.32\textwidth} p{0.32\textwidth}}
         \begin{subfigure}{0.32\textwidth}
        \includegraphics[width=\textwidth]{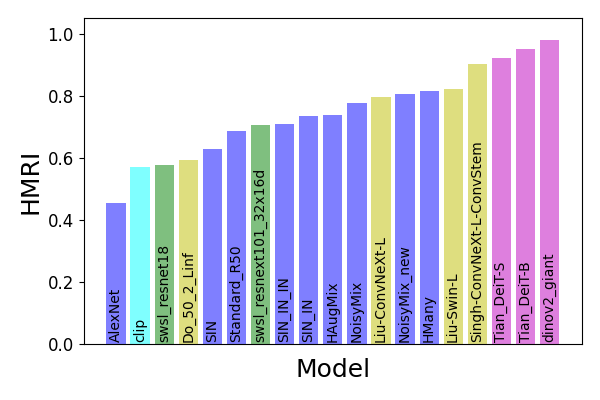}
        \caption{\scriptsize \textit{HMRI} for $\robustnesssymbol_a$ }
        \label{fig:gaussian-noise-HMRI-acc}
    \end{subfigure}& 
    \begin{subfigure}{0.32\textwidth}
        \includegraphics[width=\textwidth]{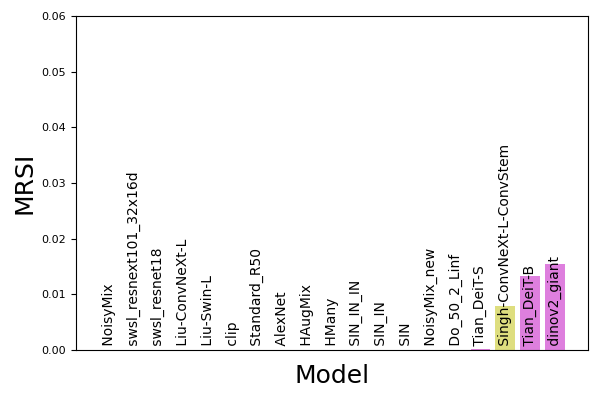}
        \caption{\scriptsize \textit{MRSI} for $\robustnesssymbol_a$}
        \label{fig:gaussian-noise-MRSI-acc}
    \end{subfigure}&
    \begin{subfigure}{0.32\textwidth}
        \includegraphics[width=\textwidth]{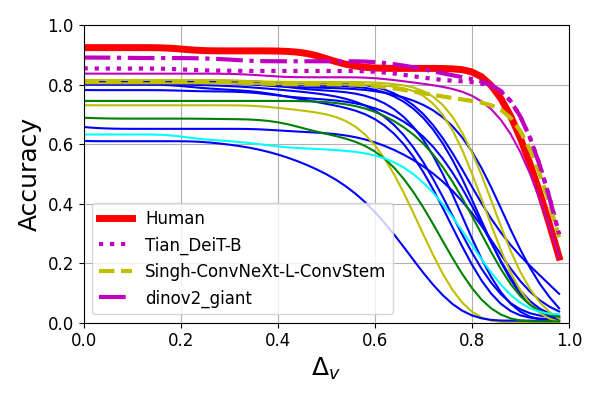}
        \caption{Estimated curves $s_a$}
        \label{fig:gaussian-noise-vcr-acc-comparison}    
    \end{subfigure}
    \\
          \begin{subfigure}{0.32\textwidth}
        \includegraphics[width=\textwidth]{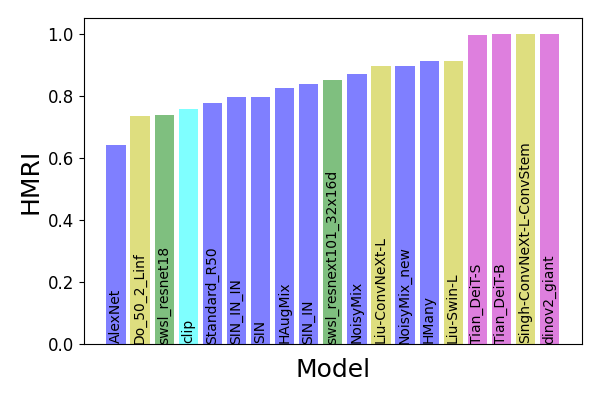}
        \caption{\scriptsize \textit{HMRI} for $\robustnesssymbol_p$}
        \label{fig:gaussian-noise-HMRI-pred}
    \end{subfigure}&
    \begin{subfigure}{0.32\textwidth}
        \includegraphics[width=\textwidth]{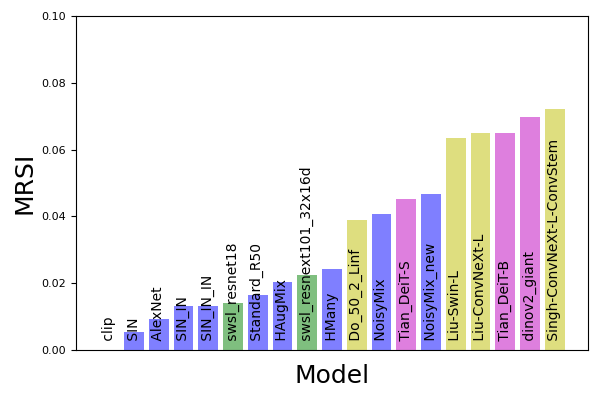}
        \caption{\scriptsize \textit{MRSI} for $\robustnesssymbol_p$}
        \label{fig:gaussian-noise-MRSI-pred}
    \end{subfigure}
    &\begin{subfigure}{0.32\textwidth}
        \includegraphics[width=\textwidth]{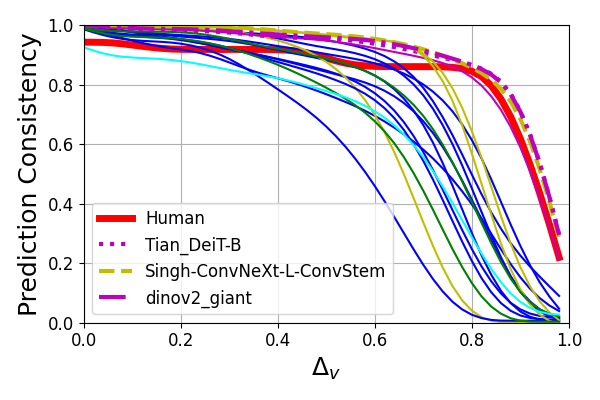}
        \caption{Estimated curves $s_p$}
        \label{fig:gaussian-noise-vcr-pred-sim-comparison} 
    \end{subfigure}

    \end{tabular}

    \vspace{-0.25in}
    
    \caption{\small VCR evaluation results for Gaussian Noise. Results include, for each NN, the estimated curves $s_a$ and $s_p$ (representing how the performance w.r.t. the robustness properties $a$ and $p$ decreases as $\Delta_v$ increases); and the corresponding \textit{HMRI} and \textit{MRSI} values. 
    %including the estimated curves $s_a$ and $s_p$ for all NN models, and their HMRI and MRSI metrics values.
    %VCR of humans and NNs with our metrics HMRI and MRSI 
    %Our measured HMRI and MRSI values for Gaussian Noise.
    Results are colored  based on their category:  \HumanColor{Human}, \ViTColor{Vision Transformer}, \SupervisedColor{Supervised Learning}, \SWSLColor{SWSL}, \AdversarialColor{Adversarial Training}, \ClipColor{CLIP}.
    }
    \label{fig:hmri-mrsi-bar-plot_gn}

    %\vspace{0.05in}

    \begin{tabular}{p{0.32\textwidth} p{0.32\textwidth} p{0.32\textwidth}}
    
      \begin{subfigure}{0.32\textwidth}
        \includegraphics[width=\textwidth]{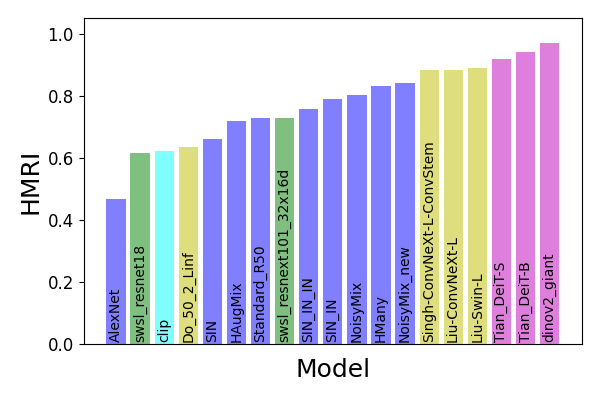}
        \caption{\scriptsize \textit{HMRI} for $\robustnesssymbol_a$ }
        \label{fig:uniform-noise-HMRI-acc}
    \end{subfigure}&
    \begin{subfigure}{0.32\textwidth}
        \includegraphics[width=\textwidth]{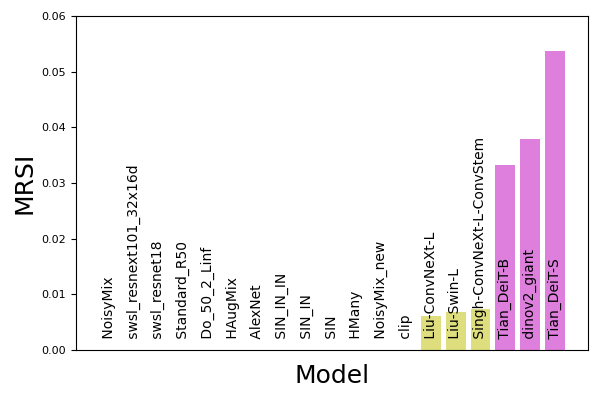}
        \caption{\scriptsize \textit{MRSI} for $\robustnesssymbol_a$}
        \label{fig:uniform-noise-MRSI-acc}
    \end{subfigure} &
    \begin{subfigure}{0.32\textwidth}
        \includegraphics[width=\textwidth]{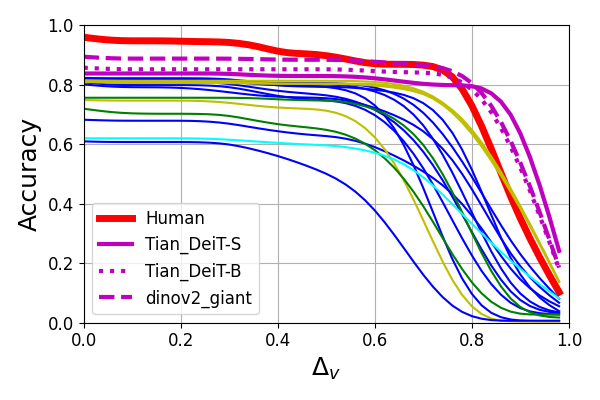}
        \caption{Estimated curves $s_a$ }
        \label{fig:uniform-noise-vcr-acc-comparison}    
    \end{subfigure}
    \\
   \begin{subfigure}{0.32\textwidth}
        \includegraphics[width=\textwidth]{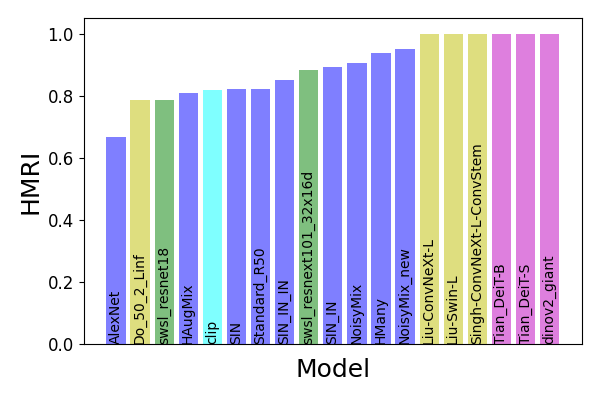}
        \caption{\scriptsize \textit{HMRI} for $\robustnesssymbol_p$}
        \label{fig:uniform-noise-HMRI-pred}
    \end{subfigure} &
    \begin{subfigure}{0.32\textwidth}
        \includegraphics[width=\textwidth]{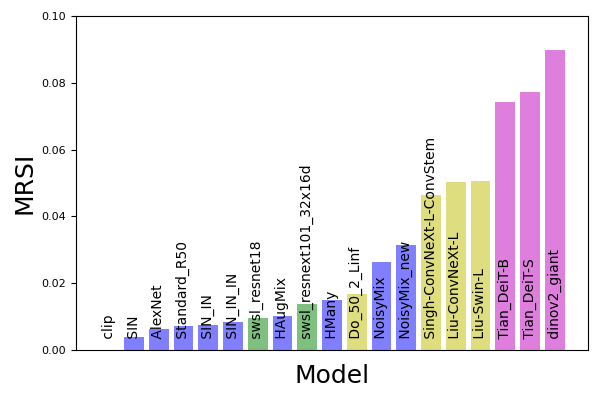}
        \caption{\scriptsize \textit{MRSI} for $\robustnesssymbol_p$}
        \label{fig:uniform-noise-MRSI-pred}
    \end{subfigure}&
\begin{subfigure}{0.32\textwidth}
        \includegraphics[width=\textwidth]{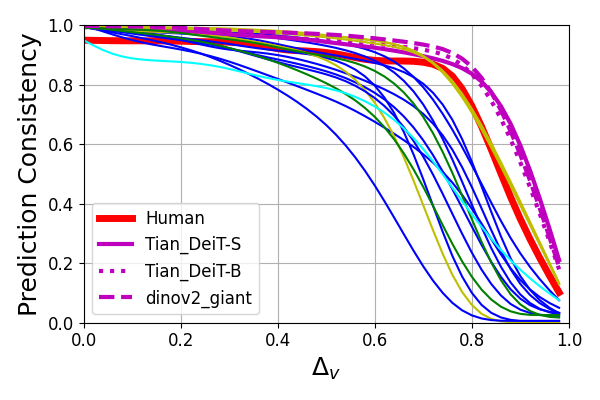}
        \caption{Estimated curves $s_p$ }
        \label{fig:uniform-noise-vcr-pred-sim-comparison}    
    \end{subfigure}

    \end{tabular}

    \vspace{-0.25in}
    
    \caption{\small  VCR evaluation results for Uniform Noise.} 
    \label{fig:hmri-mrsi-bar-plot_un}

    \vspace{-0.25in}
\end{figure}

For Gaussian Noise, Fig.~\ref{fig:hmri-mrsi-bar-plot_gn} presents our measured \textit{HMRI} and \textit{MRSI} values for $\robustnesssymbol_a$ and $\robustnesssymbol_p$. 
For both metrics, a higher value indicates better robustness. 
%As shown in Fig.~\ref{fig:gaussian-noise-HMRI-acc} and Fig.~\ref{fig:gaussian-noise-HMRI-pred}, no NN has reached $1.0$ for \textit{HMRI}, indicating that there are still unclosed gaps between human and NN robustness, with humans giving more accurate and more consistent predictions facing corruptions. The NNs \textsc{Tian\_DeiT-B}, \textsc{Tian\_DeiT-S}, and \textsc{Singh-ConvNeXt-L-ConvStem} have the highest \textit{HMRI} values (around $0.9$), making these models closest to human robustness. 
\black{As shown in Fig.~\ref{fig:gaussian-noise-HMRI-acc}, no NN has reached $1.0$ for $\textit{HMRI}_a$, and in Fig.~\ref{fig:gaussian-noise-HMRI-pred}, only 3 out of 21 NNs \textsc{dinov2\_giant}, \textsc{Tian\_DeiT-B} and \textsc{Singh-ConvNeXt-L-ConvStem} reached $1.0$ for $\textit{HMRI}_p$, indicating that there are still unclosed gaps between human and NN robustness, with humans giving more accurate and more consistent predictions facing corruptions than most SoTA NNs. These thee top-performing models have also the highest \textit{HMRI} values for both $\robustnesssymbol_a$ and $\robustnesssymbol_p$, making these models closest to human robustness.}
In Fig.~\ref{fig:gaussian-noise-MRSI-acc}, we can see that these three models have $\textit{MRSI}_a$ values above $0.0$, indicating that they surpass human accuracy in certain ranges of visual corruption. This can be visualized by checking the estimated curves $s_a$ as shown in Fig.~\ref{fig:gaussian-noise-vcr-acc-comparison}. 
The top-three models exceed human accuracy (the red curve) when $\Delta_v > 0.85$. For prediction consistency, Fig.~\ref{fig:gaussian-noise-MRSI-pred} shows that all NNs have the $\textit{MRSI}_p$ value above $0.0$ and this is because, as shown in Fig.~\ref{fig:gaussian-noise-vcr-pred-sim-comparison}, all NN curves are above the human curve when the $\Delta_v$ value is small. Specifically, the top-three models completely exceed humans in the entire $\Delta_v$ range.

%Additionally, for Shot Noise, in Fig.~\ref{fig:shot-noise-HMRI-acc} and Fig.~\ref{fig:shot-noise-MRSI-acc}, we can see that regarding $\robustnesssymbol_a$, the top performing model for \textit{MRSI} (\textsc{Singh-ConvNeXt-L-ConvStem}) leads the second place (\textsc{Tian\_DeiT-B}) by a large margin, but its \textit{HMRI} is lower than \textsc{Tian\_DeiT-B}. This suggests that \textsc{Singh-ConvNeXt-L-ConvStem} exceeds human performance more than \textsc{Tian\_DeiT-B}, but \textsc{Tian\_DeiT-B} is closer to human performance than \textsc{Singh-ConvNeXt-L-ConvStem}, which may be counter-intuitive. This can be explained with $s_a$ shown in Fig.~\ref{fig:shot-noise-vcr-acc-comparison}. The accuracy of \textsc{Singh-ConvNeXt-L-ConvStem} exceeds human performance by a lot for $\Delta_v$ between $[0.05 .. 0.4]$, then it decreases rapidly after $0.4$ to be lower than \textsc{Tian\_DeiT-B}. \textsc{Tian\_DeiT-B}, on the other hand, has accuracy close to human accuracy over a larger range of $\Delta_v$, between $[0.0 .. 0.8]$. Therefore, \textsc{Tian\_DeiT-B} has closer to human accuracy, i.e., higher \textit{HMRI}, but \textsc{Singh-ConvNeXt-L-ConvStem} can exceed humans more with small amounts of visual corruption, i.e., higher \textit{MRSI}. This suggests that both \textit{HMRI} and \textit{MRSI} are useful for comparing NN robustness, and our curves $s_a$ and $s_p$ can provide further information on NN robustness with different degrees of visual corruption.

Similarly, for Uniform Noise, as shown in Fig.~\ref{fig:uniform-noise-HMRI-acc} and Fig.~\ref{fig:uniform-noise-HMRI-pred}, no models reached $1.0$ for $\textit{HMRI}_a$ and the top-three models, reached $1.0$ for $\textit{HMRI}_p$. Together with Fig.~\ref{fig:uniform-noise-MRSI-acc} and Fig.~\ref{fig:uniform-noise-MRSI-pred}, we can see that for both $\robustnesssymbol_a$ and $\robustnesssymbol_p$, \textsc{Tian\_DeiT-B} has higher \textit{HMRI} values but \textsc{Tian\_DeiT-S} has higher \textit{MRSI} values. This suggests that while \textsc{Tian\_DeiT-B} is closer to human performance, \textsc{Tian\_DeiT-S} exceeds human performance more. This result may be counter-intuitive but can be explained with the curves $s_a$ and $s_p$ representing how the performance w.r.t. the robustness properties $a$ and $p$ decreases as $\Delta_v$ increases, as shown in Fig.~\ref{fig:uniform-noise-vcr-acc-comparison} and Fig.~\ref{fig:uniform-noise-vcr-pred-sim-comparison}. From both $s_a$ and $s_p$, we observed that for $\Delta_v$ values less than $0.8$, the performance of \textsc{Tian\_DeiT-B} is higher than \textsc{Tian\_DeiT-S} and closer to human, hence the higher \textit{HMRI} value; and after $\Delta_v = 0.8$ when human performance starts decreasing, the \textsc{Tian\_DeiT-B} performance drops rapidly to much below that of \textsc{Tian\_DeiT-S}, hence the lower \textit{MRSI} value. 
%Additionally, observe that for \textsc{Tian\_DeiT-S}, both $s_a$ and $s_p$ are flatter compared to other models. This indicates that its performance remains relatively consistent even in the presence of Uniform Noise, i.e., robust against Uniform Noise. 

This suggests that both \textit{HMRI} and \textit{MRSI} are useful for comparing NN robustness, and our curves $s_a$ and $s_p$ can provide further information on NN robustness with different degrees of visual corruption.

Overall, in both Fig.~\ref{fig:hmri-mrsi-bar-plot_gn} and Fig.~\ref{fig:hmri-mrsi-bar-plot_un}, we observed that the three ViT models (shown in purple) have the best performance for both $\mathcal{R}_a$ and $\mathcal{R}_p$, making them the models closest to human robustness. The same can also be observed for the rest of the corruption functions; see the appendix for more details. This indicates that vision transformer is the most promising architecture for reaching human-level robustness, even outperforming models trained with additional training data. The data in the appendix also indicates the biggest remaining robustness gap for blur corruptions. 
Furthermore, our generated test sets can be used during model retraining for improved robustness compared to humans, resulting in with higher \textit{HMRI} and \textit{MRSI} values.
%the results are coloured based on their characteristic like training method or architecture. 
%One pattern we can observe is that the two ViT models (purple bars) seem to perform the best in terms of both $\mathcal{R}_a$ and $\mathcal{R}_p$. Not only for gaussian noise, this pattern was observed in all transformations. The rest of the results can be found in supplementary

\noindent
\underline{Summary:} As our results suggest, when considering the full range of visually-continuous corruption, 
%no NNs can match human robustness for both accuracy and prediction consistency, and few exceed humans with only a very tiny margin for some specific degrees of corruption. 
\textbf{no NNs can match human accuracy, especially for blur corruptions, and only the best-performing ones can match human prediction consistency. For some specific degrees of corruption, few NNs can exceed humans by mostly tiny margins. }
This highlights a more substantial gap between human and NN robustness than previously identified by \cite{GeirhosNMTBWB21}. 
By evaluating VCR using our human-centric metrics, we gain deeper insights into the robustness gap, which can aid in the development of models closer to human robustness.

\subsection{Experiment 3: Training with Data Augmentation}
\label{sec:train}
%Retraining was conducted by training on all parameters of the model. Training set was sampled from the training set of ImageNet~\cite{ILSVRC2012} with a size of around 12,000. %Optimizer used was stochastic gradient descent with learning rate=0.001 and momentum=0.9. Loss function used was cross entropy loss. 5 epochs is enough to show some progress.
%

Because VCR considers a different distribution of corruptions in the images (i.e., continuous) than existing benchmarks (i.e., selected parameter values), model performance is expected to improve once the model is fine-tuned on the new distribution.
We show a small retraining example to demonstrate the usefulness of our benchmark in improving VCR.
The retraining process was carried out by fine-tuning all parameters of the image classification model. The training dataset was generated from a subset sampled from the \textsc{ImageNet}~\cite{ILSVRC2012} training set with a size of around 12,000. For optimization, we leveraged the most basic stochastic gradient descent with learning rate=0.001 and momentum=0.9. We utilized Cross-Entropy Loss as the loss function, given its effectiveness in classification tasks. The number of epochs depends on the model. Five epochs are usually enough to show some progress. %The training details can also be found in the codebase: \gitlink.
The state-of-the-art NNs are already optimized for the corruption functions included in \imagenetc; however, as shown in Tab.~\ref{tab:coverage_table}, for certain corruption functions, such as Motion Blur, Frost and Glass Blur, \imagenetc{} images do not cover a wide range of visual changes, leaving room for robustness improvement. %
In Tab.~\ref{tab:retrain_result}, we demonstrate results for NNs SIN~\cite{GeirhosRMBWB19} and Standard\_R50~\cite{croce2020robustbench} for these corruption functions, the rest can be found in the codebase$^{\ref{git}}$. 

\noindent
\underline{Summary:} Our results show that
simply retraining with tests generated with VCR can improve
all metrics comparing NN model performance relative to humans, even for models already optimized for the same corruption functions included in \imagenetc. This is because VCR considers a completely different distribution of corruption that the models were not previously exposed to. This shows that the \textbf{gap between human and NN robustness is larger than benchmarks with discrete corruptions such as \imagenetc{} can detect}. Our proposed \textbf{VCR can not only detect this gap, it also provides a step towards closing this gap!}

\begin{table*}[t!]
    \centering
%\vspace{-0.2in}
\noindent\scalebox{0.52}{
    \begin{tabular}{ |>{\columncolor{gray!20}}c| c c c c c c >{\columncolor{gray!10}}c >{\columncolor{gray!10}}c >{\columncolor{gray!10}}c >{\columncolor{gray!10}}c >{\columncolor{gray!10}}c >{\columncolor{gray!10}}c | c c c c c c >{\columncolor{gray!10}}c >{\columncolor{gray!10}}c >{\columncolor{gray!10}}c >{\columncolor{gray!10}}c >{\columncolor{gray!10}}c >{\columncolor{gray!10}}c  |}
    \hline
     & \multicolumn{12}{|c|}{Results for Standard\_R50~\cite{croce2020robustbench}} & \multicolumn{12}{c|}{Results for SIN~\cite{GeirhosRMBWB19}}\\
    
    \cline{2-25}
         & \multicolumn{6}{c}{Before Retraining}& \multicolumn{6}{>{\columncolor{gray!10}}c|}{After Retraining} & \multicolumn{6}{c}{Before Retraining}& \multicolumn{6}{>{\columncolor{gray!10}}c|}{After Retraining} \\
        \cline{2-25}
           &  \multicolumn{3}{c}{Accuracy} & \multicolumn{3}{c}{Prediction similarity}  & \multicolumn{3}{>{\columncolor{gray!10}}c}{Accuracy} & \multicolumn{3}{>{\columncolor{gray!10}}c|}{Prediction similarity}&  \multicolumn{3}{c}{Accuracy} & \multicolumn{3}{c}{Prediction similarity}  & \multicolumn{3}{>{\columncolor{gray!10}}c}{Accuracy} & \multicolumn{3}{>{\columncolor{gray!10}}c|}{Prediction similarity}\\
        \cline{2-25}
        \multirow{-4}{*}{corruption function} & $\hat{\robustnesssymbol}_a$& \textit{HMRI} & \textit{MRSI} & $\hat{\robustnesssymbol}_p$& \textit{HMRI} & \textit{MRSI} & $\hat{\robustnesssymbol}_a$& \textit{HMRI} & \textit{MRSI} & $\hat{\robustnesssymbol}_p$& \textit{HMRI} & \textit{MRSI} & $\hat{\robustnesssymbol}_a$& \textit{HMRI} & \textit{MRSI} & $\hat{\robustnesssymbol}_p$& \textit{HMRI} & \textit{MRSI} & $\hat{\robustnesssymbol}_a$& \textit{HMRI} & \textit{MRSI} & $\hat{\robustnesssymbol}_p$& \textit{HMRI} & \textit{MRSI}\\
        \hline
        % Standard R50 Model
        Median Blur &  0.532 &     0.635 &     0.000 &  0.573 &          0.673 &          0.000 & \redb{0.694} &     \redb{0.828} &     \redb{0.003} &  \redb{0.728} &          \redb{0.854} &          \redb{0.001}  & 
        0.522 & 0.624 & 0.00 & 0.605 & 0.710 & 0.00 & \redb{0.650} & \redb{0.774} & \redb{0.004} & \redb{0.729} & \redb{0.852} & \redb{0.004} \\
        \hline
        Frost & 0.429 & 0.521 & 0.011 & 0.473 & 0.572 & 0.012 & \redb{0.575} & \redb{0.690} & \redb{0.025} & \redb{0.678} & \redb{0.804} & \redb{0.031} & 
        0.423 & 0.512 & 0.015 & 0.513 & 0.618 & 0.016 & \redb{0.517} & \redb{0.625} & \redb{0.016} & \redb{0.647} & \redb{0.768} & \redb{0.031} \\
        \hline
        % Glass Blur
        Glass Blur & 0.468 & 0.569 & 0.003 & 0.502 & 0.603 & 0.003 & \redb{0.647} & \redb{0.770} & \redb{0.024} & \redb{0.744} &\redb{0.866} & \redb{0.034}&  
        0.334 & 0.407 & 0.000 & 0.397 & 0.478 & 0.000 & \redb{0.572} & \redb{0.687} & \redb{0.016} & \redb{0.684} & \redb{0.809} & \redb{0.018}  \\
               \hline 
               \multicolumn{12}{l}{Note: all numbers are rounded.}\\
        % \hline

        %
    \end{tabular}
}
    \caption{VCR comparison before and after retraining. Red indicates improvement.}
    \label{tab:retrain_result}
    \vspace{-0.25in}
\end{table*}

\subsection{Experiment 4: VCR for Visually Similar Corruption Functions}
\label{sec:similar}

%\red{During the process we discovered visually similar transformations, which we then discovered can be identified through a simple method, perfecting this is future work. (we can present this more as a discovery with initial results rather than a solution)}

%\red{experiments to be presented:
%\begin{enumerate}
%    \item transferability, with more analysis, e.g., what's the error bar for identifying ones not really similar, stress future opportunities.
%    \item retraining with one and check performance improvements on similar, for discussion of the difference between humans and DNN. 
%\end{enumerate}
%We need to stress the significance of this: understanding expectations between human and ML, save experiment cost.}

\begin{figure}[t!]
%\vspace{-0.1in}
    \centering
    \begin{subfigure}[b]{0.32\linewidth}
         \centering
         \includegraphics[trim={0 0 0 1cm },clip,width=\linewidth]{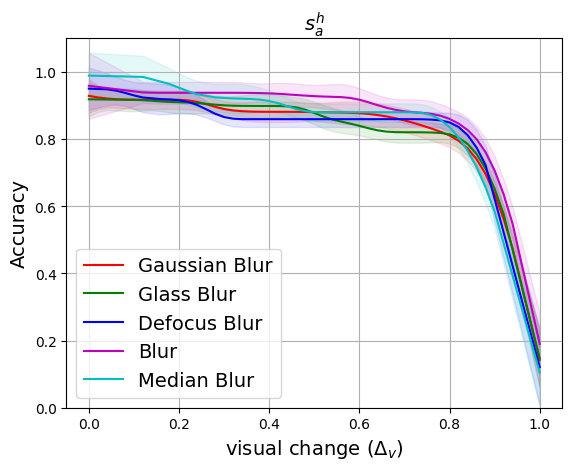}
         \vspace{-0.2in}
         \caption{ \small Blur class $s_a^h$}
         \label{fig:d}
     \end{subfigure}
     \begin{subfigure}[b]{0.32\linewidth}
         \centering
         \includegraphics[trim={0 0 0 1cm },clip,width=\linewidth]{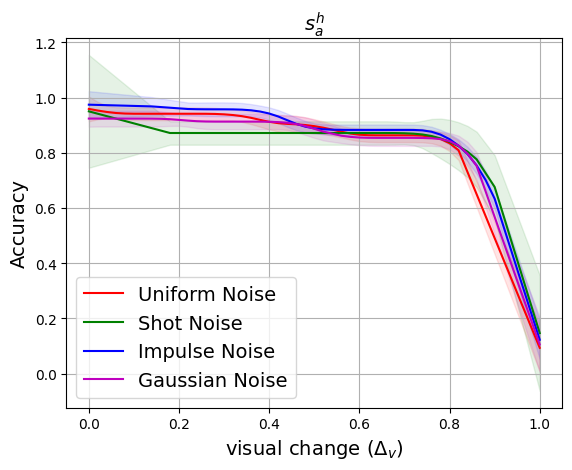}
         \vspace{-0.2in}
         \caption{\small Noise class $s_a^h$}
         \label{fig:d}
     \end{subfigure}
     \begin{subfigure}[b]{0.32\linewidth}
         \centering
         \includegraphics[trim={0 0 0 1cm },clip,width=\linewidth]{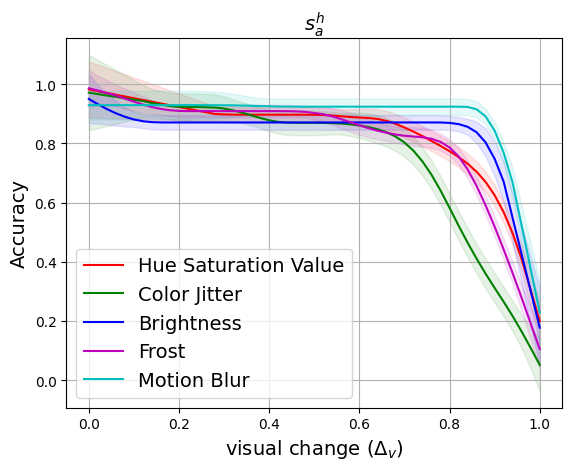}
         \vspace{-0.2in}
         \caption{\small Dissimilar $s_a^h$}
         \label{fig:d}
     \end{subfigure}
\vspace{-0.1in}
    \caption{\small Comparing human performance spline curves $s^h_a$ for similar and dissimilar corruption functions. For each curve, the coloured region around the curve is the $83\%$ confidence interval used for comparison of similarity. See $s^h_p$ in Appendix.~\ref{sec:app_extra}. }
    \label{fig:similar}
    
    \vspace{-0.25in}
\end{figure}

%\red{update story}
%\red{TODO: find a model, plot VCR curve clustered by similar transformation (noise/blur). Find a model that's very different}
One noteworthy observation we made from our experiments with humans is the existence of \emph{visually similar} corruption functions. This can contribute towards reducing experiment costs and a better understanding of differences between humans and NNs.

Different corruptions change different aspects of the images, e.g., image colour, contrast, and the amount of additive visual noise, and thus affect human perception differently~\cite{Geirhos2018GeneralisationIH}. Also, multiple different corruption functions can be implemented for the same visual effect, such as Gaussian noise and Impulse noise for noise addition. Although the difference between Gaussian noise and Impulse noise can be picked up by complex NN models, an average human %without high computational power 
would struggle to distinguish between the two. Therefore, for a specific visual effect, there should exist a class of corruption functions implementing the effect that an average human is unable to tell them apart.
We call corruption functions in the same class \emph{visually similar}. We postulate that since visually similar functions, by definition, affect human perception similarly, they would affect human robustness similarly as well. Therefore, human data for one function can be reused for other similar functions in the same class possibly reducing experiment costs.

Since \robustness{} is estimated with the spline curves $s_a^h$ and $s_p^h$, if the difference among the curves of a set of functions is statistically insignificant, human data (i.e., the spline curves) can be reused among the functions in this set. In Fig.~\ref{fig:similar}, we plot the smoothed spline curves $s_a^h$ and $s_p^h$ obtained for all 14 corruption functions included in our experiments.
We can observe that, for all corruption functions shown, human performance decreases slowly for small values of visual degrade ($\Delta_v$), but once $\Delta_v$ reaches a turning point, human performance starts decreasing more rapidly. Then, we observe that spline curves obtained for certain blur and noise transformations have similar shapes, while those for dissimilar transformations start decreasing at different turning points with different slopes. %decrease differently.  
More specifically, the differences between two spline curves are statistically insignificant if their $83\%$ confidence intervals overlap~\cite{koenker-94}.
%, hu-et-al-22}. 

\noindent
\underline{Summary:} By checking statistical significance with $83\%$ confidence interval for each corruption function, we empirically observed two classes of visually similar corruptions in our experiments with humans: (1) the noise class: Shot Noise, Impulse Noise, Gaussian Noise, and Uniform Noise; and (2) the blur class: Blur, Median Blur, Gaussian Blur, Glass blur, Defocus Blur.
The remaining corruptions are dissimilar (see Fig.~\ref{fig:similar}).
%: Motion Blur, Hue Saturation Value, Color Jitter, Brightness, and Frost. See supplementary for the full demonstration. }

\vskip 0.1in
\noindent
\textbf{NN Robustness for Visually Similar Corruption Functions.} 
%
%\subsubsection{\red{TODO} NN Robustness for Visually Similar Transformations. } 
Because of the central difference between humans and NNs, e.g., computational powers, it is intuitive that NNs might react completely differently to corruptions visually similar to humans, and using VCR, we can empirically analyze such difference. For example, during deployment, noise with unknown distributions (ranging from Uniform, Gaussian, Poisson etc.), can be encountered.  While noise distribution does not affect humans as we showed in Fig.~\ref{fig:similar}, NNs which are particularly susceptible to a certain distribution might raise safety concerns. 
For example, two visually similar transformations Gaussian Noise and Uniform Noise add an additional noise to the images with the Gaussian and the Uniform distribution, respectively. 
%From Fig.~\ref{fig:similar}, we can see that Gaussian Noise and Shot Noise affect human robustness similarly, i.e., the difference in noise distribution is too subtle for humans to pick up. 
However, our results in Fig.~\ref{fig:hmri-mrsi-bar-plot_gn} and Fig.~\ref{fig:hmri-mrsi-bar-plot_un} suggest that the distribution difference is picked up by NNs. 
\black{We can observe that most models have higher \textit{HMRI} and \textit{MRSI} values for Uniform Noise than Gaussian Noise. For small amounts of corruption ($\Delta_v < 0.8$), the difference between the estimated $s_a$ and $s_p$ curves for both corruptions is not statistically significant, i.e.,  NN models perform similarly when facing small amounts of Uniform and Gaussian Noise. 
For $\Delta_v$ values between $[0.8 .. 1.0]$, most visual information required for humans to recognize objects is corrupted by the noise, human performance decreases quickly, but the most robust models, e.g., \textsc{dinov2\_giant} and \textsc{Tian\_DeiT-S}, are able to pick up more information than humans and make reasonable recognition. When the added noise is from a uniform distribution, NN models perform better than when it is from a Gaussian distribution.}
%More specifically, for $\robustnesssymbol_a$, from Fig.~\ref{fig:gaussian-noise-vcr-acc-comparison} and Fig.~\ref{fig:gaussian-noise-vcr-pred-sim-comparison}, we can see that for Gaussian Noise, best-performing NNs (e.g., \textsc{Singh2023} and \textsc{Tian\_DeiT-B}) exceed human robustness with a higher degree of corruption ($\Delta_v > 0.8$), while for Shot Noise, they outperform humans with a lower degree of corruption ($\Delta_v < 0.4$). Additionally, most supervised and adversarially trained NNs have a lower accuracy overall for Shot Noise compared to Gaussian Noise. Similar results can be observed for $\robustnesssymbol_p$: best-performing NNs are more robust against small amounts of Shot Noise and large amounts of Gaussian Noise. 
Therefore, studying VCR also allows us to empirically analyze how changing the noise distribution, which would not affect humans, affects NN performance for different degrees of corruption. In the case of unknown or shifting distributions, such analysis would require human data for all distributions which is impractical and expensive. Identifying classes of visually similar corruption functions and reusing human data would significantly reduce the experiment costs.

%are more sensitive to an exponential distribution of additional noise than humans but can be robust against small exponential noises and .
%In this case, performance comparison with humans would require human data for all distributions, which is impractical and expensive. Knowing classes of corruption functions visually similar to human perception can significantly reduce experiment costs by allowing reuse of data and also allows us to empirically evaluate the differences between humans and NNs. 

%they are visually similar for humans, 
%\purple{Pending new results for retraining with one and check performance improvements on similar, for discussion of the difference between humans and DNN.
%figures to include: noise, blur and others
%1. all human data plots to show similarity;
%2. evaluation results for similar ones (figure 2 of partial but same model but different transformation in one plot, different colors represent different groups);
%3. retraining results for similar ones (will keep tables);
%}

\vskip 0.1in
\noindent
\textbf{Identifying Visually Similar Transformations.}
%\subsubsection{Identifying Visually Similar Corruptions.} 
We provide a naive method for identifying classes of visually similar corruptions.
To identify whether two corruptions are similar enough to reuse human data, the goal is to determine whether the difference between them is distinguishable to a human. \black{This can be done through a set of relatively inexpensive experiments. Without knowing the specific corruptions introduced to the images, participants are shown corrupted images and asked if the presented images are corrupted with the same corruption function. }
%This can be done through a set of relatively inexpensive experiments where the participants, without any knowledge of which corruptions are being applied in the images, are shown corrupted images and asked whether the presented images are corrupted with the same corruption.
% {\bf MC:  UGH \blue{rewritten}} 
Presented images can be corrupted with the same or different corruption functions. Then, by repeating the experiments with different sampled images, the accuracy of distinguishing the corruptions can be calculated. 
\black{We hypothesize that if the corruption functions are indistinguishable, human accuracy should be close to random.}
%We hypothesize that if humans cannot distinguish the differences, the accuracy should be close to random. 
% {\bf MC:  UGH \blue{rewritten}} 
Then, since each experiment is either successfully distinguished or not, we use a binomial test to check whether the accuracy is statistically significant to not be random. Visually similar transformations included in this paper can be detected with this naive method.
%See more details in supplementary materials.
Our experiments showed that for each pair of transformations, results with statistical significance can be reached in less than a minute. Compared to the full set of experiments with $2{,}000$ images and five different participants for each experiment, identifying similar transformations significantly decreased the experiment time, from approximately $5.55$ hours to $5$ minutes. 

\noindent
\underline{Limitation:} Note that the results of this method can be highly dependent on the opinion of the participants; thus, it is more optimal to select participants with a normal eyesight and a basic knowledge of image corruptions. We acknowledge that this naive method cannot give the most accurate identification of visually similar transformations. For example, it is reasonable to assume that two transformations can have very different visual effects but still affect human robustness in the same way, and this case would not be detected with this method. Nevertheless, we hope that our findings will promote future investigations of how NNs and humans react differently to corruptions.

\section{Conclusion}
\label{sec:conclusion}
%\noindent - Summary\\
%    \indent\quad - What's done?\\
%    \indent\quad - What is the experiment result?\\
%    \indent\quad - Contribution\\
%    \indent\quad - Limitations and future work: type of applicable transformations; depends on availability of human data (+ applicability to other types of model, e.g., detection); transferability cannot completely replace human experiment; our metrics may not be suited to be used during training (requires lots of data, or more work needs to be done which is future work)?\\
%    \indent\quad - negative societal impacts: environmental impact, large computation power needed for generating validation images, but still much less compared to training large models.

In this paper, %to better understand NN robustness relative to humans, 
we revisit corruption robustness to consider
%introduce a new concept \emph{visual-corruption robustness} (VCR) that considers 
it in relation to the wide and continuous range of corruptions to human perceptive quality, defining \emph{visually-continuous corruption robustness} (VCR); along with two novel human-aware metrics for NN evaluation.
Our results showed that \textbf{the robustness gap between human and NNs is bigger than previously detected}, especially for blur corruptions. We found that using the full and continuous range of visual change is necessary when estimating robustness, as \textbf{insufficient coverage can lead to biased results}. We also discovered classes of image corruptions that affect human perception similarly and identifying them can \textbf{help reduce the cost of measuring human robustness} and assessing disparities between human perception and computational models. 
In our study, we only considered the comparison of object recognition between humans and NNs; however, human and machine vision can be compared in many different ways, e.g., against neural data~\cite{yamins2014performance, kubilius2019brain}, contrasting Gestalt effects~\cite{kim2019neural}, object similarity judgments~\cite{hebart2020revealing}, or mid-level properties~\cite{storrs2021unsupervised}. 
%In no way do we claim our results can represent these studies, but we beleive our results are 
Still, we hope our results will inspire future robustness studies.   We also provide our benchmark datasets with human performance data and our code as open source.

\bibliographystyle{splncs04}
\bibliography{main}
\newpage

%\section{Supplementary Material}
\appendix

\section{Implementation and Data}
\label{sup:link}
Data and implementation can be found at \gitlink.

\section{Overview of \method }

\begin{figure}[h]
    \centering
    \includegraphics[width=\linewidth]{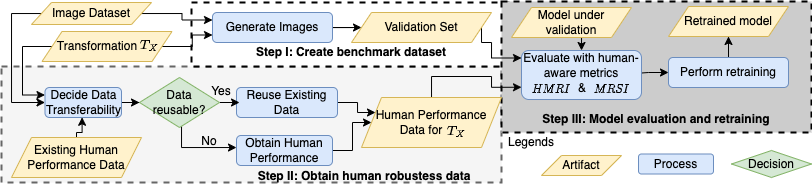}
    \caption{Our proposed method {\method} for benchmarking ML robustness with humans. }
    \label{fig:intro-paper-design-flowchart}

\end{figure}

Our method for benchmarking VCR ({\method}) is outlined in  Fig.~\ref{fig:intro-paper-design-flowchart}. Step~\RN{1} generates a validation set that covers the full continuous range of visual changes. This is achieved by uniformly sampling from the entire domain of corruption function parameters. Step~\RN{2} obtains human robustness performance data needed to compute our two newly-proposed human-aware evaluation metrics: \emph{Human-Relative Model Robustness Index} (HMRI) and \emph{Model Robustness Superiority Index} (MRSI), which quantify the extent to which a NN can replicate or surpasses human performance, respectively. Since measuring human performance for every single image corruption function is expensive and impractical, we propose a method to reduce the cost by generalizing existing human performance data obtained for one corruption function to a class of corruption functions with similar visual effects. For example, images transformed with Gaussian Blur and Glass Blur have very similar visual effects on humans, unlike Motion Blur and Brightness.  Thus, Gaussian Blur and Glass Blur, but not with Motion Blur and Brightness, thus they belong to the same class of similar corruption functions, and human performance data for one can be transferred to the other. Step~\RN{3} of \method{} evaluates the model using the validation dataset and our human-aware metrics. Then it retrains the model to improve its robustness.

\section{VCR and Its Estimation}
\label{sec:app_vcr}

\vskip 0.05in
\noindent
{\bf Background: Image Quality Assessment (IQA)}. 
%IQA metrics are quantitative measures of human objective image quality~\cite{SSIM}. Given the original image and the transformed image, the IQA metrics automatically predict the perceived image quality by measuring the perceptual `distance' between the two images~\cite{VIF}. This `distance' is different from pixel distance and its calculation depends on the design of the IQA metric. VIF (\emph{Visual Information Fidelity}~\cite{VIF}) measures the information fidelity by analyzing the statistics of the natural scenes in the images. VIF returns a value between 0 and 1 if the changes degrade perceived image quality, with 1 indicating the perfect quality compared to the original image; and it returns a value  $>1$ if the changes enhances image quality~\cite{VIF}. VIF is empirically shown to be the closest to human opinions when compared to all other IQA metrics~\cite{Sheikh2006}. VIF is applicable to transformations that can be described locally by a combination of signal attenuation and additive Gaussian noise in the sub-bands in the wavelet domain~\cite{VIF}. 
%
IQA metrics serve as quantitative measures of human objective image quality~\cite{SSIM}. By comparing the original image and the transformed image, IQA metrics automatically estimate the perceived image quality by evaluating the perceptual ``distance'' between the two images~\cite{VIF}. This ``distance'' differs from simple pixel distance and varies depending on the specific IQA metric used.

One such metric is VIF (Visual Information Fidelity)~\cite{VIF}, which evaluates the fidelity of information by analyzing the statistical properties of natural scenes within the images. VIF returns a value between 0 and 1 if the changes degrade perceived image quality, with 1 indicating the perfect quality compared to the original image; and it returns a value  $>1$ if the changes enhances image quality~\cite{VIF}. More precisely, VIF defines the visual quality of a distorted image as a ratio of the amount of information a human can extract from the distorted image versus the original reference image. The method models statistically (i) images in the wavelet domain with coefficients drawn from a Gaussian scale mixture, (ii) distortions as attenuation and additive Gaussian noise in the wavelet domain, and (iii) the human visual system (HVS) as additive white Gaussian noise in each sub-band of the wavelet decomposition. The amount of information that a human can extract from the distorted image is measured as the mutual information between the distorted image and the output of the HVS model for that image. Similarly, the amount of information that a human can extract from the reference image is measured as the mutual information between the reference image and the output of the HVS model for that image. Empirical studies have shown that VIF aligns closely with human opinions when compared to other IQA metrics~\cite{Sheikh2006}.

We choose VIF, since it is well-established, computationally efficient, applicable to our transformations, and still performing competitively compared to newer metrics. More recent research has explored the use of feature spaces computed by deep NNs as a basis to define IQA metrics (e.g., LPIPS~\cite{LPIPS} and DISTS~\cite{DISTS}). Even though these metrics may be applicable to a wider class of transformations than VIF, including those that affect both structure and textures, their scope may depend on the training datasets in potentially unpredictable ways. On the other hand, the scope of VIF is well-defined based on the metric's mathematical definition. In particular, VIF is suitable for evaluating corruption functions that can be locally described as a combination of signal attenuation and additive Gaussian noise in the sub-bands of the wavelet domain~\cite{VIF}. The transformations in our experiments are local corruptions that are well within this scope.  Moreover, VIF performs still competitively when compared to even the newer DNN-based metrics across multiple datasets (e.g., see Table 1 in~\cite{DISTS}). However, future work should explore VCR using other IQA metrics.

\vskip 0.05in
\noindent
{\bf Visual Change ($\Delta_v$)}. The metric $(\Delta_v)$ defined using the IQA metric VIF, as shown in Def.~\ref{def:visual_degrade}, is proposed by Hu et al.~\cite{hu-et-al-22} to quantitatively measure the amount of visual changes in the images perceived by human observers. % that is independent of particular images and transformations. %Using $(\Delta_v)$ while studying human robustness removes the complexity of different transformation parameter domains and effects; and also how transformations may impact individual images differently.                  
%
%$\Delta_v$ is defined with a Image Quality Assessment (IQA) metric, which given the original image, automatically predict the perceived image quality by measuring the perceptual ‘distance’ between the original and transformed images~\cite{hu-et-al-22}. 
%Specifically, $\Delta_v$ uses an IQA metric VIF~\cite{VIF} for considering imperceptible and visible changes respectively~\cite{hu-et-al-22}. 

\begin{definition}
Let an image $x$, an applicable corruption function $T_X$ with a parameter domain $C$ and a parameter $c\in C$, s.t. $x' = T_X(x,c)$ be given.
Visual change $\Delta_v(x,x')$ is a function defined as follows:
    {\small  \[ \begin{cases}
      0  & $If VIF$(x,x')>1\\
       1-$VIF$(x, x')& $Otherwise  $
   \end{cases}
\]}
\label{def:visual_degrade}
\end{definition}

$\Delta_v$ returns a value between 0 and 1, with 0 indicating no degradation to visual quality and 1 indicating all visual information in the original image has been changed. The first case of $\Delta_v$ corresponds to changes that enhance the visual quality (when VIF$(x,x')>1$), indicating changes do not impact human recognition of the images negatively, hence $\Delta_v = 0$. The other case deals with visible changes that degrade visual quality. Since VIF returns 1 for perfect quality compared to the original image, the degradation is one minus the image quality score. 

\begin{figure}[h]
\centering
\scalebox{0.9}{
    \begin{tabular}{c c c c}
        %\toprule
         %& & & & \\
         \begin{subfigure}{.16\textwidth}
          \includegraphics[width=\linewidth]{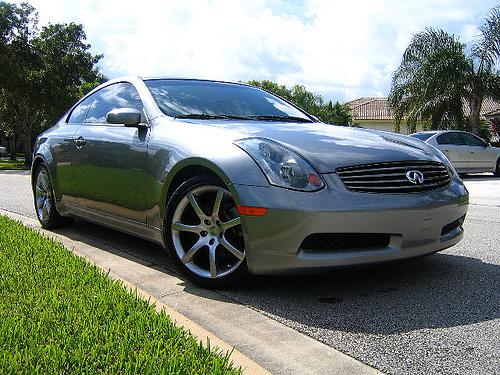}\caption{original image, $\Delta_v=0$}\label{fig:a_sup}
         \end{subfigure} &
         \begin{subfigure}{.16\textwidth}
            \includegraphics[width=\linewidth]{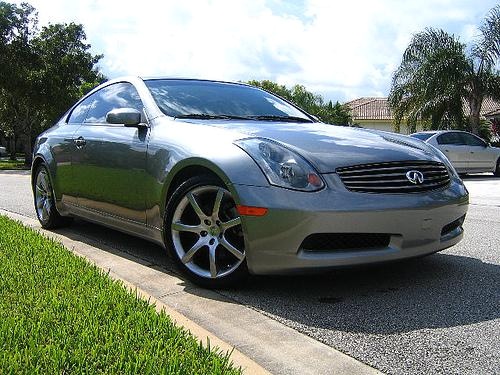}\caption{ $\Delta_v=0.005$}\label{fig:b_sup}
         \end{subfigure}  &
         \begin{subfigure}{.16\textwidth}
            \includegraphics[width=\linewidth]{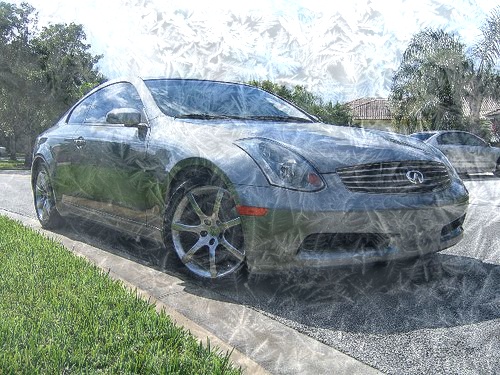}\caption{ $\Delta_v=0.71$}\label{fig:c_sup}
         \end{subfigure} &
         \begin{subfigure}{.16\textwidth}
         \includegraphics[width=\linewidth]{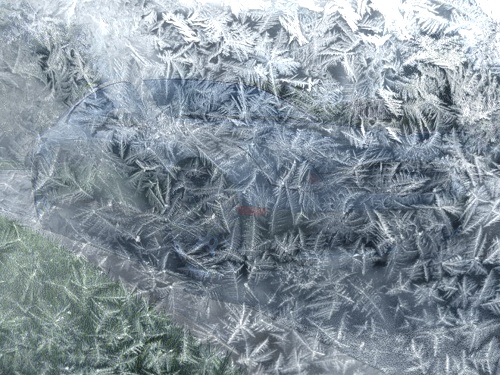}\caption{$\Delta_v=0.96$ }\label{fig:d_sup}
         \end{subfigure}   
    \end{tabular}
}
\caption{\small Examples of images from Imagenet~\cite{ILSVRC2012} with different levels of added frost.}
%The images are from Imagenet~\cite{ILSVRC2012} and the transformation is from Imagenet-c~\cite{hendrycks2019robustness}. }\
\label{fig:example_sup}
\end{figure}

\noindent
\underline{Example:} In Fig.~\ref{fig:example_sup}, the visual change of the original image Fig.~\ref{fig:a_sup} is $0$, since no changes are applied; and Fig.~\ref{fig:b_sup} has minimal frost added, which caused minimal change in visual quality so $\Delta_v = 0.005$; and Fig.~\ref{fig:c_sup} and Fig.~\ref{fig:d_sup} have more frost and thus higher $\Delta_v$ values $0.71$ and $0.96$, respectively.

\begin{algorithm}[h]
\caption{VCR Estimation}
\label{alg:vcr-est}
% \begin{algorithmic}
    \vspace{2pt}
    \KwIn{$\begin{cases}
   \text{model } f(x)\\
   \text{transformation } T_X, \text{with parameter domain } C\\
   \text{input dataset } \{x_k\} \text{ for consistency [or } \{(x_k,y_k)\} \text{ for accuracy]}\\
   \text{generated test set size } N\\
   \text{visual change resolution } M\\
   \text{minimum number of points per bin } L
    \end{cases}$}
    \KwOut{estimated VCR $\hat{\robustnesssymbol}_p$ [or $\hat{\robustnesssymbol}_a$]\vspace{0.5em}}
    Initialize histograms $\textit{count}_j$ and $\textit{correct}_j$ with empty counts, for all $j\in[0..M-1]$\\
    Initialize performance data array $P_j$ with $-1$, denoting missing data points for $j$, for all $j\in[0..M-1]$\\
    \For{$i \gets 0 \mathbf{~to~} N-1$}{
        draw random $x$ from $\{x_k\}$ [or $(x,y)$ from $\{(x_k,y_k)\}$]\\
        $c \sim \textit{Uniform}(C)$\\
        $x' \gets T_X(x,c)$\\
        $v \gets \Delta_v(x,x')$\\
        $j \gets \floor{v(M-1)}$\\
        $\textit{count}_j \gets \textit{count}_j + 1$\\
        \uIf{$f(x')=f(x)$ {\normalfont[or} $f(x')=y${\normalfont]}}{
            $\textit{correct}_j \gets \textit{correct}_j + 1$
        }
    }
    \For{$j \gets 0 \mathbf{~to~} M-1$}{
        \uIf{$\textit{count}_j\geq L$}{
            $P_j \gets \frac{\textit{correct}_j}{\textit{count}_j}$
        }
    }
    $s \gets \textit{FitMonotonicSpline}(P)$\\
    $\hat{\robustnesssymbol} \gets \int_0^{1} s(v) dv$\\
    \textbf{return} $\hat{\robustnesssymbol}$\\[2.5pt]
% \end{algorithmic}
\end{algorithm}

\noindent
{\bf VCR Estimation Algorithm}.
Algorithm~\ref{alg:vcr-est} gives the pseudo-code of the VCR estimation procedure described under ``Testing VCR'' in the main body of the paper. The algorithm takes a model $f(x)$; a transformation $T_X$ with its parameter domain $C$; an input dataset; the size $N$ of the dataset of transformed images to be generated; the visual change resolution $M$, over which the model performance will be estimated; and the minimum size $L$ of a bin to be used to estimate the performance for that bin. The input dataset consist of images $x_k\sim P_X$ for estimating VCR wrt. consistency, or images and their labels for estimating VCR wrt. accuracy. We use $M=40$ in our experiments, which is a standard choice for calculating average precision in object detection; for example, it is used in the current version of the KITTI benchmark~\cite{Geiger2013IJRR}.
%We also use $L=20$ as a minimum threshold to compute the performance for a given $v$.

Our algorithm first initializes two histogram arrays to keep the counts of the tested data points and their consistent or accurate predictions, respectively, and an array to keep the performance data, with each of the three arrays having size $M$. In each iteration, the first for-loop samples an image $x$ and transformation parameter $c$, and produces a transformed image $x'$. It then computes the visual change value $v$ and records the result of testing $f(x')$ in the histograms. The second for-loop computes the performance data as a relative frequency of correct predictions. A monotonic smoothing spline is fit into the performance data, and the VCR is computed as the area under the spline.

Note that this algorithm samples $c$ uniformly, which will lead to a varying number of performance samples per point in the performance data array $P$. As already discussed, the number of performance samples impacts the performance estimate uncertainty at this point, and in an extreme case some of the $\Delta_v$ bins in $P_i$ may be even empty (i.e., have value -1). These missing points are mitigated by fitting the spline over the entire $\Delta_v$ range, while anchoring it with known values for the first and last bins. In particular, the accuracy spline $s_a$ always starts at the left with the accuracy for clean images, and the consistency spline $s_p$ starts with 1 for models (assuming deterministic NNs).

A possible approach to obtain a sample set with a more uniform coverage of $\Delta_v$ would be to (1) fit a strictly monotonic spline into $(c,\Delta_v)$ values obtained from $c\sim \textit{Uniform}(C)$ as in Alg.~\ref{alg:vcr-est}, (2) take a set of samples $\Delta_v\sim \textit{Uniform}(0,1)$, (3) map the latter to a new sample from $C$ using the inverted spline, and repeat these steps now using the new sample from $C$. These steps would need to be run iteratively until a sufficient coverage is obtained. Such an algorithm would be computationally expensive, however.

\section{Comparison of $\Delta_v$ Distribution}
\label{sec:app_cov}
\begin{figure}[h]
\centering
\scalebox{0.7}{
    \begin{tabular}{c c  >{\columncolor{gray!10}}c >{\columncolor{gray!10}}c >{\columncolor{gray!25}}c >{\columncolor{gray!25}}c}
        %\toprule
         %& & & & \\
         \multicolumn{2}{c}{\small Brightness} &\multicolumn{2}{>{\columncolor{gray!10}}c}{\small Defocus Blur} & \multicolumn{2}{>{\columncolor{gray!25}}c}{\small Gaussian Noise}\\
          \begin{subfigure}[b]{0.20\textwidth}
         \centering
         \includegraphics[width=\textwidth]{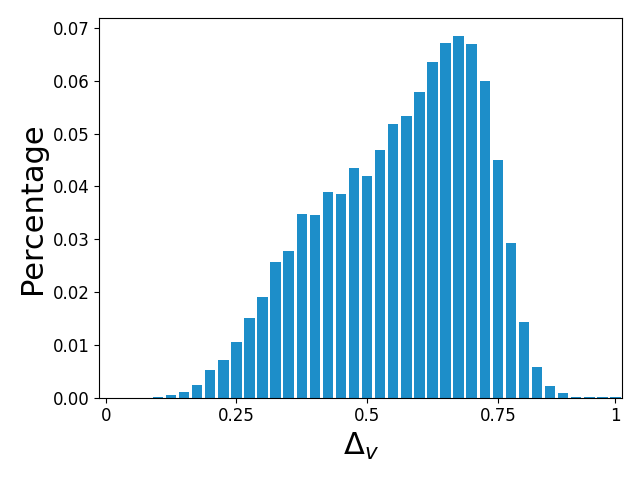}
         \caption{\scriptsize \imagenetc}
         \label{fig:brightness_c}
     \end{subfigure}
     &
     \begin{subfigure}[b]{0.20\textwidth}
         \centering
         \includegraphics[width=\textwidth]{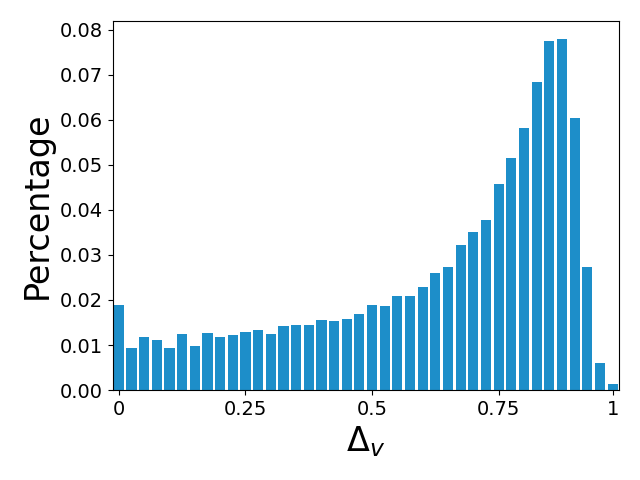}
         \caption{\scriptsize \method}
         \label{fig:brightness_v}
     \end{subfigure}
     &
     \begin{subfigure}[b]{0.20\textwidth}
         \centering
         \includegraphics[width=\textwidth]{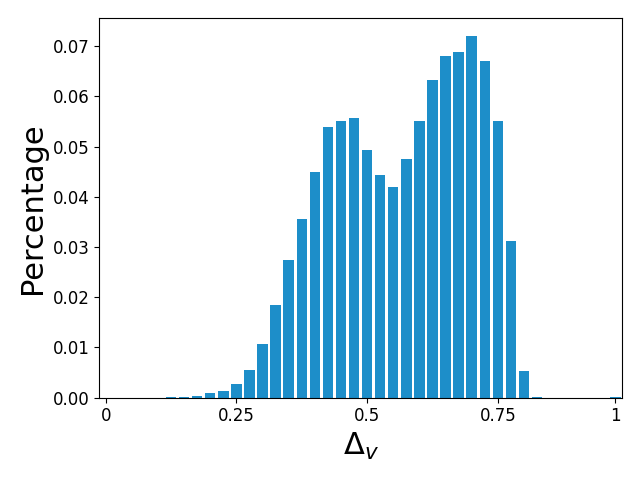}
         \caption{\scriptsize \imagenetc}
         \label{fig:defocus_c}
     \end{subfigure}
     &
     \begin{subfigure}[b]{0.20\textwidth}
         \centering
         \includegraphics[width=\textwidth]{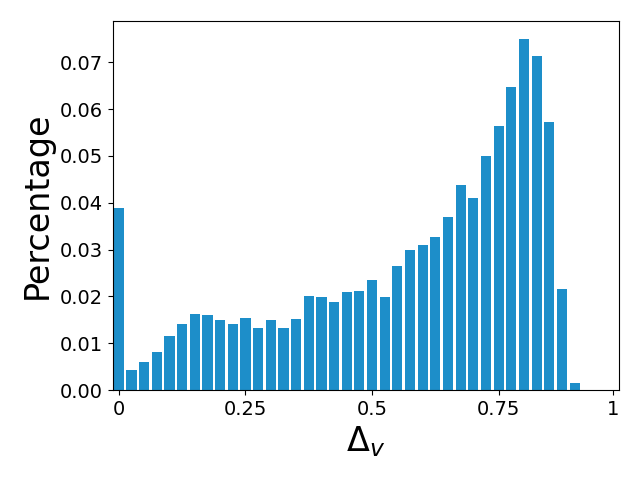}
         \caption{\scriptsize \method}
         \label{fig:defocus_v}
     \end{subfigure}
     & \begin{subfigure}[b]{0.20\textwidth}
         \centering
         \includegraphics[width=\textwidth]{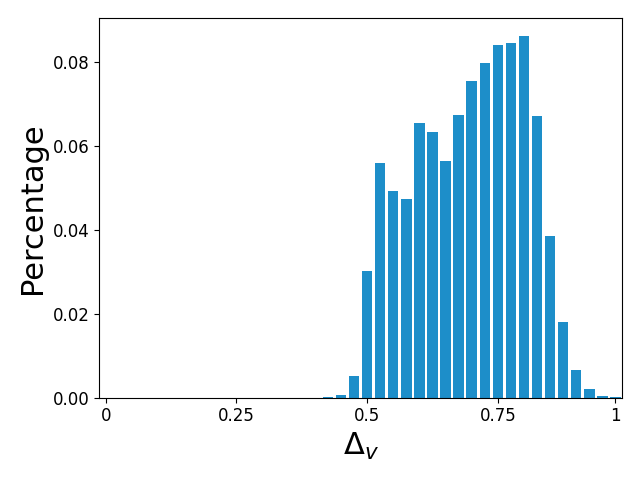}
         \caption{\scriptsize \imagenetc}
         \label{fig:gauss_noise_c}
     \end{subfigure}
     &
     \begin{subfigure}[b]{0.20\textwidth}
         \centering
         \includegraphics[width=\textwidth]{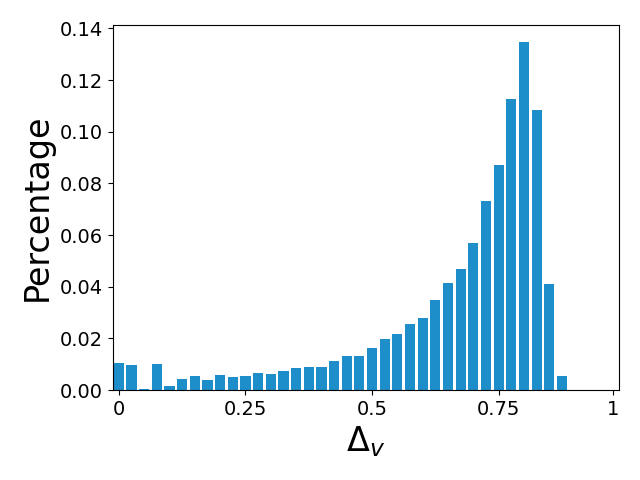}
         \caption{\scriptsize \method}
         \label{fig:gauss_noise_v}
     \end{subfigure}
     \\

     \multicolumn{2}{c}{\small Glass Blur} &\multicolumn{2}{>{\columncolor{gray!10}}c}{\small Impulse Noise} &\multicolumn{2}{>{\columncolor{gray!25}}c}{\small Shot Noise}\\
        
     \begin{subfigure}[b]{0.20\textwidth}
         \centering
         \includegraphics[width=\textwidth]{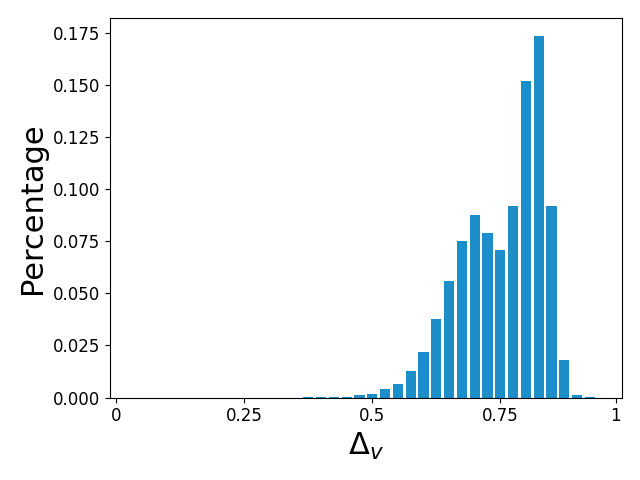}
         \caption{\scriptsize \imagenetc}
         \label{fig:glass_blur_c}
     \end{subfigure}
     &
     \begin{subfigure}[b]{0.20\textwidth}
         \centering
         \includegraphics[width=\textwidth]{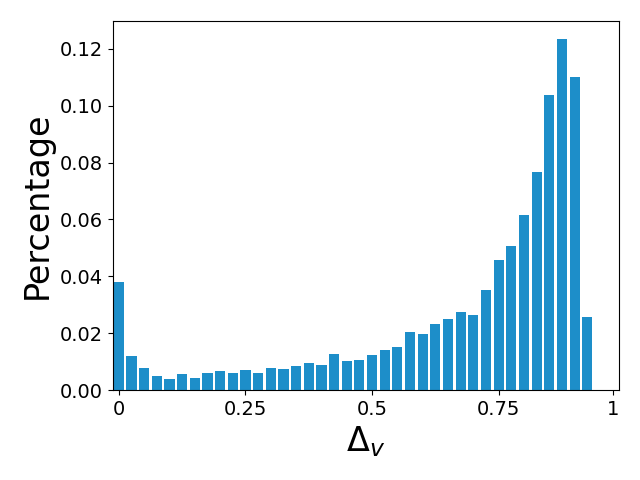}
         \caption{\scriptsize \method}
         \label{fig:glass_blur_v}
     \end{subfigure} 
     & \begin{subfigure}[b]{0.20\textwidth}
         \centering
         \includegraphics[width=\textwidth]{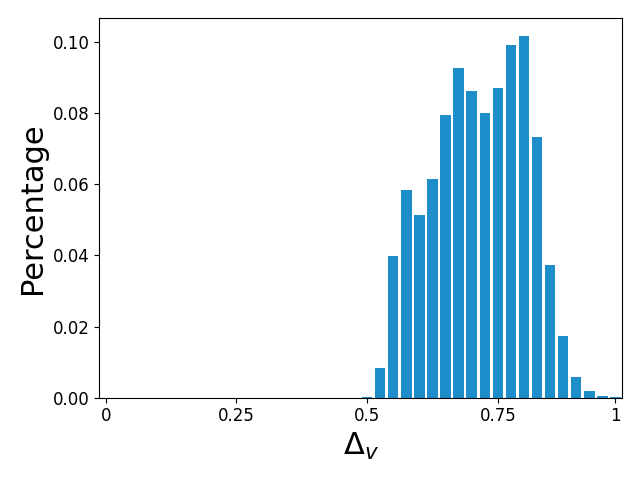}
         \caption{\scriptsize \imagenetc}
         \label{fig:impulse_c}
     \end{subfigure}
     &
     \begin{subfigure}[b]{0.20\textwidth}
         \centering
         \includegraphics[width=\textwidth]{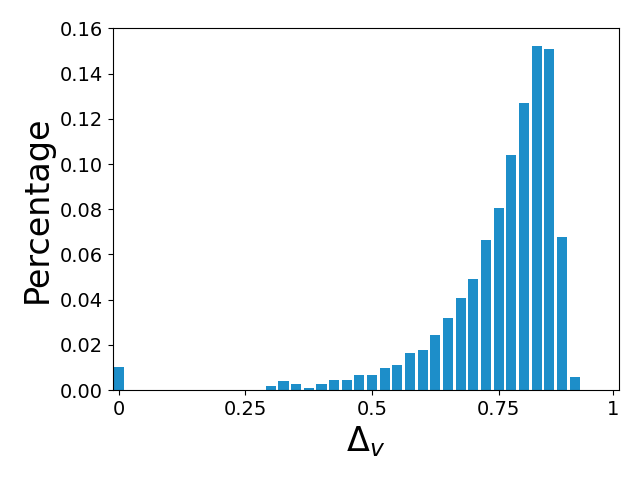}
         \caption{\scriptsize \method}
         \label{fig:impulse_v}
     \end{subfigure}
     &
     \begin{subfigure}[b]{0.20\textwidth}
         \centering
         \includegraphics[width=\textwidth]{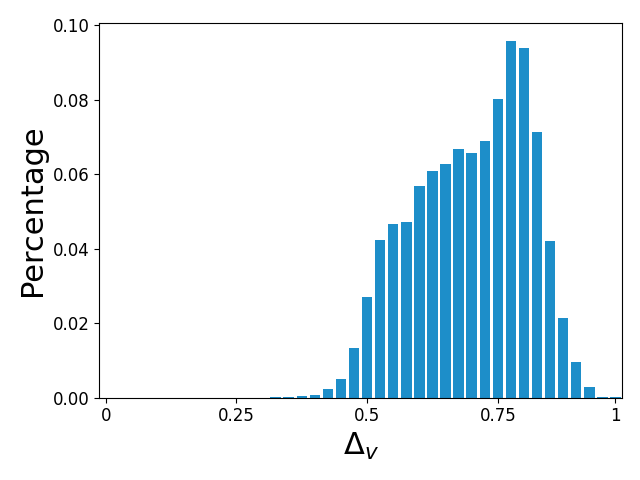}
         \caption{\scriptsize \imagenetc}
         \label{fig:shot_c}
     \end{subfigure}
     &
     \begin{subfigure}[b]{0.20\textwidth}
         \centering
         \includegraphics[width=\textwidth]{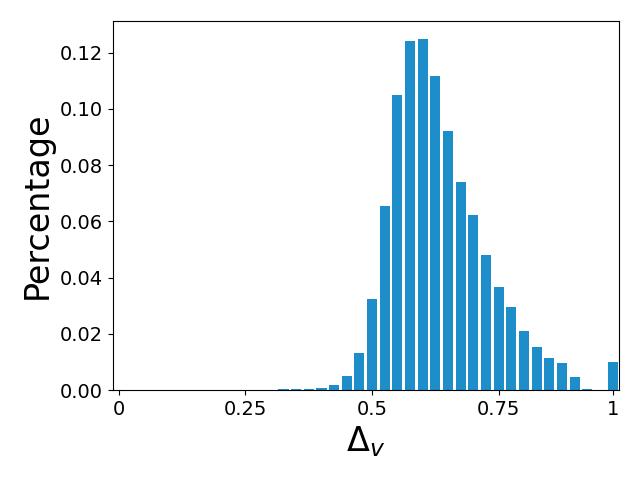}
         \caption{\scriptsize \method}
         \label{fig:shot_v}
     \end{subfigure}\\
     
     \multicolumn{2}{c}{\small Frost} &\multicolumn{2}{>{\columncolor{gray!10}}c}{\small Gaussian Blur} &\multicolumn{2}{>{\columncolor{gray!25}}c}{\small Motion Blur} \\
         \begin{subfigure}[b]{0.20\textwidth}
         \centering
         \includegraphics[width=\textwidth]{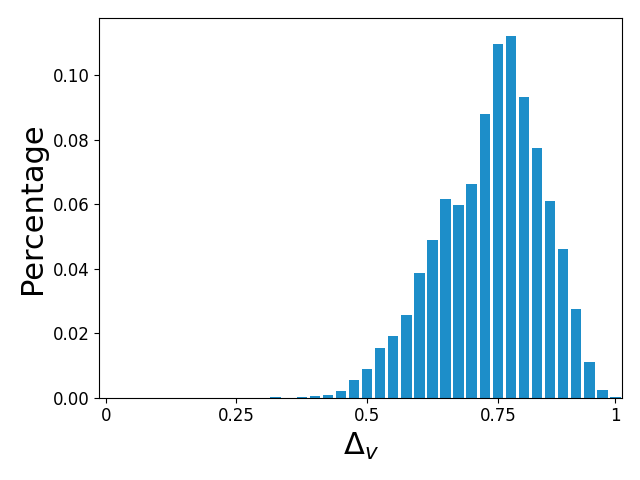}
         \caption{\scriptsize \imagenetc}
         \label{fig:frost_c}
     \end{subfigure}
     &
     \begin{subfigure}[b]{0.20\textwidth}
         \centering
         \includegraphics[width=\textwidth]{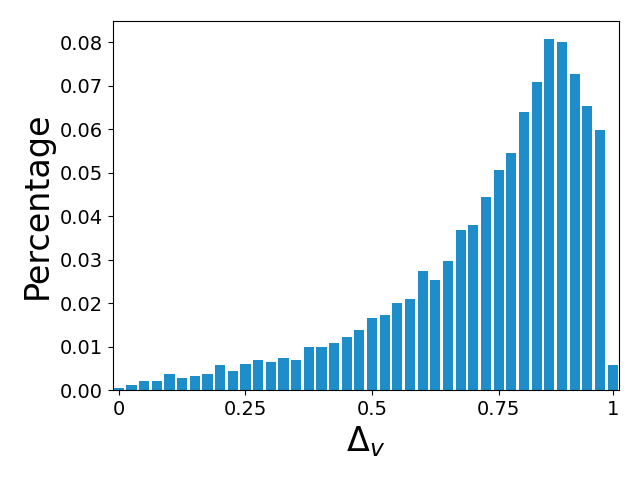}
         \caption{\scriptsize \method}
         \label{fig:frost_v}
     \end{subfigure}
     &
     \begin{subfigure}[b]{0.20\textwidth}
         \centering
         \includegraphics[width=\textwidth]{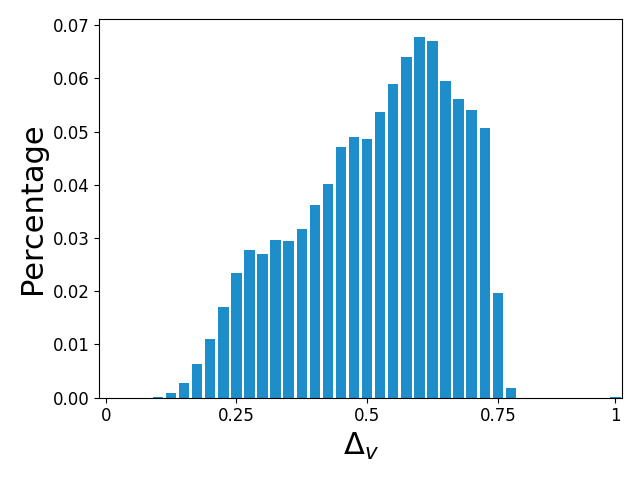}
         \caption{\scriptsize \imagenetc}
         \label{fig:gauss_blur_c}
     \end{subfigure}
     &
     \begin{subfigure}[b]{0.20\textwidth}
         \centering
         \includegraphics[width=\textwidth]{resources/our_distribution/gaussian_blur.png}
         \caption{\scriptsize \method}
         \label{fig:gauss_blur_v}
     \end{subfigure}
      &
     \begin{subfigure}[b]{0.20\textwidth}
         \centering
         \includegraphics[width=\textwidth]{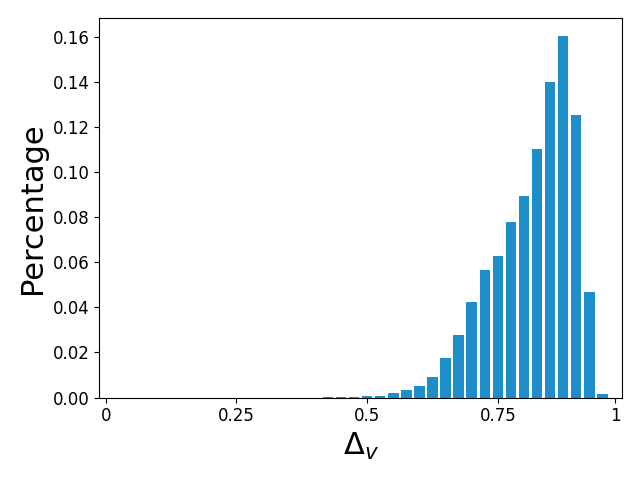}
         \caption{\scriptsize \imagenetc}
         \label{fig:motion_blur_c}
     \end{subfigure}
     &
     \begin{subfigure}[b]{0.20\textwidth}
         \centering
         \includegraphics[width=\textwidth]{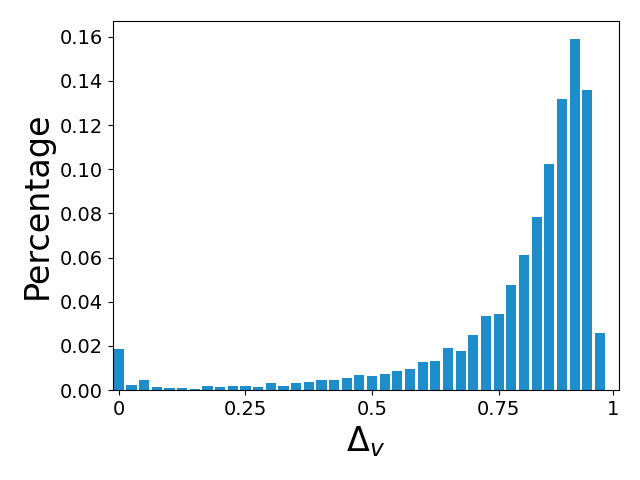}
         \caption{\scriptsize \method}
         \label{fig:motion_blur_v}
     \end{subfigure}\\
     
    \end{tabular}
}
%\vspace{-0.1in}
\caption{\small Comparison of $\Delta_v$ distribution between \imagenetc{} and \method. The figures are histograms, where x-axis is $\Delta_v$, y-axis is percentage of images.}
%The images are from Imagenet~\cite{ILSVRC2012} and the transformation is from Imagenet-c~\cite{hendrycks2019robustness}. }\
\label{fig:coverage}
%\vspace{-0.2in}
\end{figure}

In Fig.~\ref{fig:coverage} below we compare the $\Delta_v$ distribution of validation images from \imagenetc{} and those generated by our benchmark. We include all 9 corruption functions shared between \imagenetc{} and our benchmark. 
Note that all of our images are generated by sampling uniformly in the parameter domain, while \imagenetc{} images are generated with 5 pre-selected parameter values. We can observe two major differences in the distributions. 
First we can see that because of difference in the parameter values used, the $\Delta_v$ distributions between \imagenetc{} and our benchmark peak at different values. For example, for Brightness in Fig.~\ref{fig:brightness_c} and Fig.~\ref{fig:brightness_v}, most \imagenetc{} images have $\Delta_v$ values between $0.4$ to $0.8$, while most \method{} images are between $0.6$ and $0.9$; a similar observation holds for Defocus Blur and Gaussian Blur. Second, we notice that \imagenetc{} images cannot cover all $\Delta_v$ values. Specifically, Fig.~\ref{fig:defocus_c} for Defocus Blur shows that \imagenetc{} validation set does not contain images with $\Delta_v$ greater than $0.8$ and less than $0.2$. The same can be observed for all corruption functions shown in Fig.~\ref{fig:coverage}. 
These two differences indicate that, when considering the full range of visual changes that a corruption function can incur, using \imagenetc{} can lead to biased results.
%, Fig.~\ref{fig:gauss_noise_c} for Gaussian Noise, Fig.~\ref{}
%This section shows the $\Delta_v$ distribution of all \imagenetc{} transformations, you can see that most transformations cannot cover the entire domain of .
%\red{TODO: Replace our fixed transformation images (e.g. brightness) in Fig.~\ref{fig:coverage}}

\section{Extra Evaluation Results}
\label{sec:app_extra}

\subsection{Prediction Similarity of Visually Similar Corruption Functions}
\begin{figure}[h]
    \centering
    \begin{subfigure}[b]{0.32\textwidth}
         \centering
         \includegraphics[width=\textwidth]{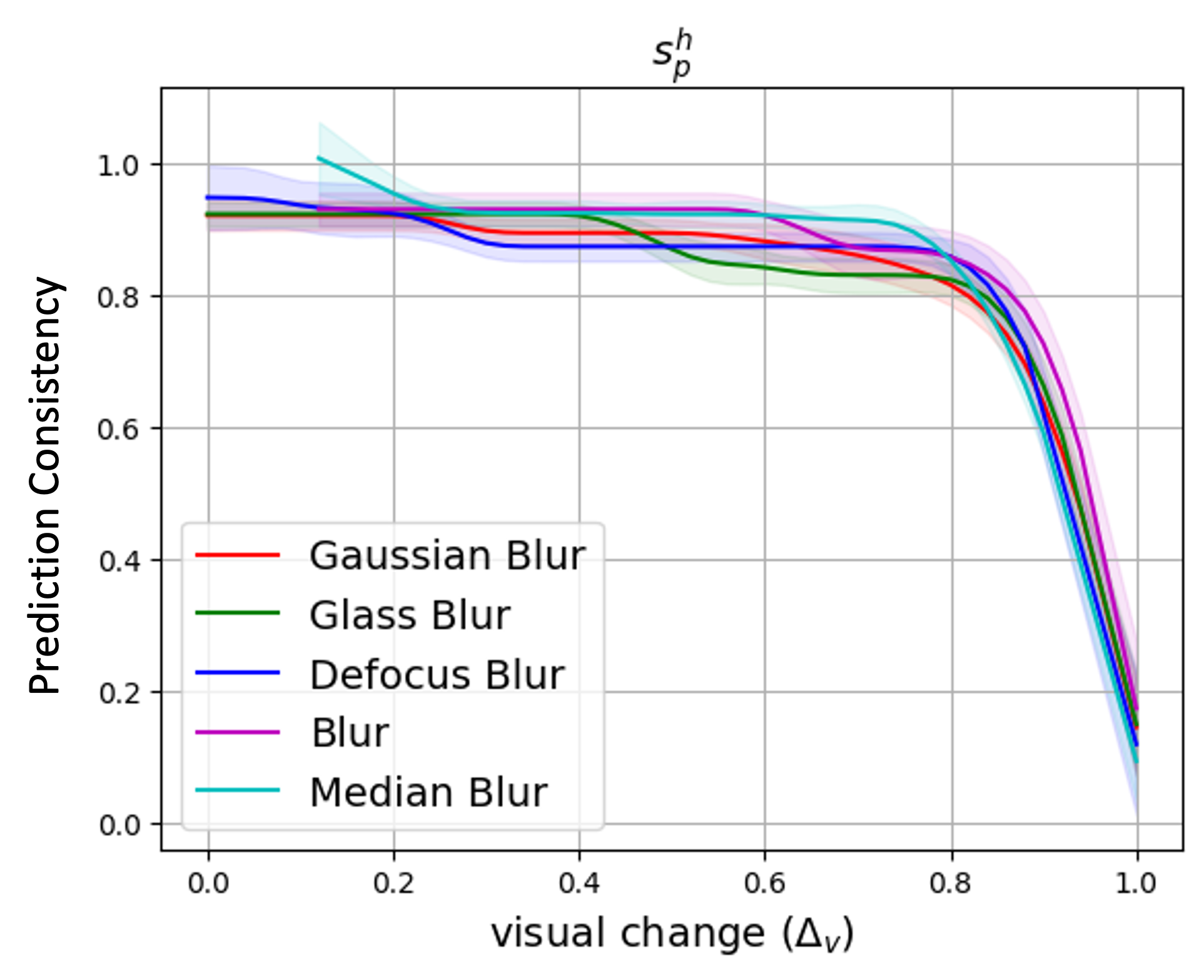}
         \caption{ \small Blur corruption functions}
         \label{fig:blur_sup}
     \end{subfigure}
     \begin{subfigure}[b]{0.32\textwidth}
         \centering
         \includegraphics[width=\textwidth]{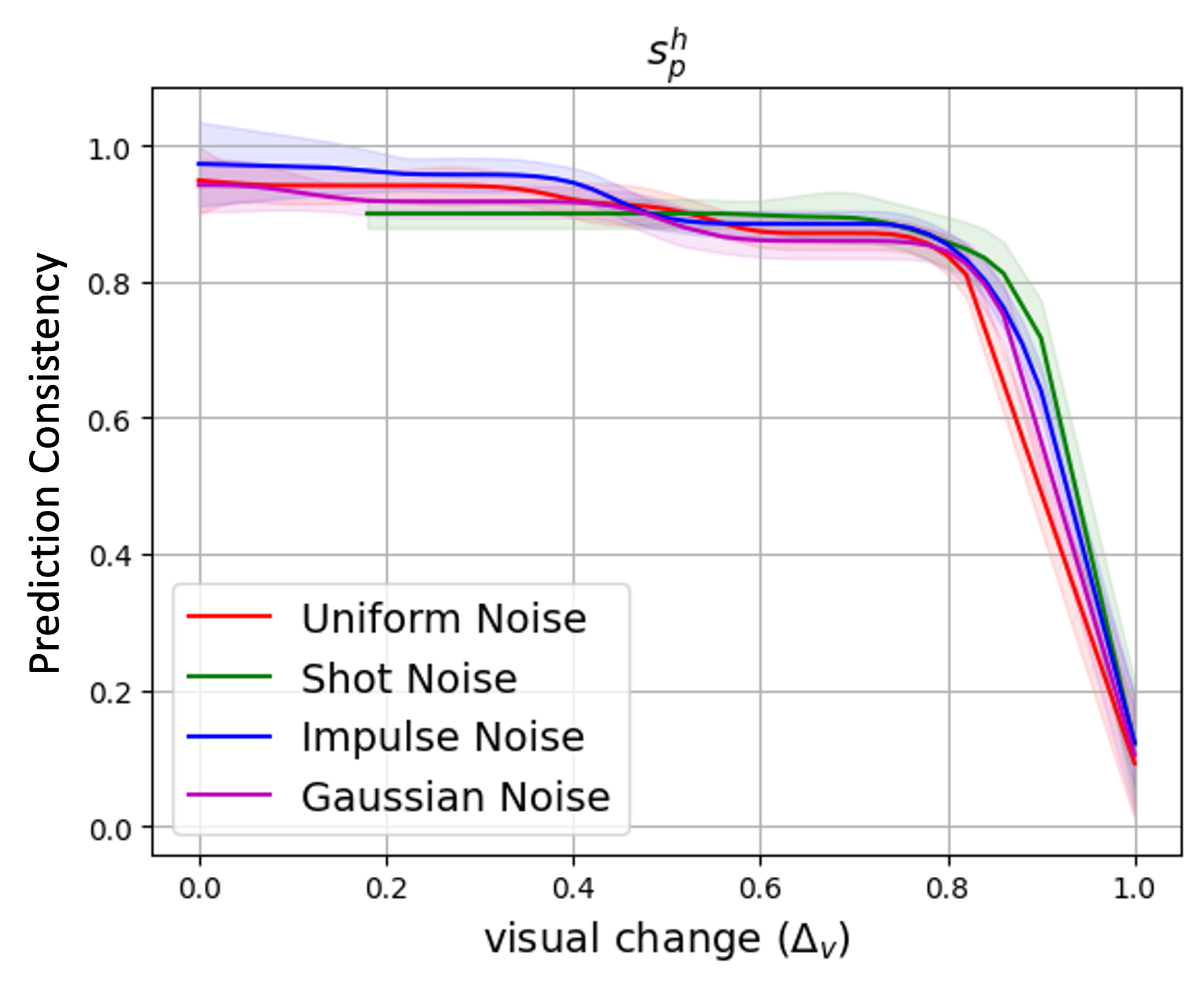}
         \caption{\small Noise corruption functions}
         \label{fig:noise_sup}
     \end{subfigure}
     \begin{subfigure}[b]{0.32\textwidth}
         \centering
         \includegraphics[width=\textwidth]{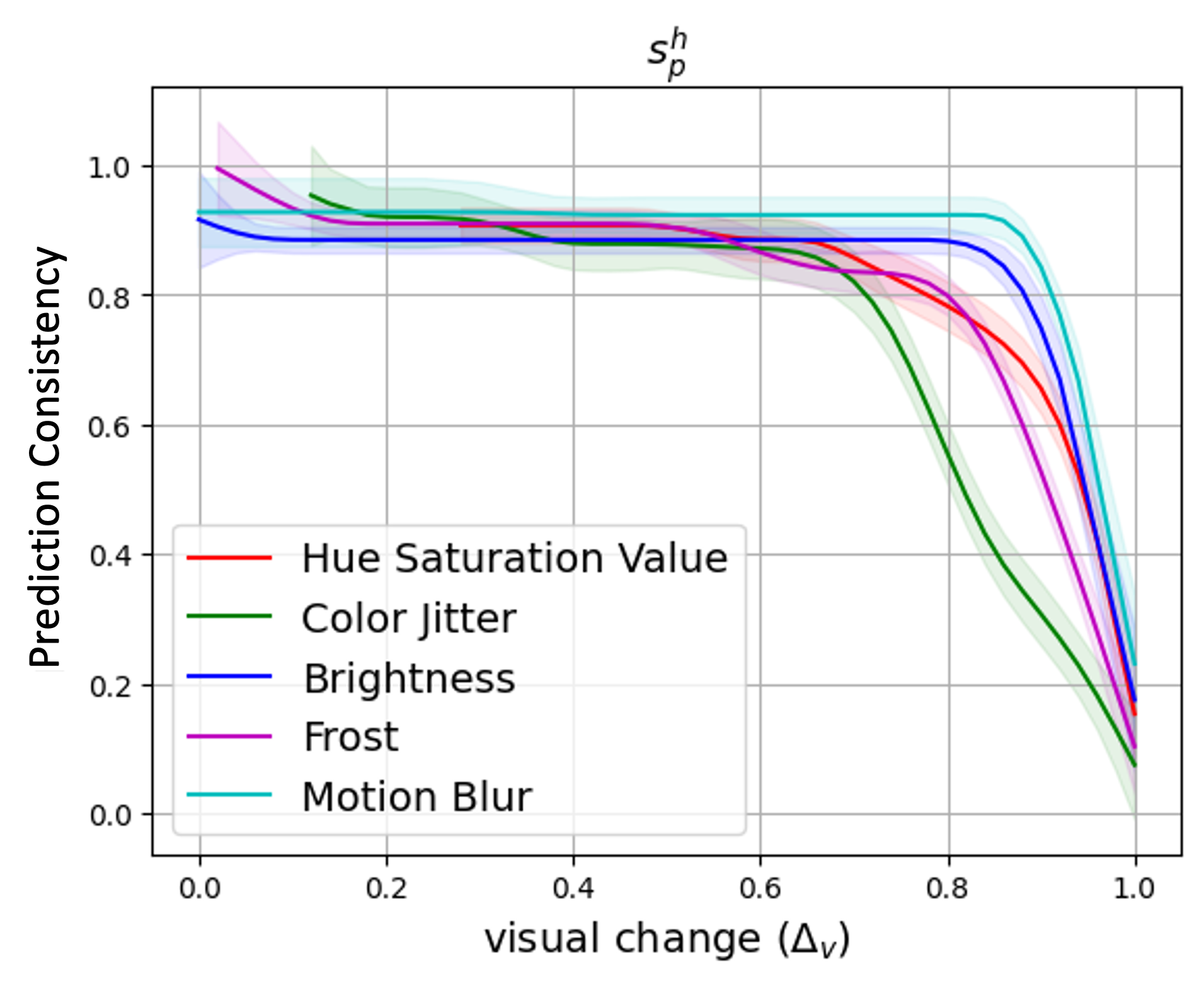}
         \caption{\small Dissimilar corruption functions}
         \label{fig:non_sup}
     \end{subfigure}
    \caption{\small Comparing human performance spline curves $s^h_p$ for similar and dissimilar corruption functions. For each curve, the coloured region around the curve is the $83\%$ confidence interval used for comparison of similarity~\cite{koenker-94}.}
    \label{fig:similar_sup}
    %\vspace{-0.25in}
\end{figure}

In the paper, to check that human robustness data is transferable between two similar corruption functions, we checked whether the $83\%$ confidence interval of the spine curves $s^h_a$ and $s^h_p$ for similar corruption functions overlap. The results for $s^h_a$ in Fig.~\ref{fig:similar}. We also include results for $s^h_p$ in Fig.~\ref{fig:similar_sup}. We can observe that, similar to $s^h_a$, $s^h_p$ for similar corruption functions are similar, thus human data is transferable. 

%we included spline curves estimated from experiment data and their $83\%$ confidence interval for all the corruption functions.

\subsection{CO2 Emission}

CO2 Emission is calculated as \texttt{CO2 emissions (kg) = (Power consumption in kilowatts) x (Daily usage time in hours) x (Emissions factor in kgCO2/kWh)}

Our carbon intensity is around 25\,g/kWh.
During benchmark dataset generation, there is no GPU usage, and the CPU usage is 200\,W. Each corruption function takes around 1.5 hour to generate a dataset with 50,000 images.
During evaluation, the CPU power usage is around 160\,W; and GPU power usage ranges between 50-170\,W depending on the model. Each evaluation takes 30-60 minutes, depending on the corruption function type. 
Let's assume the power usage of other components is 50\,W in total.
If we assume the total power usage is $((200+50)\times 1.5+(170+160+50))/1{,}000=0.755\,$kWh for each experiment,  the CO2 emission is $0.755\times25=18.875$\,g for each experiment (corruption function type).

\subsection{\robustness{} Evaluation}
% We presented VCR evaluation results for Gaussian Noise and Shot Noise in the main body of the paper, and 
% we include all 12 other corruption functions we studied below.

In the main body of the paper, we have compared VCR robustness results with \imagenetc{} on Gaussian Noise, and we presented the assessing VCR in relation to human performance with our human-aware metrics \textit{HMRI} and \textit{MRSI} for Gaussian Noise and Shot Noise. Below, we first include the comparison between VCR and \imagenetc{} for all \imagenetc{} 9 corruption functions we studied. Then, include detailed evaluation results with our human-aware metrics for all 12 other corruption functions we studied.

%already presented the VCR evaluation results for gaussian noise and shot noise. Additionally, we have compared VCR with \imagenetc{} on gaussian noise. In this section, we include detailed evaluation results for 12 other corruption functions we studied. Furthermore, we also include comparison between VCR and \imagenetc{} for all 9 corruptions they share.

% We presented VCR evaluation results for Gaussian Noise and Shot Noise in the main body of the paper, as well as the comparison between \imagenetc{} and VCR using gaussian noise. Below, we include all 12 other corruption functions we studied below, as well as comparison with \imagenetc{} for the 9 common corruptions.

\newpage
% Compare VCCR with ImageNet-C

\begin{figure}
    \centering
    \begin{subfigure}{0.32\textwidth}
        \includegraphics[width=\textwidth]{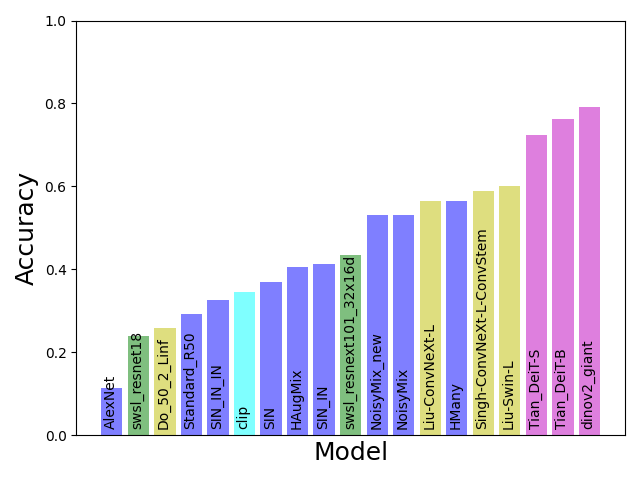}
        \caption{\scriptsize \textsc{ImageNet-C} Gaussian Noise Accuracy}
        \label{fig:imagenetc-gaussian-noise-acc-bar}
    \end{subfigure}
    \begin{subfigure}{0.32\textwidth}
        \includegraphics[width=\textwidth]{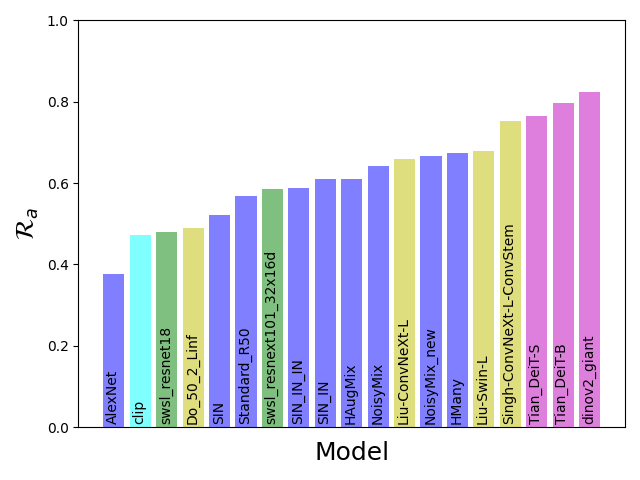}
        \caption{\scriptsize Gaussian Noise $\hat{\mathcal{R}}_a$}
        \label{fig:vcr-gaussian-noise-acc-bar}
    \end{subfigure}
    \begin{subfigure}{0.32\textwidth}
        \includegraphics[width=\textwidth]{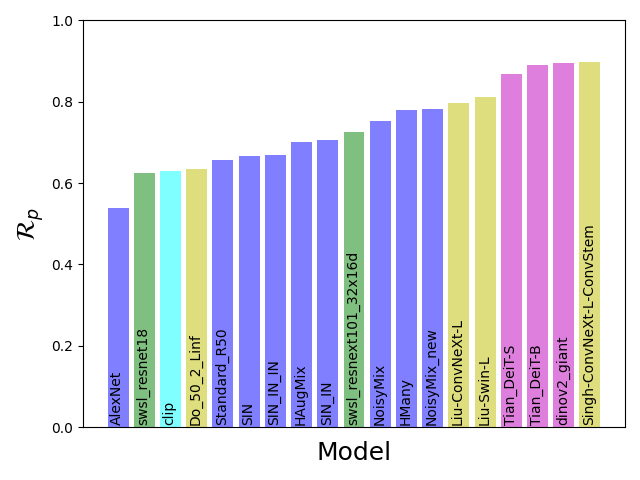}
        \caption{\scriptsize Gaussian Noise $\hat{\mathcal{R}}_p$}
        \label{fig:vcr-gaussian-noise-pred-sim-bar}
    \end{subfigure}
    \caption{Comparison between \textsc{ImageNet-C} and VCR with Gaussian Noise.}
    \label{fig:compare-imagenetc-vcr-gaussian-noise}

\end{figure}

\begin{figure}
    \centering
    \begin{subfigure}{0.32\textwidth}
        \includegraphics[width=\textwidth]{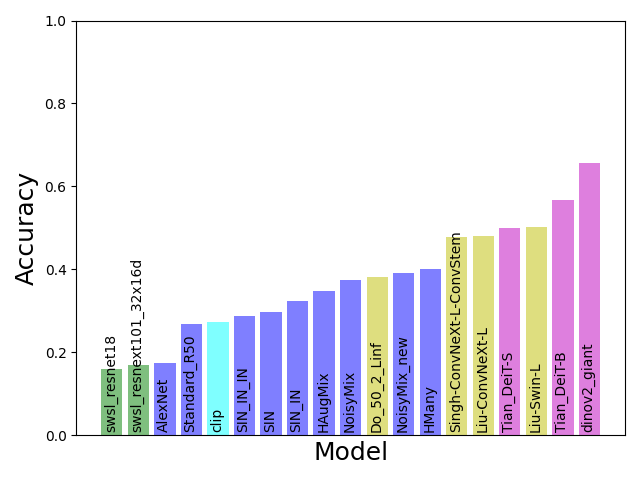}
        \caption{\scriptsize \textsc{ImageNet-C} Glass Blur Accuracy}
        \label{fig:imagenetc-glass-blur-acc-bar}
    \end{subfigure}
    \begin{subfigure}{0.32\textwidth}
        \includegraphics[width=\textwidth]{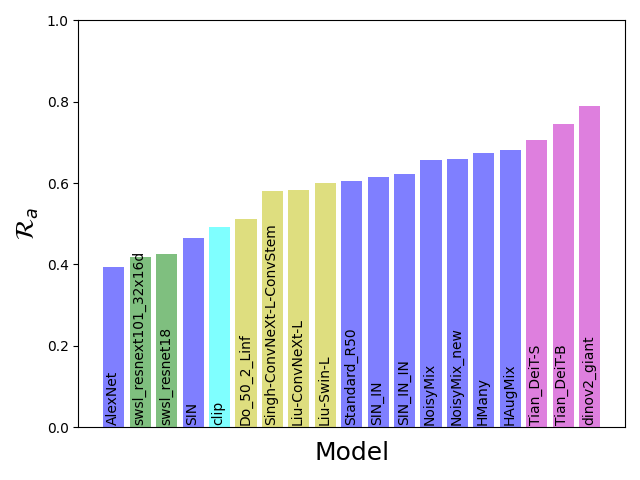}
        \caption{\scriptsize Glass Blur $\hat{\mathcal{R}}_a$}
        \label{fig:vcr-glass-blur-acc-bar}
    \end{subfigure}
    \begin{subfigure}{0.32\textwidth}
        \includegraphics[width=\textwidth]{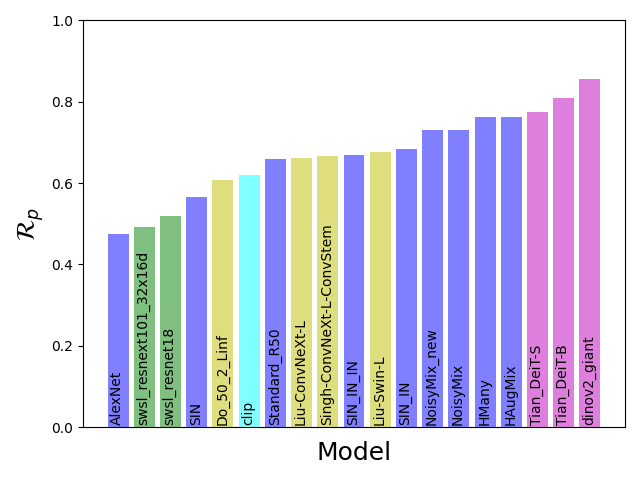}
        \caption{\scriptsize Glass Blur $\hat{\mathcal{R}}_p$}
        \label{fig:vcr-glass-blur-pred-sim-bar}
    \end{subfigure}
    \caption{Comparison between \textsc{ImageNet-C} and VCR with Glass Blur.}
    \label{fig:compare-imagenetc-vcr-glass-blur}

\end{figure}

\begin{figure}
    \centering
    \begin{subfigure}{0.32\textwidth}
        \includegraphics[width=\textwidth]{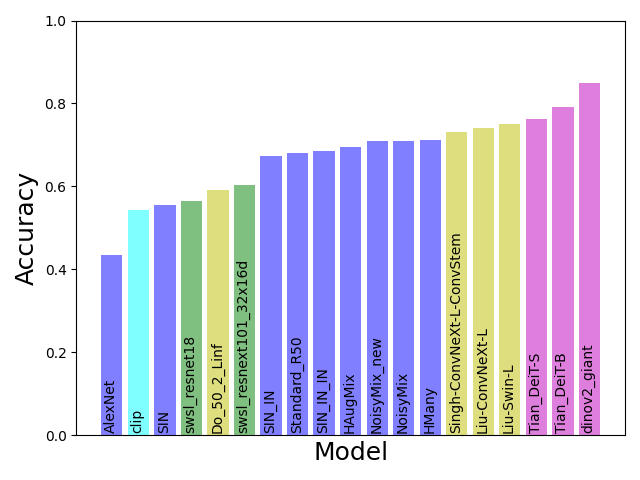}
        \caption{\scriptsize \textsc{ImageNet-C} Brightness Accuracy}
        \label{fig:imagenetc-brightness-acc-bar}
    \end{subfigure}
    \begin{subfigure}{0.32\textwidth}
        \includegraphics[width=\textwidth]{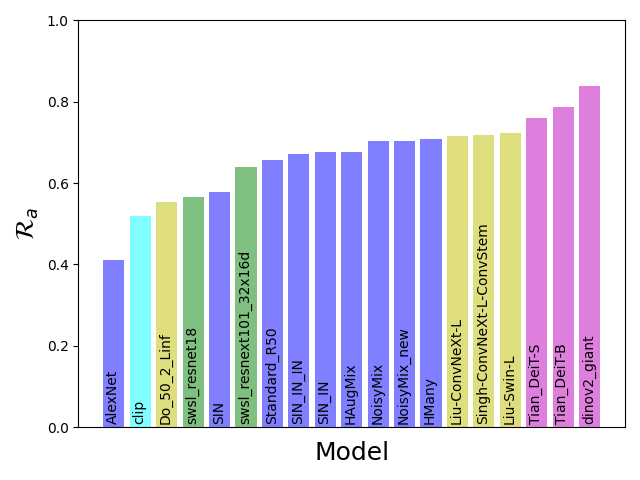}
        \caption{\scriptsize Brightness $\hat{\mathcal{R}}_a$}
        \label{fig:vcr-brightness-acc-bar}
    \end{subfigure}
    \begin{subfigure}{0.32\textwidth}
        \includegraphics[width=\textwidth]{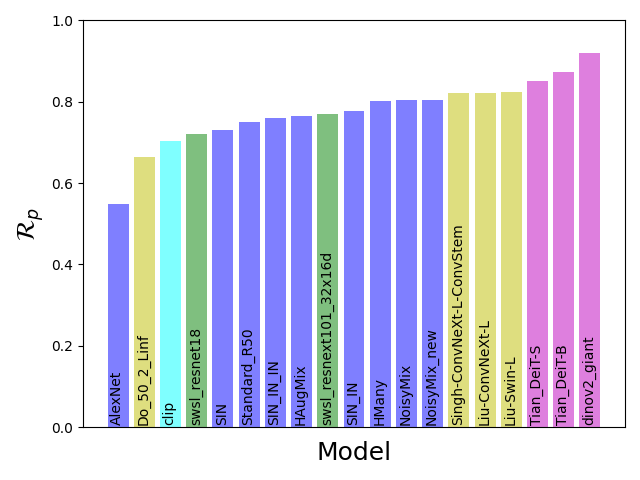}
        \caption{\scriptsize Brightness $\hat{\mathcal{R}}_p$}
        \label{fig:vcr-brightness-pred-sim-bar}
    \end{subfigure}
    \caption{Comparison between \textsc{ImageNet-C} and VCR with Brightness.}
    \label{fig:compare-imagenetc-vcr-brightness}

\end{figure}

\begin{figure}
    \centering
    \begin{subfigure}{0.32\textwidth}
        \includegraphics[width=\textwidth]{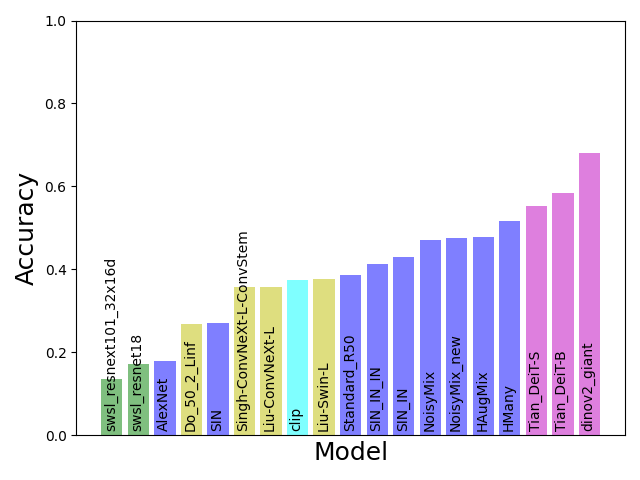}
        \caption{\scriptsize \textsc{ImageNet-C} Defocus Blur Accuracy}
        \label{fig:imagenetc-defocus-blur-acc-bar}
    \end{subfigure}
    \begin{subfigure}{0.32\textwidth}
        \includegraphics[width=\textwidth]{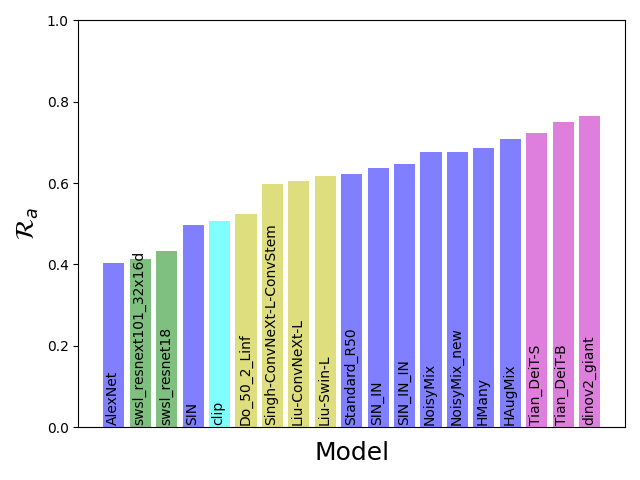}
        \caption{\scriptsize Defocus Blur $\hat{\mathcal{R}}_a$}
        \label{fig:vcr-defocus-blur-acc-bar}
    \end{subfigure}
    \begin{subfigure}{0.32\textwidth}
        \includegraphics[width=\textwidth]{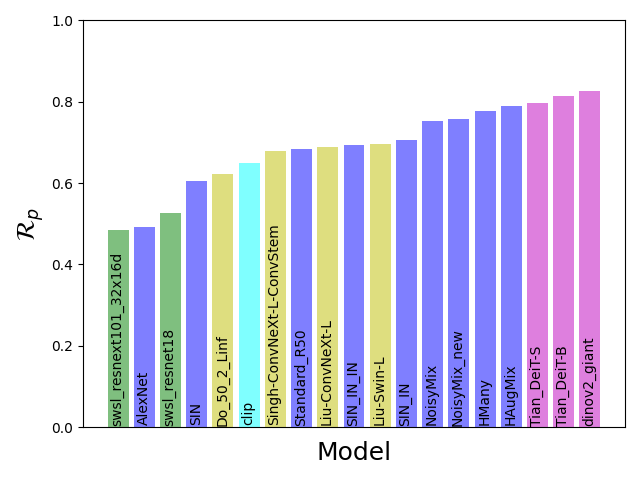}
        \caption{\scriptsize Defocus Blur $\hat{\mathcal{R}}_p$}
        \label{fig:vcr-defocus-blur-pred-sim-bar}
    \end{subfigure}
    \caption{Comparison between \textsc{ImageNet-C} and VCR with Defocus Blur.}
    \label{fig:compare-imagenetc-vcr-defocus-blur}

\end{figure}

\begin{figure}
    \centering
    \begin{subfigure}{0.32\textwidth}
        \includegraphics[width=\textwidth]{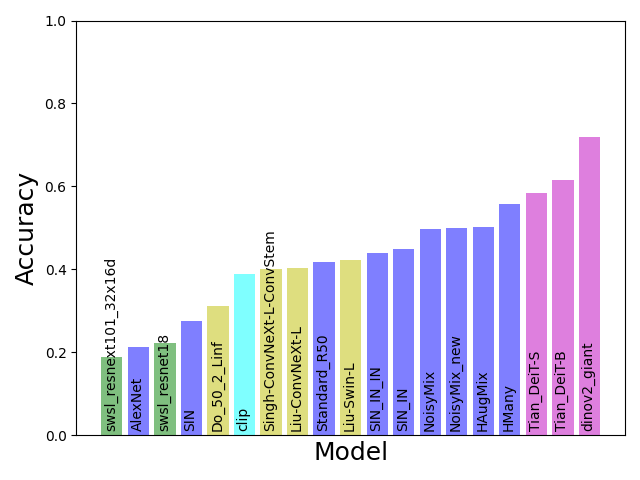}
        \caption{\scriptsize \textsc{ImageNet-C} Gaussian Blur Accuracy}
        \label{fig:imagenetc-gaussian-blur-acc-bar}
    \end{subfigure}
    \begin{subfigure}{0.32\textwidth}
        \includegraphics[width=\textwidth]{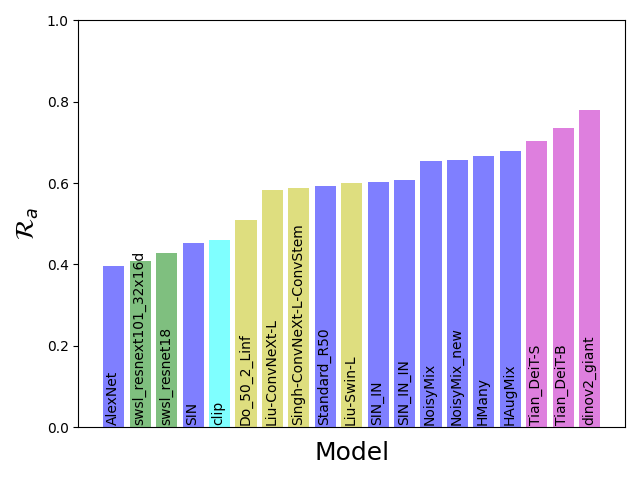}
        \caption{\scriptsize Gaussian Blur $\hat{\mathcal{R}}_a$}
        \label{fig:vcr-gaussian-blur-acc-bar}
    \end{subfigure}
    \begin{subfigure}{0.32\textwidth}
        \includegraphics[width=\textwidth]{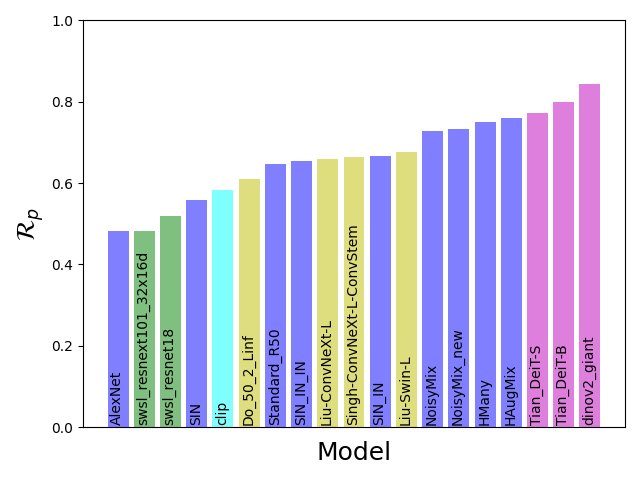}
        \caption{\scriptsize Gaussian Blur $\hat{\mathcal{R}}_p$}
        \label{fig:vcr-gaussian-blur-pred-sim-bar}
    \end{subfigure}
    \caption{Comparison between \textsc{ImageNet-C} and VCR with Gaussian Blur.}
    \label{fig:compare-imagenetc-vcr-gaussian-blur}

\end{figure}

\begin{figure}
    \centering
    \begin{subfigure}{0.32\textwidth}
        \includegraphics[width=\textwidth]{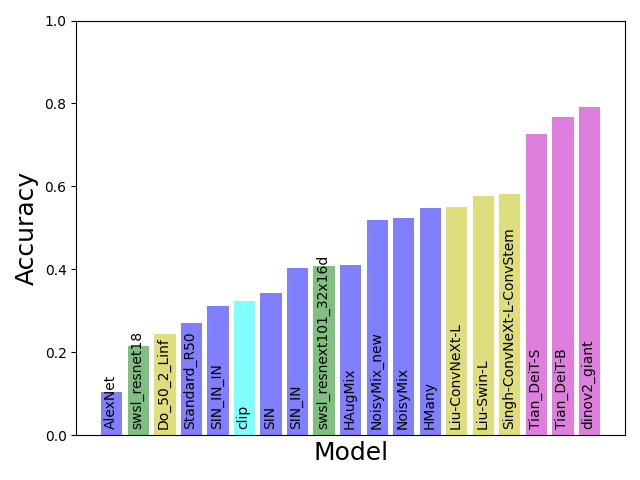}
        \caption{\scriptsize \textsc{ImageNet-C} Shot Noise Accuracy}
        \label{fig:imagenetc-shot-noise-acc-bar}
    \end{subfigure}
    \begin{subfigure}{0.32\textwidth}
        \includegraphics[width=\textwidth]{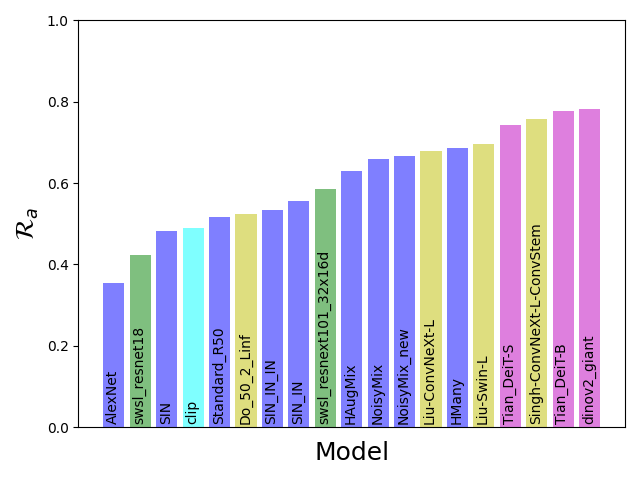}
        \caption{\scriptsize Shot Noise $\hat{\mathcal{R}}_a$}
        \label{fig:vcr-shot-noise-acc-bar}
    \end{subfigure}
    \begin{subfigure}{0.32\textwidth}
        \includegraphics[width=\textwidth]{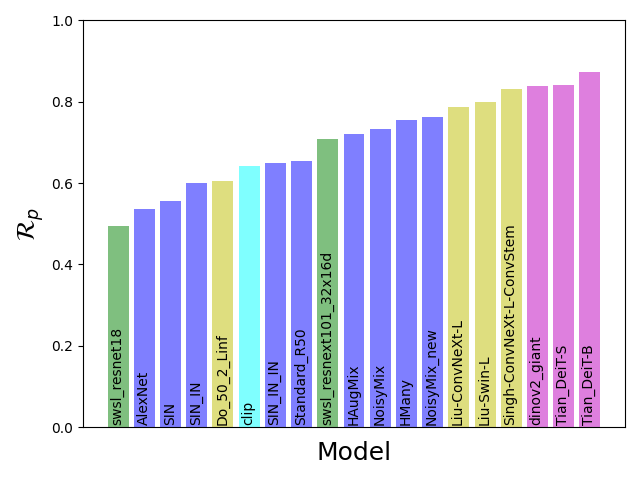}
        \caption{\scriptsize Shot Noise $\hat{\mathcal{R}}_p$}
        \label{fig:vcr-shot-noise-pred-sim-bar}
    \end{subfigure}
    \caption{Comparison between \textsc{ImageNet-C} and VCR with Shot Noise.}
    \label{fig:compare-imagenetc-vcr-shot-noise}

\end{figure}

\begin{figure}
    \centering
    \begin{subfigure}{0.32\textwidth}
        \includegraphics[width=\textwidth]{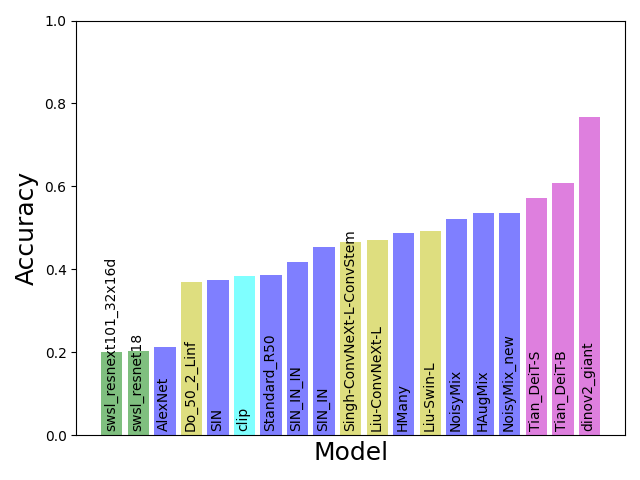}
        \caption{\scriptsize \textsc{ImageNet-C} Motion Blur Accuracy}
        \label{fig:imagenetc-motion-blur-acc-bar}
    \end{subfigure}
    \begin{subfigure}{0.32\textwidth}
        \includegraphics[width=\textwidth]{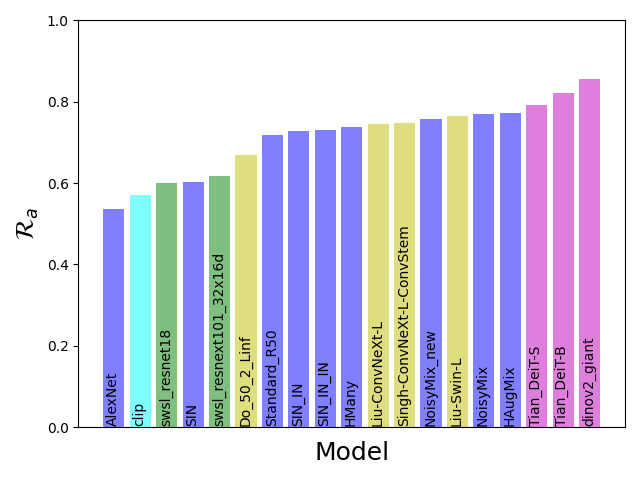}
        \caption{\scriptsize Motion Blur $\hat{\mathcal{R}}_a$}
        \label{fig:vcr-motion-blur-acc-bar}
    \end{subfigure}
    \begin{subfigure}{0.32\textwidth}
        \includegraphics[width=\textwidth]{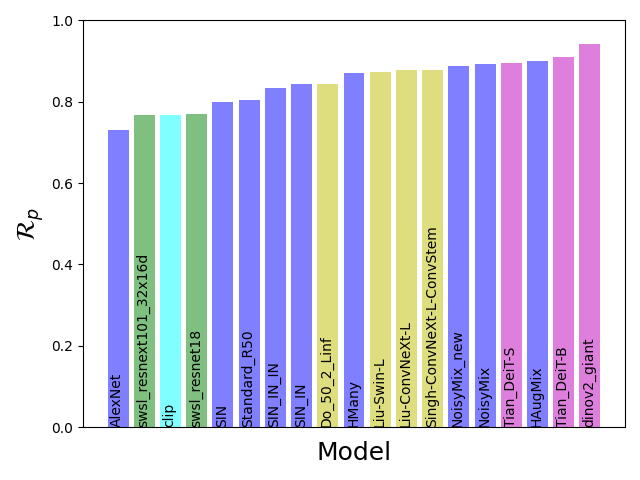}
        \caption{\scriptsize Motion Blur $\hat{\mathcal{R}}_p$}
        \label{fig:vcr-motion-blur-pred-sim-bar}
    \end{subfigure}
    \caption{Comparison between \textsc{ImageNet-C} and VCR with Motion Blur.}
    \label{fig:compare-imagenetc-vcr-motion-blur}

\end{figure}

\begin{figure}
    \centering
    \begin{subfigure}{0.32\textwidth}
        \includegraphics[width=\textwidth]{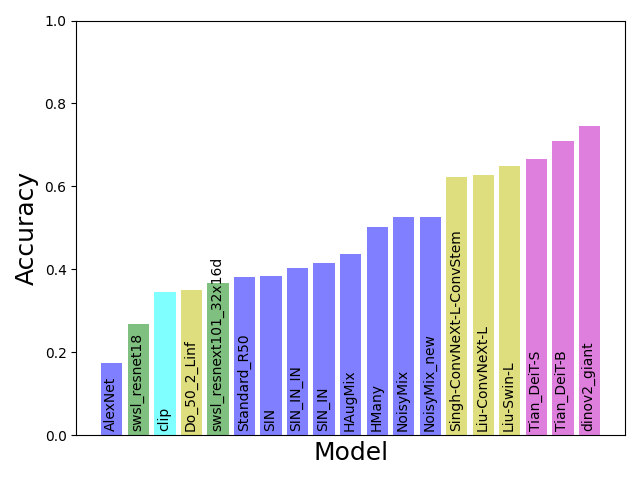}
        \caption{\scriptsize \textsc{ImageNet-C} Frost Accuracy}
        \label{fig:imagenetc-frost-acc-bar}
    \end{subfigure}
    \begin{subfigure}{0.32\textwidth}
        \includegraphics[width=\textwidth]{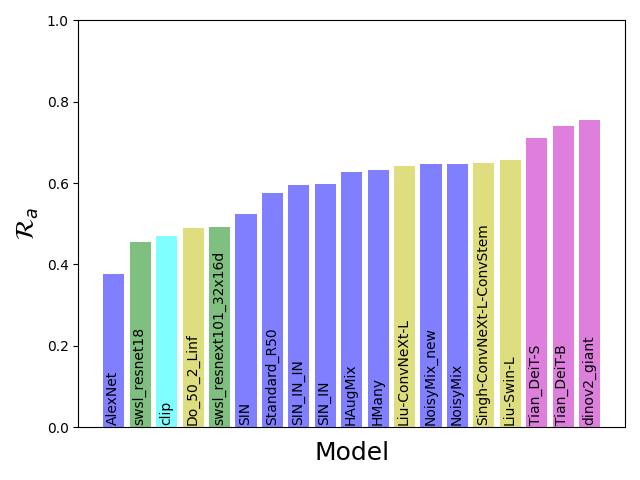}
        \caption{\scriptsize Frost $\hat{\mathcal{R}}_a$}
        \label{fig:vcr-frost-acc-bar}
    \end{subfigure}
    \begin{subfigure}{0.32\textwidth}
        \includegraphics[width=\textwidth]{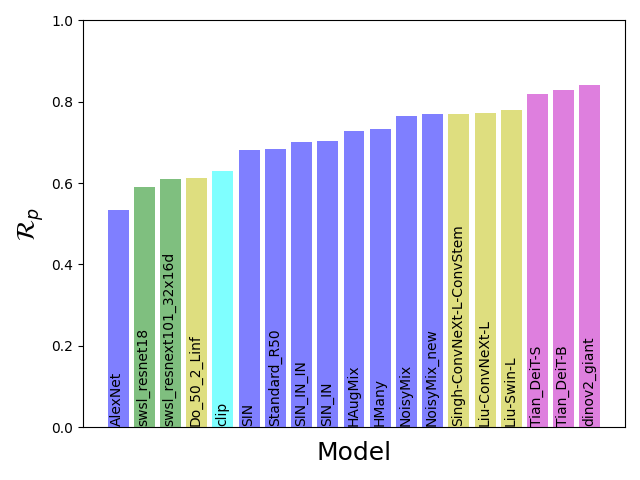}
        \caption{\scriptsize Frost $\hat{\mathcal{R}}_p$}
        \label{fig:vcr-frost-pred-sim-bar}
    \end{subfigure}
    \caption{Comparison between \textsc{ImageNet-C} and VCR with Frost.}
    \label{fig:compare-imagenetc-vcr-frost}

\end{figure}

\begin{figure}
    \centering
    \begin{subfigure}{0.32\textwidth}
        \includegraphics[width=\textwidth]{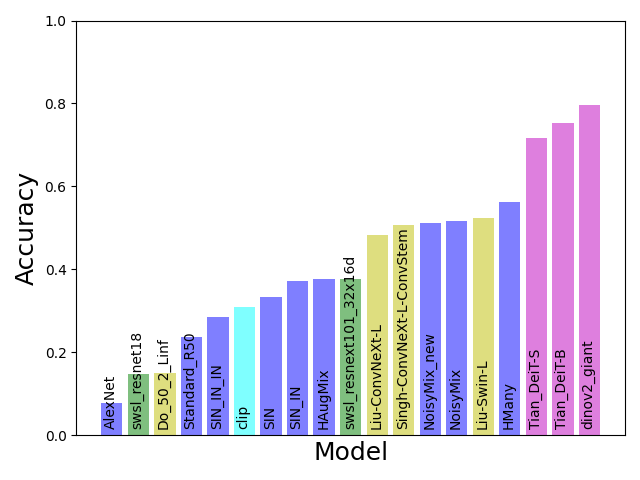}
        \caption{\scriptsize \textsc{ImageNet-C} Impulse Noise Accuracy}
        \label{fig:imagenetc-impulse-noise-acc-bar}
    \end{subfigure}
    \begin{subfigure}{0.32\textwidth}
        \includegraphics[width=\textwidth]{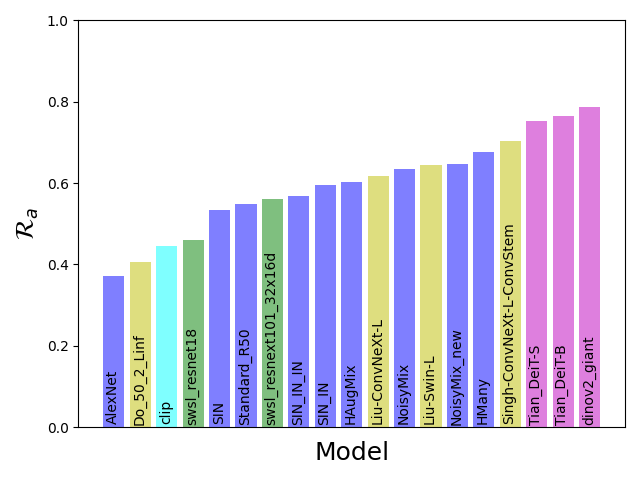}
        \caption{\scriptsize Impulse Noise $\hat{\mathcal{R}}_a$}
        \label{fig:vcr-impulse-noise-acc-bar}
    \end{subfigure}
    \begin{subfigure}{0.32\textwidth}
        \includegraphics[width=\textwidth]{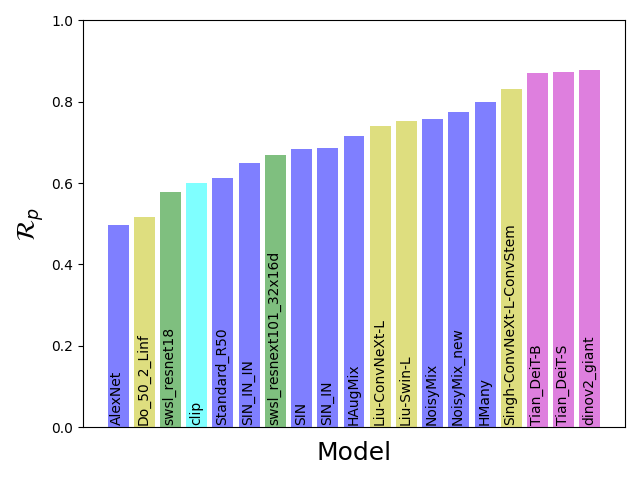}
        \caption{\scriptsize Impulse Noise $\hat{\mathcal{R}}_p$}
        \label{fig:vcr-impulse-noise-pred-sim-bar}
    \end{subfigure}
    \caption{Comparison between \textsc{ImageNet-C} and VCR with Impulse Noise.}
    \label{fig:compare-imagenetc-vcr-impulse-noise}

\end{figure}

\begin{figure}[t]
    \centering
    \begin{tabular}{p{0.32\textwidth} p{0.32\textwidth} p{0.32\textwidth}}
    
      \begin{subfigure}{0.32\textwidth}
        \includegraphics[width=\textwidth]{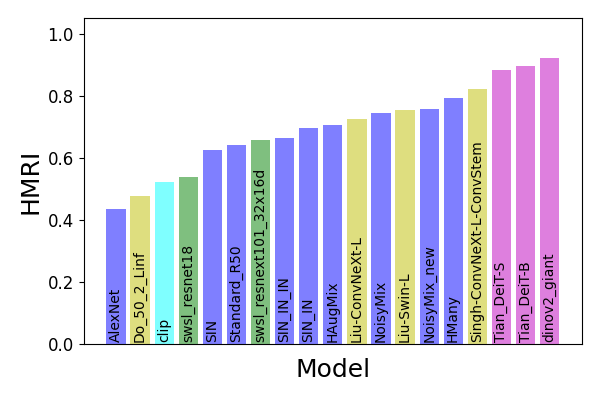}
        \caption{\scriptsize \textit{HMRI} for $\robustnesssymbol_a$ }
        \label{fig:impulse-noise-HMRI-acc}
    \end{subfigure}&
    \begin{subfigure}{0.32\textwidth}
        \includegraphics[width=\textwidth]{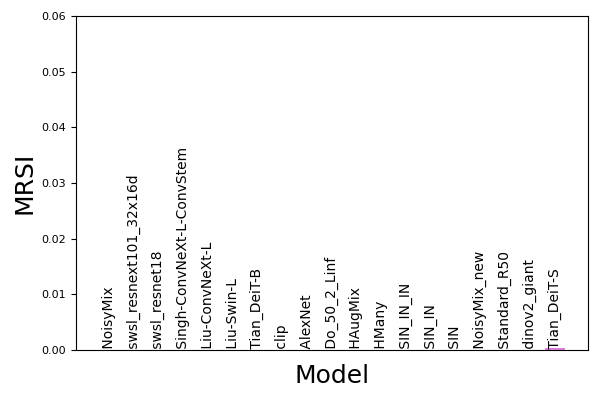}
        \caption{\scriptsize \textit{MRSI} for $\robustnesssymbol_a$}
        \label{fig:impulse-noise-MRSI-acc}
    \end{subfigure} &
    \begin{subfigure}{0.32\textwidth}
        \includegraphics[width=\textwidth]{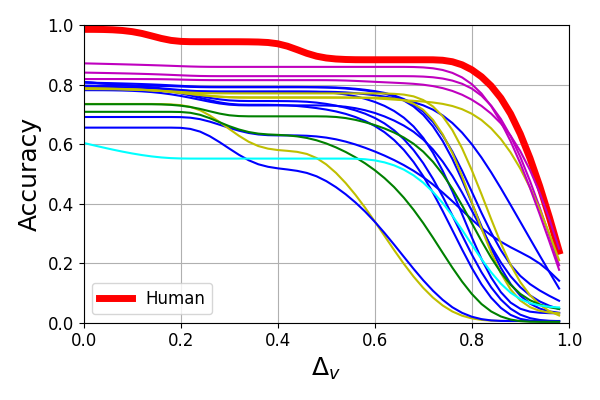}
        \caption{Estimated curves $s_a$}
        \label{fig:impulse-noise-vcr-acc-comparison}    
    \end{subfigure}
    \\
   \begin{subfigure}{0.32\textwidth}
        \includegraphics[width=\textwidth]{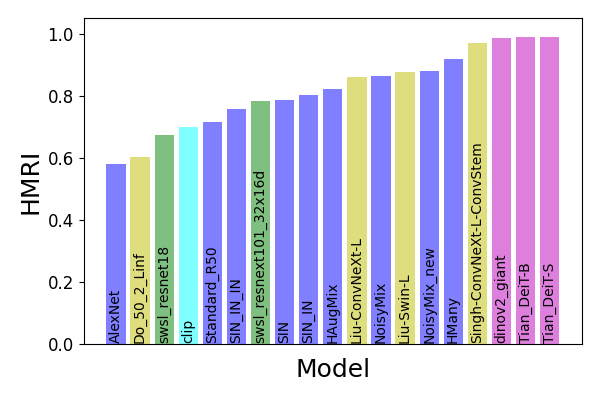}
        \caption{\scriptsize \textit{HMRI} for $\robustnesssymbol_p$}
        \label{fig:impulse-noise-HMRI-pred}
    \end{subfigure} &
    \begin{subfigure}{0.32\textwidth}
        \includegraphics[width=\textwidth]{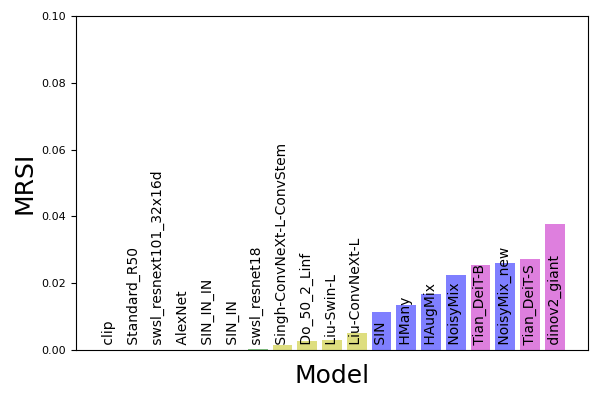}
        \caption{\scriptsize \textit{MRSI} for $\robustnesssymbol_p$}
        \label{fig:impulse-noise-MRSI-pred}
    \end{subfigure}&
\begin{subfigure}{0.32\textwidth}
        \includegraphics[width=\textwidth]{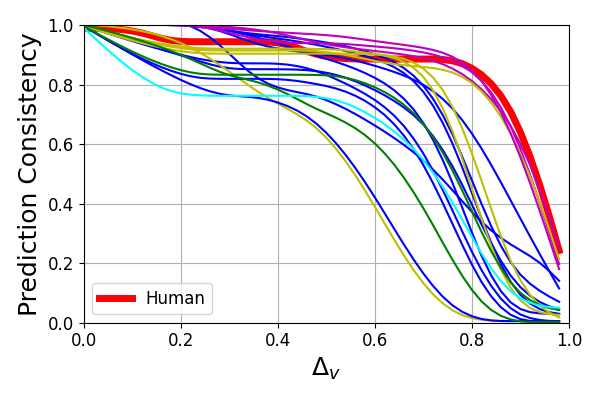}
        \caption{Estimated curves $s_p$}
        \label{fig:impulse-noise-vcr-pred-sim-comparison}    
    \end{subfigure}
    
    \\
    \end{tabular}
    \caption{VCR evaluation results for Impulse Noise.}
    \label{fig:hmri-mrsi-bar-plot_3}
\end{figure}

\begin{figure}[t]
    \centering
    \begin{tabular}{p{0.32\textwidth} p{0.32\textwidth} p{0.32\textwidth}}
    
      \begin{subfigure}{0.32\textwidth}
        \includegraphics[width=\textwidth]{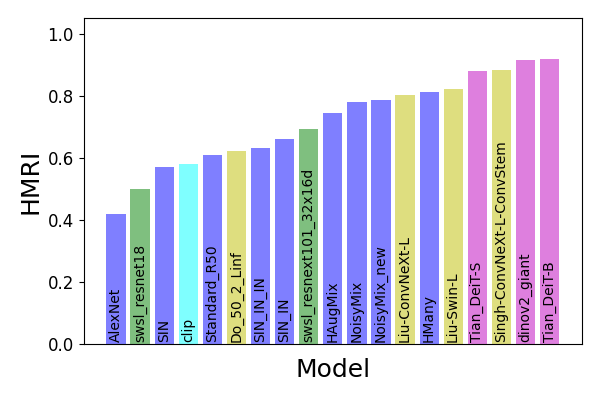}
        \caption{\scriptsize \textit{HMRI} for $\robustnesssymbol_a$ }
        \label{fig:shot-noise-HMRI-acc}
    \end{subfigure}&
    \begin{subfigure}{0.32\textwidth}
        \includegraphics[width=\textwidth]{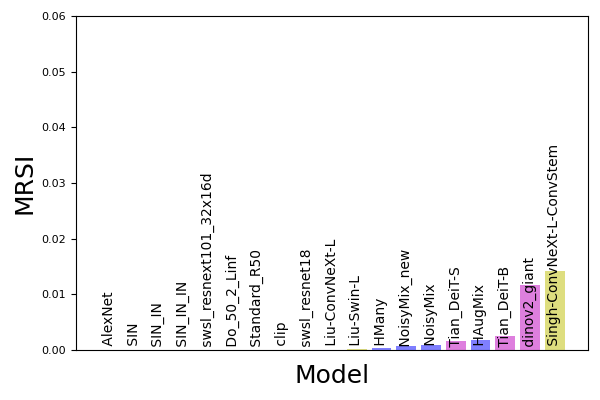}
        \caption{\scriptsize \textit{MRSI} for $\robustnesssymbol_a$}
        \label{fig:shot-noise-MRSI-acc}
    \end{subfigure} &
    \begin{subfigure}{0.32\textwidth}
        \includegraphics[width=\textwidth]{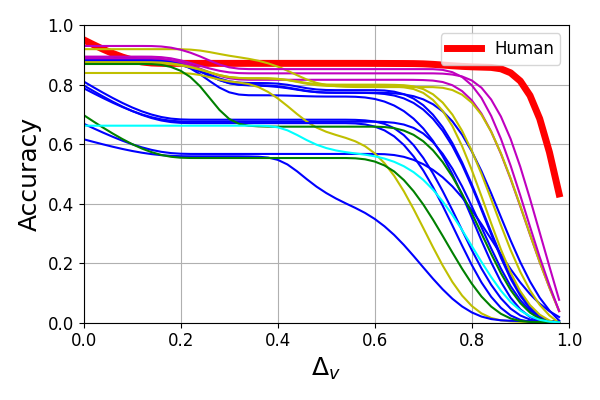}
        \caption{Estimated curves $s_a$}
        \label{fig:shot-noise-vcr-acc-comparison}    
    \end{subfigure}
    \\
   \begin{subfigure}{0.32\textwidth}
        \includegraphics[width=\textwidth]{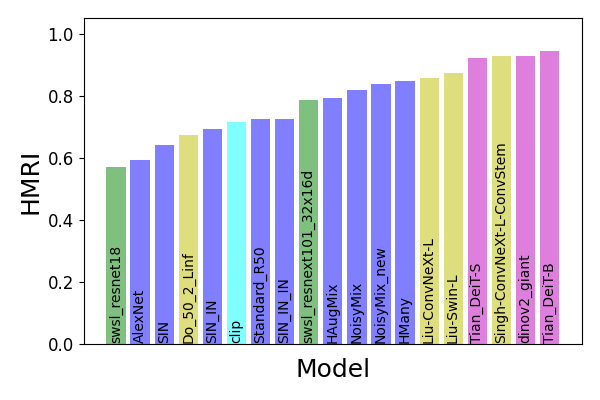}
        \caption{\scriptsize \textit{HMRI} for $\robustnesssymbol_p$}
        \label{fig:shot-noise-HMRI-pred}
    \end{subfigure} &
    \begin{subfigure}{0.32\textwidth}
        \includegraphics[width=\textwidth]{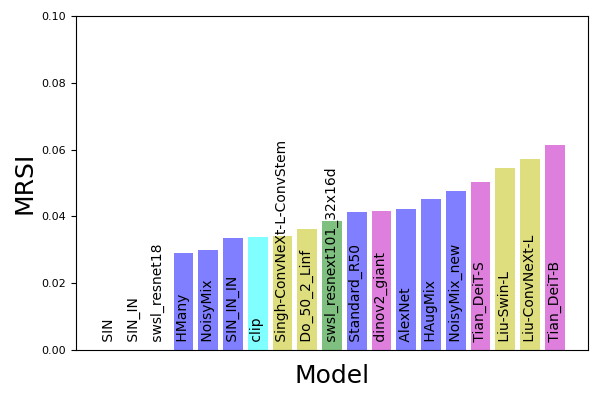}
        \caption{\scriptsize \textit{MRSI} for $\robustnesssymbol_p$}
        \label{fig:shot-noise-MRSI-pred}
    \end{subfigure}&
\begin{subfigure}{0.32\textwidth}
        \includegraphics[width=\textwidth]{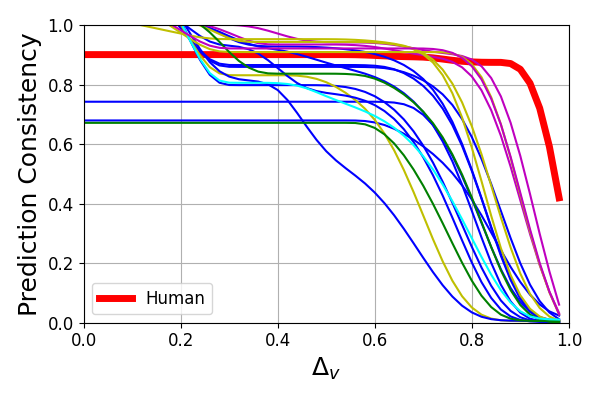}
        \caption{Estimated curves $s_p$}
        \label{fig:shot-noise-vcr-pred-sim-comparison}    
    \end{subfigure}
    
    \\
    \end{tabular}
    \caption{VCR evaluation results for Shot Noise.}
    \label{fig:hmri-mrsi-bar-plot_4}
\end{figure}

\begin{figure}[t]
    \centering
    \begin{tabular}{p{0.32\textwidth} p{0.32\textwidth} p{0.32\textwidth}}
    
      \begin{subfigure}{0.32\textwidth}
        \includegraphics[width=\textwidth]{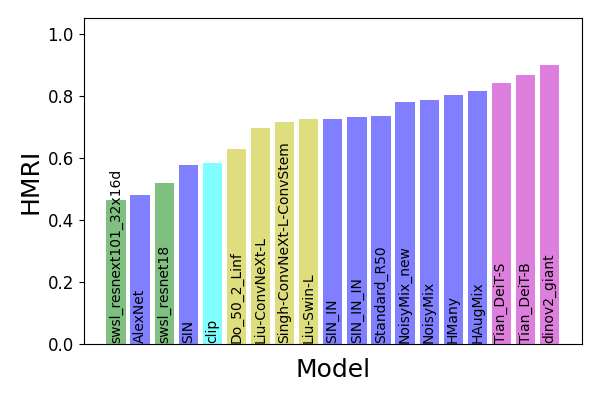}
        \caption{\scriptsize \textit{HMRI} for $\robustnesssymbol_a$ }
        \label{fig:blur-wrap-HMRI-acc}
    \end{subfigure}&
    \begin{subfigure}{0.32\textwidth}
        \includegraphics[width=\textwidth]{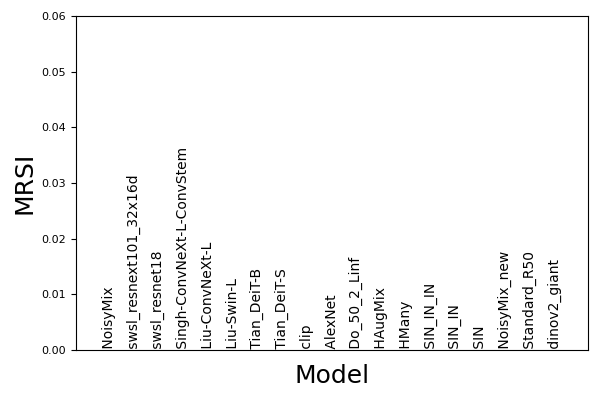}
        \caption{\scriptsize \textit{MRSI} for $\robustnesssymbol_a$}
        \label{fig:blur-wrap-MRSI-acc}
    \end{subfigure} &
    \begin{subfigure}{0.32\textwidth}
        \includegraphics[width=\textwidth]{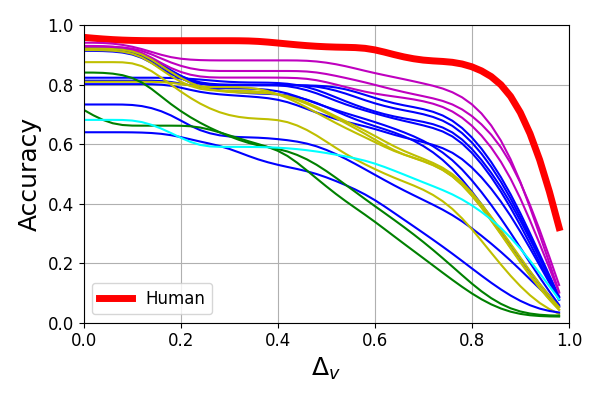}
        \caption{Estimated curves $s_a$}
        \label{fig:blur-wrap-vcr-acc-comparison}    
    \end{subfigure}
    \\
   \begin{subfigure}{0.32\textwidth}
        \includegraphics[width=\textwidth]{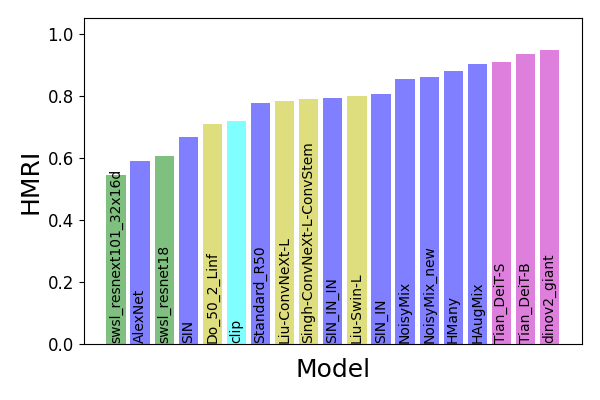}
        \caption{\scriptsize \textit{HMRI} for $\robustnesssymbol_p$}
        \label{fig:blur-wrap-HMRI-pred}
    \end{subfigure} &
    \begin{subfigure}{0.32\textwidth}
        \includegraphics[width=\textwidth]{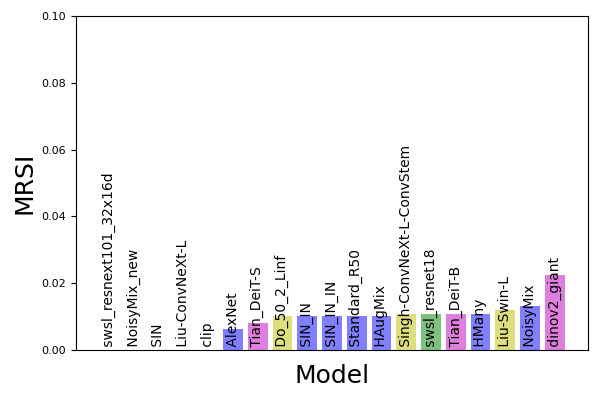}
        \caption{\scriptsize \textit{MRSI} for $\robustnesssymbol_p$}
        \label{fig:blur-wrap-MRSI-pred}
    \end{subfigure}&
\begin{subfigure}{0.32\textwidth}
        \includegraphics[width=\textwidth]{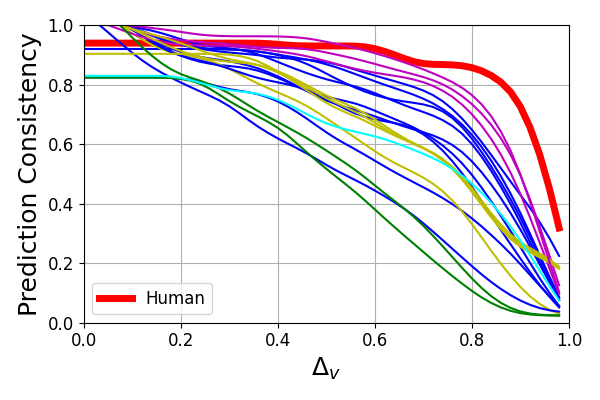}
        \caption{Estimated curves $s_p$}
        \label{fig:blur-wrap-vcr-pred-sim-comparison}    
    \end{subfigure}
    
    \\
    \end{tabular}
    \caption{VCR evaluation results for Blur.}
    \label{fig:hmri-mrsi-bar-plot_5}
\end{figure}

\begin{figure}[t]
    \centering
    \begin{tabular}{p{0.32\textwidth} p{0.32\textwidth} p{0.32\textwidth}}
    
      \begin{subfigure}{0.32\textwidth}
        \includegraphics[width=\textwidth]{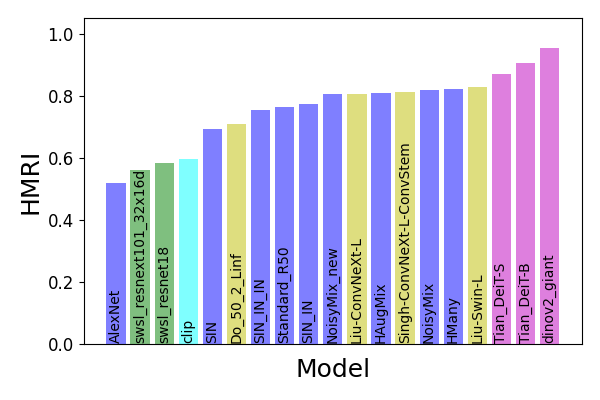}
        \caption{\scriptsize \textit{HMRI} for $\robustnesssymbol_a$ }
        \label{fig:median-blur-HMRI-acc}
    \end{subfigure}&
    \begin{subfigure}{0.32\textwidth}
        \includegraphics[width=\textwidth]{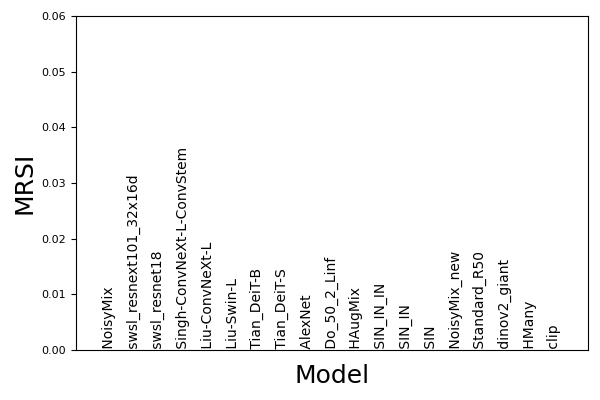}
        \caption{\scriptsize \textit{MRSI} for $\robustnesssymbol_a$}
        \label{fig:median-blur-MRSI-acc}
    \end{subfigure} &
    \begin{subfigure}{0.32\textwidth}
        \includegraphics[width=\textwidth]{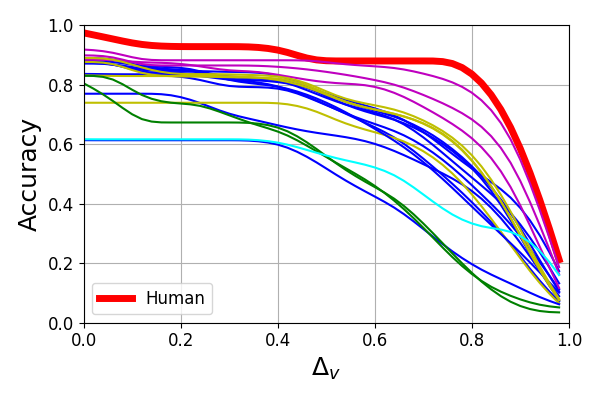}
        \caption{Estimated curves $s_a$}
        \label{fig:median-blur-vcr-acc-comparison}    
    \end{subfigure}
    \\
   \begin{subfigure}{0.32\textwidth}
        \includegraphics[width=\textwidth]{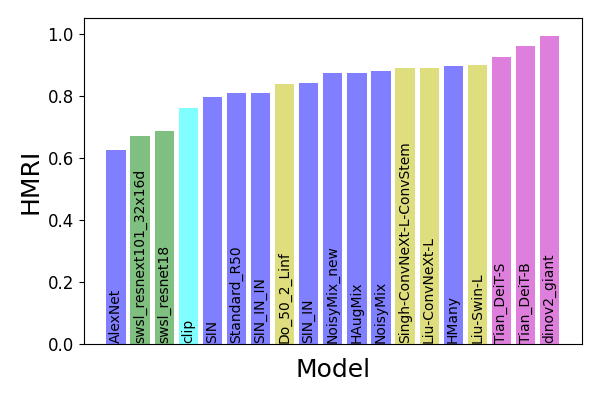}
        \caption{\scriptsize \textit{HMRI} for $\robustnesssymbol_p$}
        \label{fig:median-blur-HMRI-pred}
    \end{subfigure} &
    \begin{subfigure}{0.32\textwidth}
        \includegraphics[width=\textwidth]{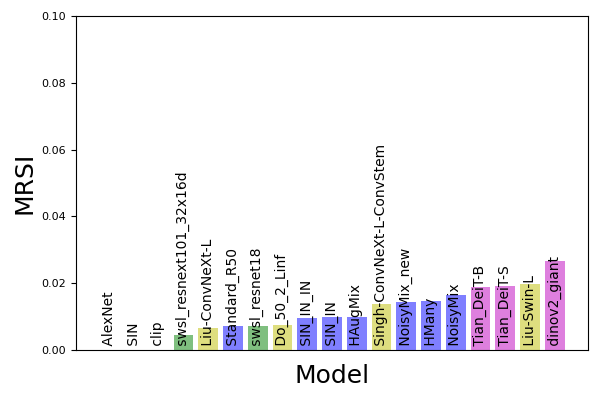}
        \caption{\scriptsize \textit{MRSI} for $\robustnesssymbol_p$}
        \label{fig:median-blur-MRSI-pred}
    \end{subfigure}&
\begin{subfigure}{0.32\textwidth}
        \includegraphics[width=\textwidth]{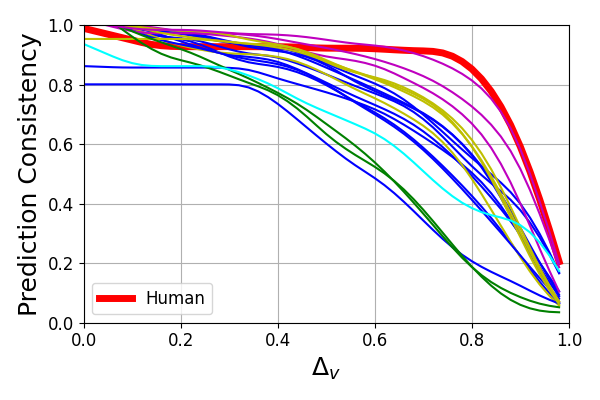}
        \caption{Estimated curves $s_p$}
        \label{fig:median-blur-vcr-pred-sim-comparison}    
    \end{subfigure}
    
    \\
    \end{tabular}
    \caption{VCR evaluation results for Median Blur.}
    \label{fig:hmri-mrsi-bar-plot_6}
\end{figure}

\begin{figure}[t]
    \centering
    \begin{tabular}{p{0.32\textwidth} p{0.32\textwidth} p{0.32\textwidth}}
    
      \begin{subfigure}{0.32\textwidth}
        \includegraphics[width=\textwidth]{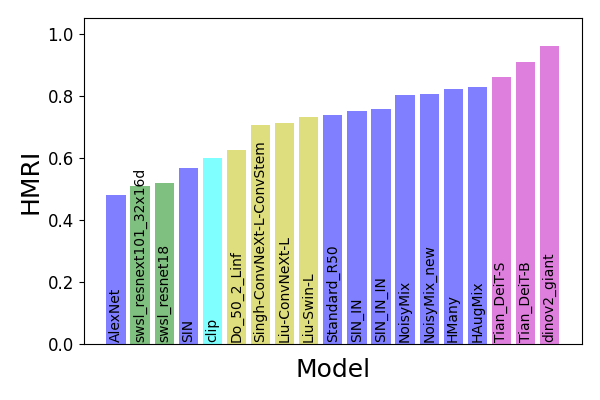}
        \caption{\scriptsize \textit{HMRI} for $\robustnesssymbol_a$ }
        \label{fig:glass-blur-HMRI-acc}
    \end{subfigure}&
    \begin{subfigure}{0.32\textwidth}
        \includegraphics[width=\textwidth]{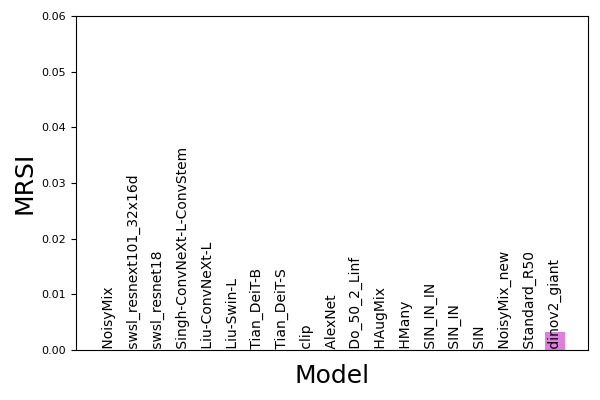}
        \caption{\scriptsize \textit{MRSI} for $\robustnesssymbol_a$}
        \label{fig:glass-blur-MRSI-acc}
    \end{subfigure} &
    \begin{subfigure}{0.32\textwidth}
        \includegraphics[width=\textwidth]{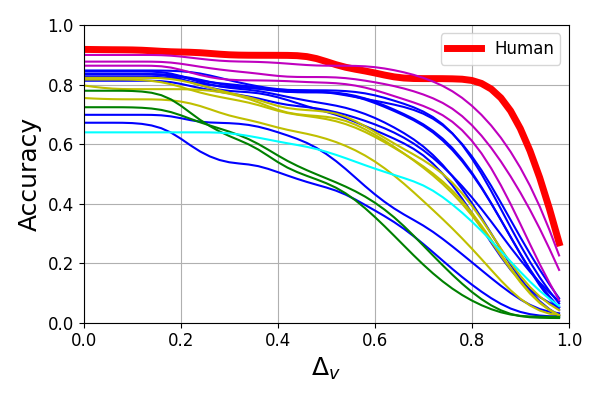}
        \caption{Estimated curves $s_a$}
        \label{fig:glass-blur-vcr-acc-comparison}    
    \end{subfigure}
    \\
   \begin{subfigure}{0.32\textwidth}
        \includegraphics[width=\textwidth]{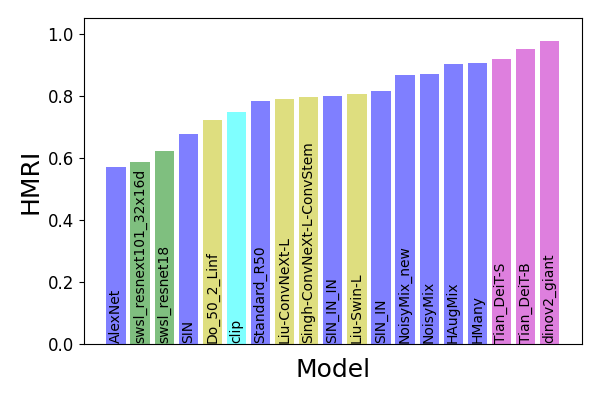}
        \caption{\scriptsize \textit{HMRI} for $\robustnesssymbol_p$}
        \label{fig:glass-blur-HMRI-pred}
    \end{subfigure} &
    \begin{subfigure}{0.32\textwidth}
        \includegraphics[width=\textwidth]{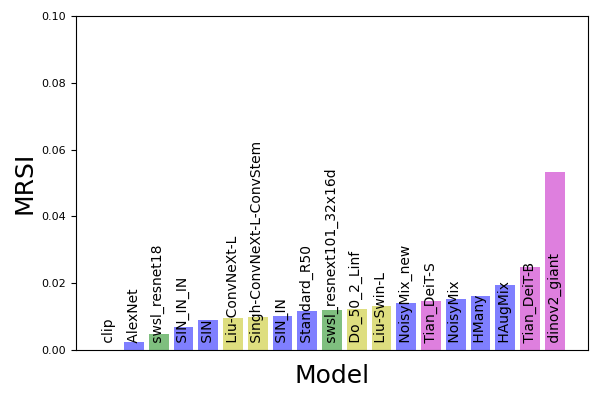}
        \caption{\scriptsize \textit{MRSI} for $\robustnesssymbol_p$}
        \label{fig:glass-blur-MRSI-pred}
    \end{subfigure}&
\begin{subfigure}{0.32\textwidth}
        \includegraphics[width=\textwidth]{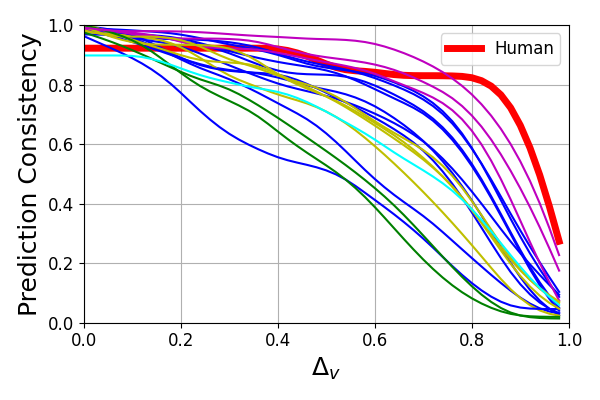}
        \caption{Estimated curves $s_p$}
        \label{fig:glass-blur-vcr-pred-sim-comparison}    
    \end{subfigure}
    
    \\
    \end{tabular}
    \caption{VCR evaluation results for Glass Blur.}
    \label{fig:hmri-mrsi-bar-plot_7}
\end{figure}

\begin{figure}[t]
    \centering
    \begin{tabular}{p{0.32\textwidth} p{0.32\textwidth} p{0.32\textwidth}}
    
      \begin{subfigure}{0.32\textwidth}
        \includegraphics[width=\textwidth]{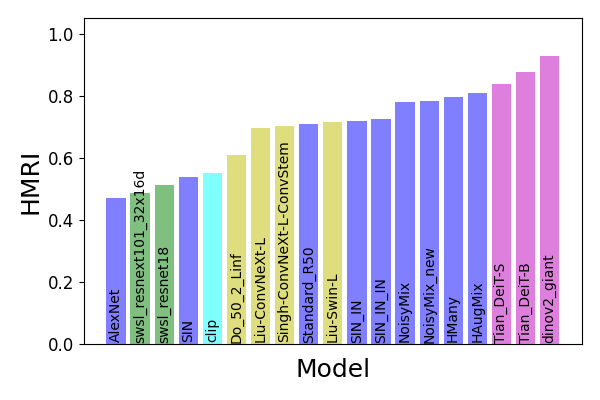}
        \caption{\scriptsize \textit{HMRI} for $\robustnesssymbol_a$ }
        \label{fig:gaussian-blur-HMRI-acc}
    \end{subfigure}&
    \begin{subfigure}{0.32\textwidth}
        \includegraphics[width=\textwidth]{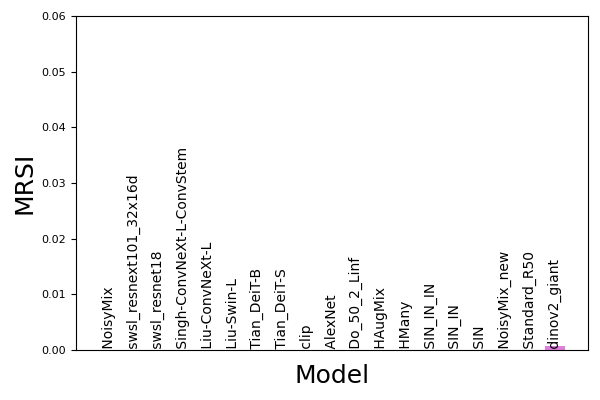}
        \caption{\scriptsize \textit{MRSI} for $\robustnesssymbol_a$}
        \label{fig:gaussian-blur-MRSI-acc}
    \end{subfigure} &
    \begin{subfigure}{0.32\textwidth}
        \includegraphics[width=\textwidth]{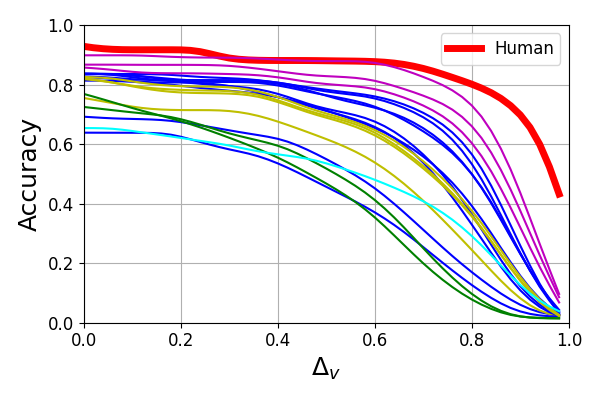}
        \caption{Estimated curves $s_a$}
        \label{fig:gaussian-blur-vcr-acc-comparison}    
    \end{subfigure}
    \\
   \begin{subfigure}{0.32\textwidth}
        \includegraphics[width=\textwidth]{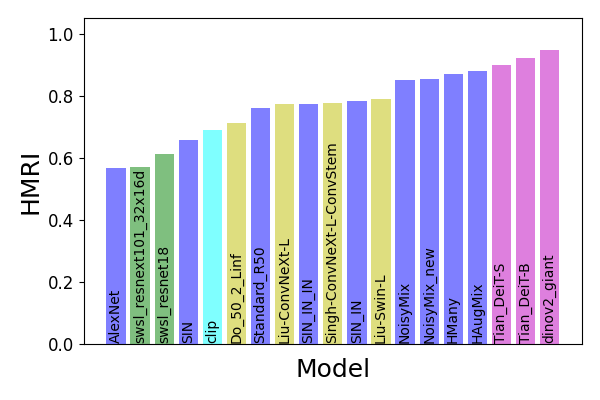}
        \caption{\scriptsize \textit{HMRI} for $\robustnesssymbol_p$}
        \label{fig:gaussian-blur-HMRI-pred}
    \end{subfigure} &
    \begin{subfigure}{0.32\textwidth}
        \includegraphics[width=\textwidth]{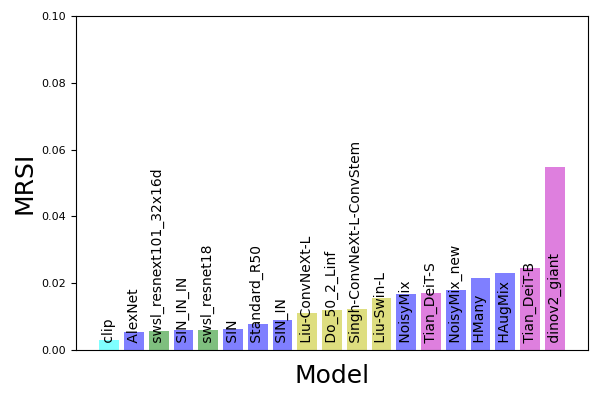}
        \caption{\scriptsize \textit{MRSI} for $\robustnesssymbol_p$}
        \label{fig:gaussian-blur-MRSI-pred}
    \end{subfigure}&
\begin{subfigure}{0.32\textwidth}
        \includegraphics[width=\textwidth]{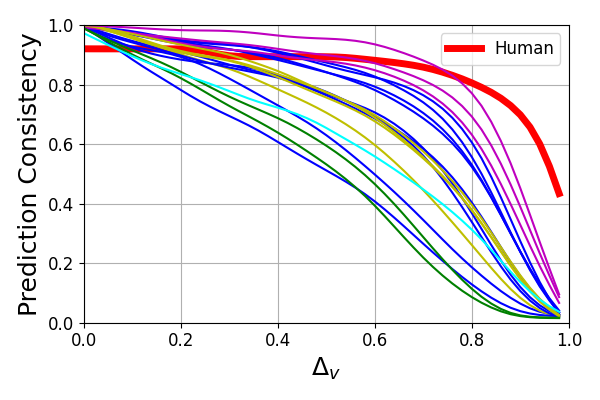}
        \caption{Estimated curves $s_p$}
        \label{fig:gaussian-blur-vcr-pred-sim-comparison}    
    \end{subfigure}
    
    \\
    \end{tabular}
    \caption{VCR evaluation results for Gaussian Blur.}
    \label{fig:hmri-mrsi-bar-plot_8}
\end{figure}

\begin{figure}[t]
    \centering
    \begin{tabular}{p{0.32\textwidth} p{0.32\textwidth} p{0.32\textwidth}}
    
      \begin{subfigure}{0.32\textwidth}
        \includegraphics[width=\textwidth]{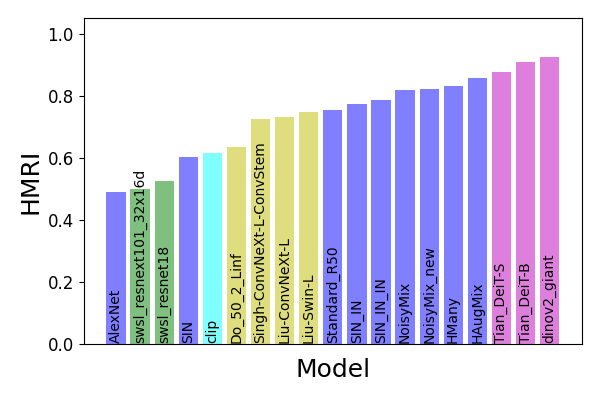}
        \caption{\scriptsize \textit{HMRI} for $\robustnesssymbol_a$ }
        \label{fig:defocus-blur-HMRI-acc}
    \end{subfigure}&
    \begin{subfigure}{0.32\textwidth}
        \includegraphics[width=\textwidth]{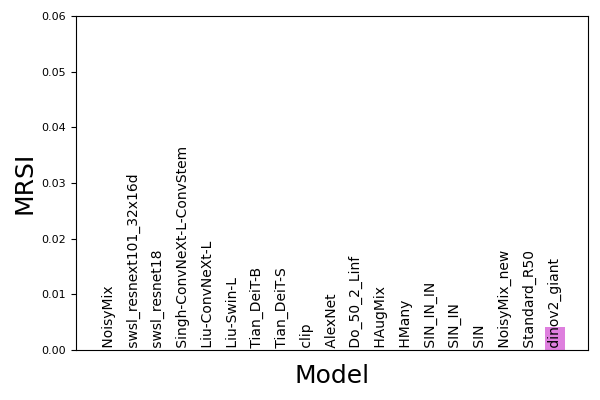}
        \caption{\scriptsize \textit{MRSI} for $\robustnesssymbol_a$}
        \label{fig:defocus-blur-MRSI-acc}
    \end{subfigure} &
    \begin{subfigure}{0.32\textwidth}
        \includegraphics[width=\textwidth]{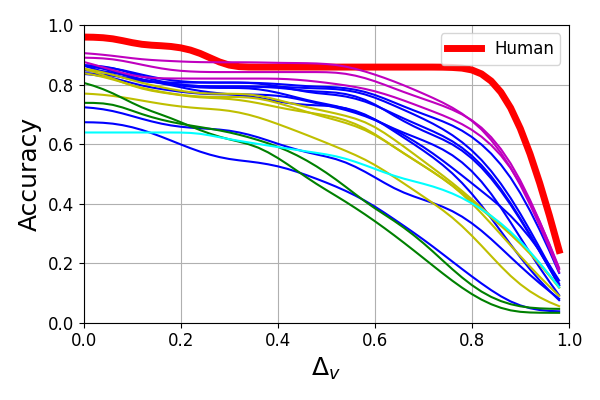}
        \caption{Estimated curves $s_a$}
        \label{fig:defocus-blur-vcr-acc-comparison}    
    \end{subfigure}
    \\
   \begin{subfigure}{0.32\textwidth}
        \includegraphics[width=\textwidth]{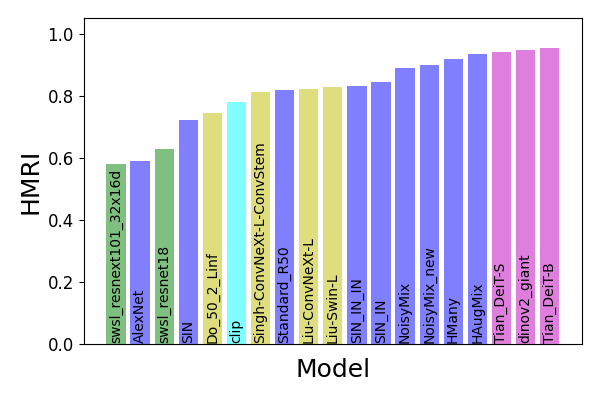}
        \caption{\scriptsize \textit{HMRI} for $\robustnesssymbol_p$}
        \label{fig:defocus-blur-HMRI-pred}
    \end{subfigure} &
    \begin{subfigure}{0.32\textwidth}
        \includegraphics[width=\textwidth]{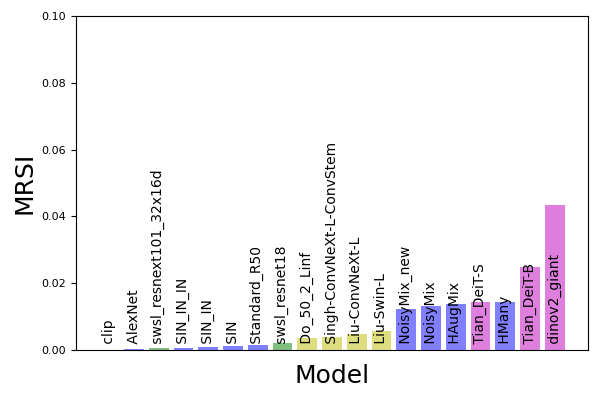}
        \caption{\scriptsize \textit{MRSI} for $\robustnesssymbol_p$}
        \label{fig:defocus-blur-MRSI-pred}
    \end{subfigure}&
\begin{subfigure}{0.32\textwidth}
        \includegraphics[width=\textwidth]{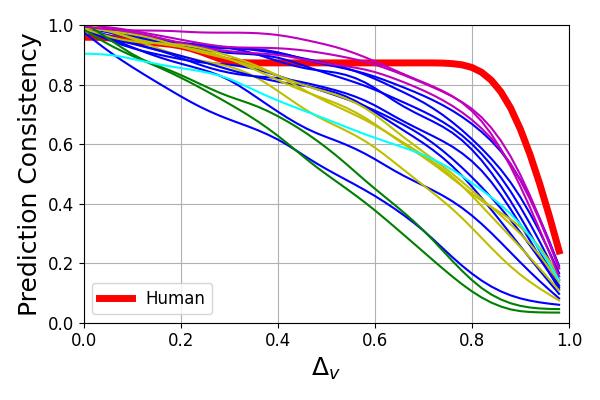}
        \caption{Estimated curves $s_p$}
        \label{fig:defocus-blur-vcr-pred-sim-comparison}    
    \end{subfigure}
    
    \\
    \end{tabular}
    \caption{VCR evaluation results for Defocus Blur.}
    \label{fig:hmri-mrsi-bar-plot_9}
\end{figure}

\begin{figure}[t]
    \centering
    \begin{tabular}{p{0.32\textwidth} p{0.32\textwidth} p{0.32\textwidth}}
    
      \begin{subfigure}{0.32\textwidth}
        \includegraphics[width=\textwidth]{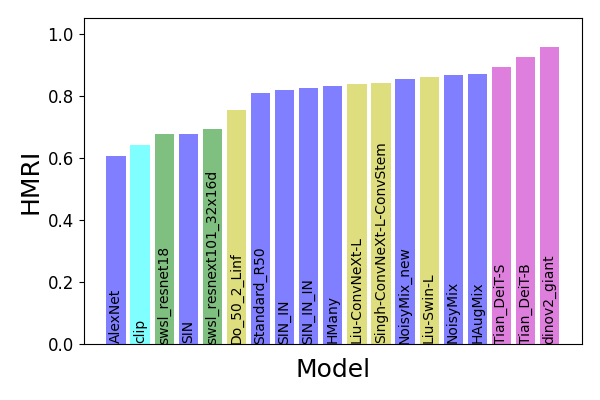}
        \caption{\scriptsize \textit{HMRI} for $\robustnesssymbol_a$ }
        \label{fig:motion-blur-HMRI-acc}
    \end{subfigure}&
    \begin{subfigure}{0.32\textwidth}
        \includegraphics[width=\textwidth]{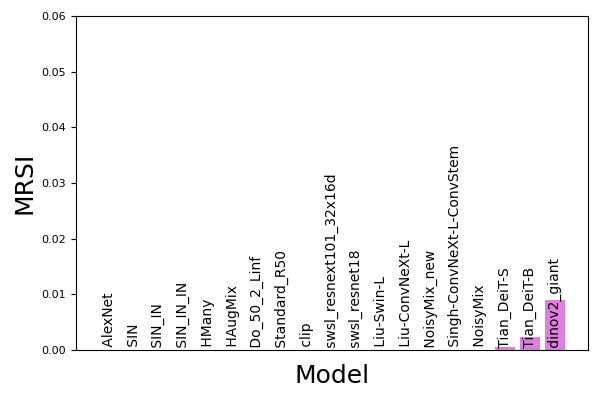}
        \caption{\scriptsize \textit{MRSI} for $\robustnesssymbol_a$}
        \label{fig:motion-blur-MRSI-acc}
    \end{subfigure} &
    \begin{subfigure}{0.32\textwidth}
        \includegraphics[width=\textwidth]{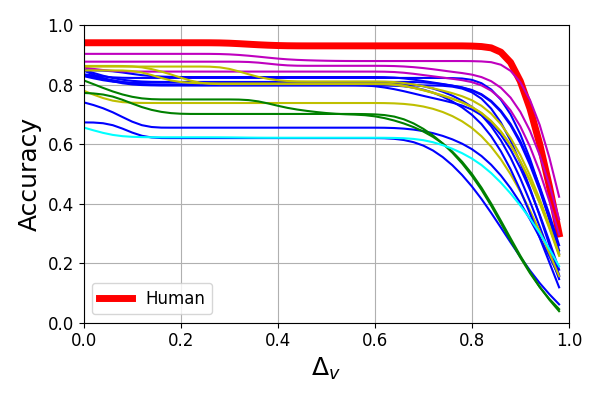}
        \caption{Estimated curves $s_a$}
        \label{fig:motion-blur-vcr-acc-comparison}    
    \end{subfigure}
    \\
   \begin{subfigure}{0.32\textwidth}
        \includegraphics[width=\textwidth]{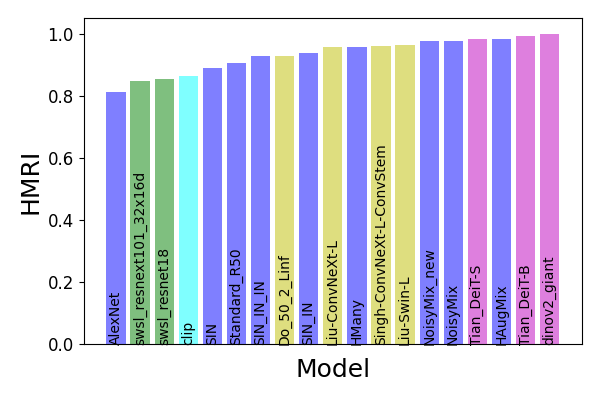}
        \caption{\scriptsize \textit{HMRI} for $\robustnesssymbol_p$}
        \label{fig:motion-blur-HMRI-pred}
    \end{subfigure} &
    \begin{subfigure}{0.32\textwidth}
        \includegraphics[width=\textwidth]{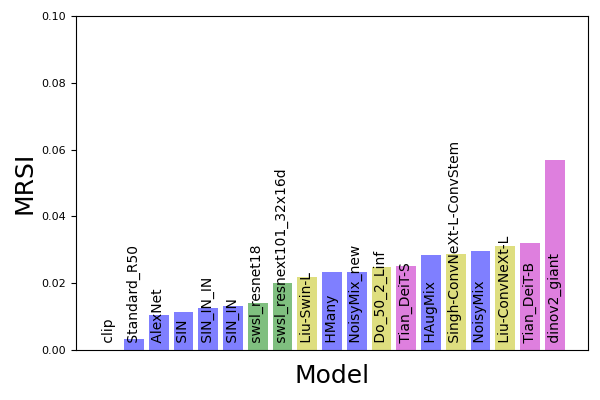}
        \caption{\scriptsize \textit{MRSI} for $\robustnesssymbol_p$}
        \label{fig:motion-blur-MRSI-pred}
    \end{subfigure}&
\begin{subfigure}{0.32\textwidth}
        \includegraphics[width=\textwidth]{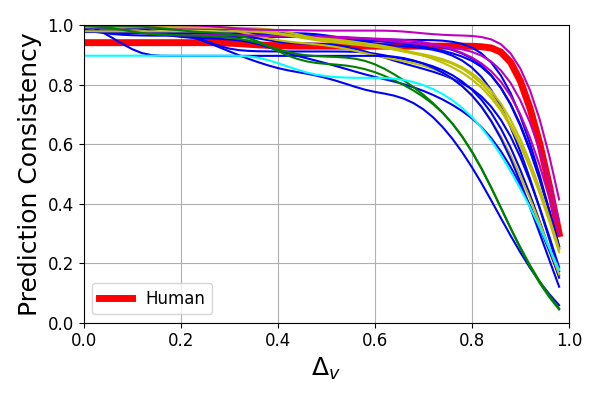}
        \caption{Estimated curves $s_p$}
        \label{fig:motion-blur-vcr-pred-sim-comparison}    
    \end{subfigure}
    
    \\
    \end{tabular}
    \caption{VCR evaluation results for Motion Blur. }
    \label{fig:hmri-mrsi-bar-plot_10}
\end{figure}

\begin{figure}[t]
    \centering
    \begin{tabular}{p{0.32\textwidth} p{0.32\textwidth} p{0.32\textwidth}}
    
      \begin{subfigure}{0.32\textwidth}
        \includegraphics[width=\textwidth]{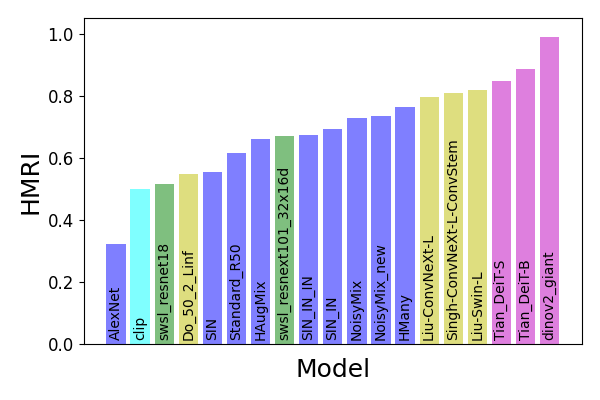}
        \caption{\scriptsize \textit{HMRI} for $\robustnesssymbol_a$ }
        \label{fig:hue-saturation-HMRI-acc}
    \end{subfigure}&
    \begin{subfigure}{0.32\textwidth}
        \includegraphics[width=\textwidth]{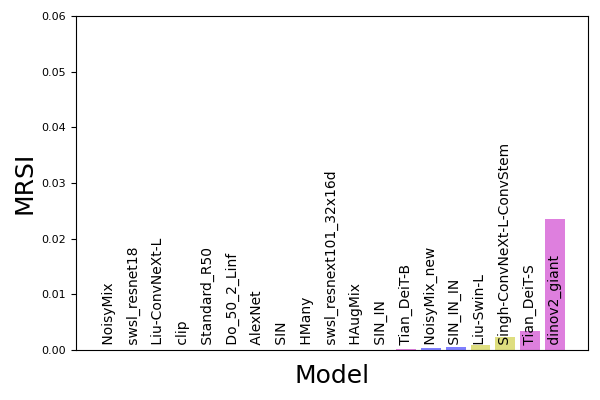}
        \caption{\scriptsize \textit{MRSI} for $\robustnesssymbol_a$}
        \label{fig:hue-saturation-MRSI-acc}
    \end{subfigure} &
    \begin{subfigure}{0.32\textwidth}
        \includegraphics[width=\textwidth]{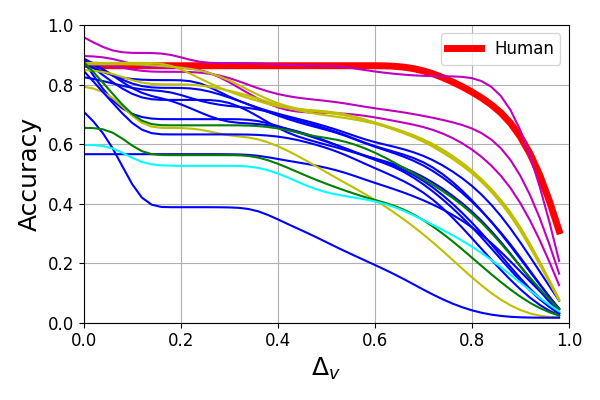}
        \caption{Estimated curves $s_a$}
        \label{fig:hue-saturation-vcr-acc-comparison}    
    \end{subfigure}
    \\
   \begin{subfigure}{0.32\textwidth}
        \includegraphics[width=\textwidth]{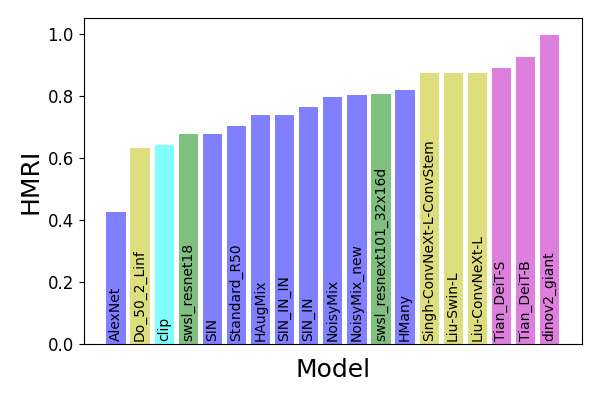}
        \caption{\scriptsize \textit{HMRI} for $\robustnesssymbol_p$}
        \label{fig:hue-saturation-HMRI-pred}
    \end{subfigure} &
    \begin{subfigure}{0.32\textwidth}
        \includegraphics[width=\textwidth]{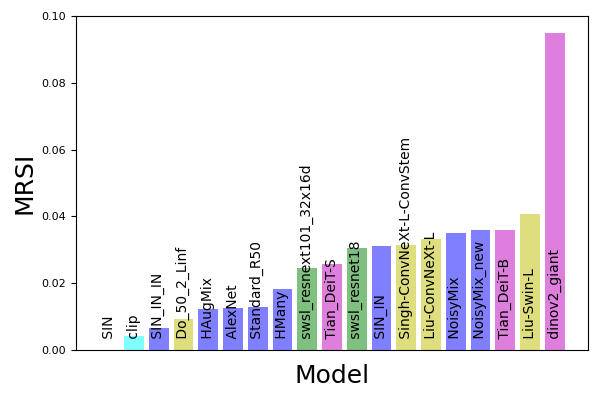}
        \caption{\scriptsize \textit{MRSI} for $\robustnesssymbol_p$}
        \label{fig:hue-saturation-MRSI-pred}
    \end{subfigure}&
\begin{subfigure}{0.32\textwidth}
        \includegraphics[width=\textwidth]{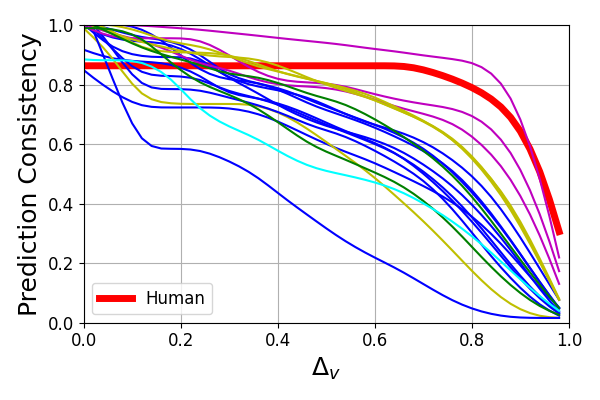}
        \caption{Estimated curves $s_p$}
        \label{fig:hue-saturation-vcr-pred-sim-comparison}    
    \end{subfigure}
    
    \\
    \end{tabular}
    \caption{VCR evaluation results for Hue Saturation Value.}
    \label{fig:hmri-mrsi-bar-plot_11}
\end{figure}

\begin{figure}[t]
    \centering
    \begin{tabular}{p{0.32\textwidth} p{0.32\textwidth} p{0.32\textwidth}}
    
      \begin{subfigure}{0.32\textwidth}
        \includegraphics[width=\textwidth]{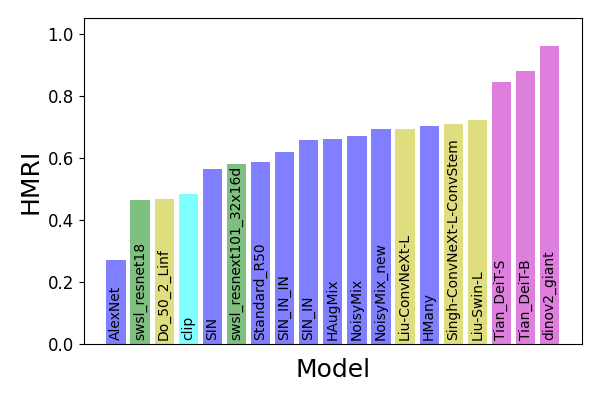}
        \caption{\scriptsize \textit{HMRI} for $\robustnesssymbol_a$ }
        \label{fig:color-jitter-HMRI-acc}
    \end{subfigure}&
    \begin{subfigure}{0.32\textwidth}
        \includegraphics[width=\textwidth]{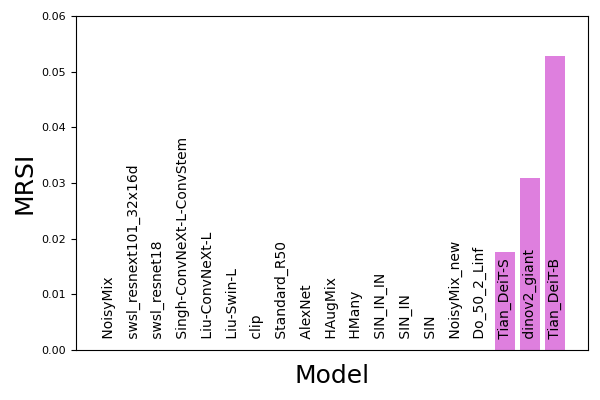}
        \caption{\scriptsize \textit{MRSI} for $\robustnesssymbol_a$}
        \label{fig:color-jitter-MRSI-acc}
    \end{subfigure} &
    \begin{subfigure}{0.32\textwidth}
        \includegraphics[width=\textwidth]{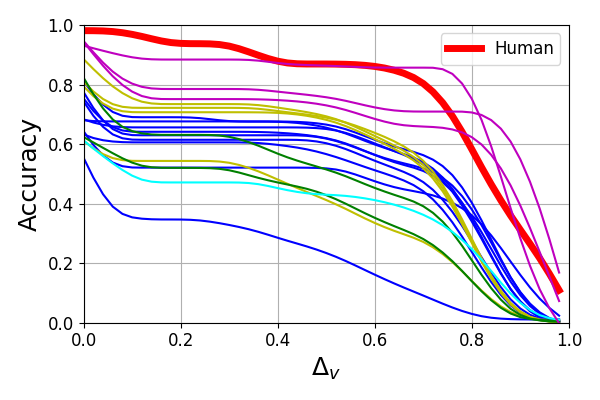}
        \caption{Estimated curves $s_a$}
        \label{fig:color-jitter-vcr-acc-comparison}    
    \end{subfigure}
    \\
   \begin{subfigure}{0.32\textwidth}
        \includegraphics[width=\textwidth]{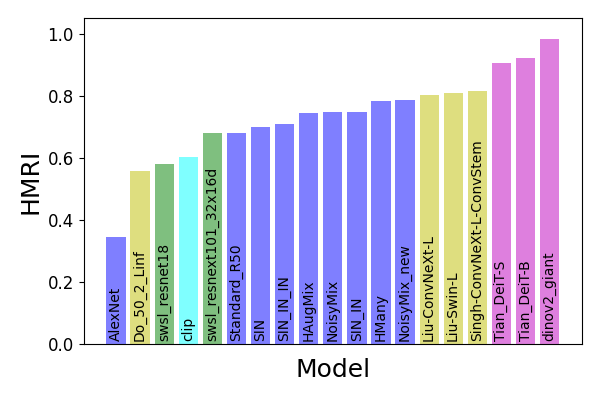}
        \caption{\scriptsize \textit{HMRI} for $\robustnesssymbol_p$}
        \label{fig:color-jitter-HMRI-pred}
    \end{subfigure} &
    \begin{subfigure}{0.32\textwidth}
        \includegraphics[width=\textwidth]{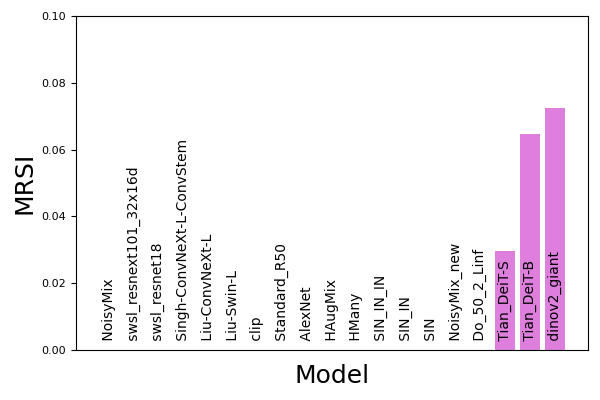}
        \caption{\scriptsize \textit{MRSI} for $\robustnesssymbol_p$}
        \label{fig:color-jitter-MRSI-pred}
    \end{subfigure}&
\begin{subfigure}{0.32\textwidth}
        \includegraphics[width=\textwidth]{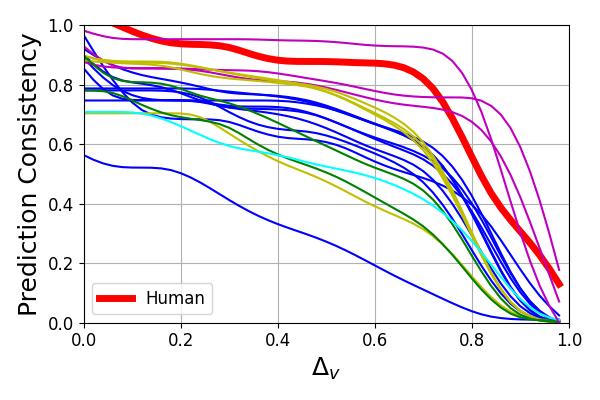}
        \caption{Estimated curves $s_p$}
        \label{fig:color-jitter-vcr-pred-sim-comparison}    
    \end{subfigure}
    
    \\
    \end{tabular}
    \caption{VCR evaluation results for Color Jitter.}
    \label{fig:hmri-mrsi-bar-plot_12}
\end{figure}

\begin{figure}[t]
    \centering
    \begin{tabular}{p{0.32\textwidth} p{0.32\textwidth} p{0.32\textwidth}}
    
      \begin{subfigure}{0.32\textwidth}
        \includegraphics[width=\textwidth]{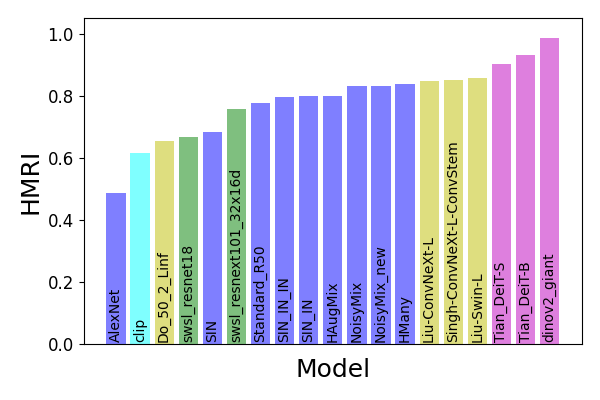}
        \caption{\scriptsize \textit{HMRI} for $\robustnesssymbol_a$ }
        \label{fig:brightness-HMRI-acc}
    \end{subfigure}&
    \begin{subfigure}{0.32\textwidth}
        \includegraphics[width=\textwidth]{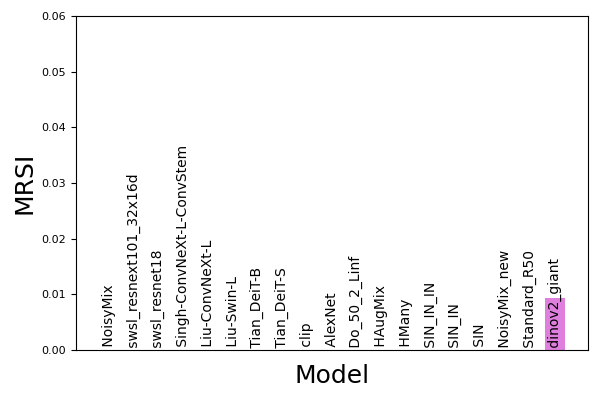}
        \caption{\scriptsize \textit{MRSI} for $\robustnesssymbol_a$}
        \label{fig:brightness-MRSI-acc}
    \end{subfigure} &
    \begin{subfigure}{0.32\textwidth}
        \includegraphics[width=\textwidth]{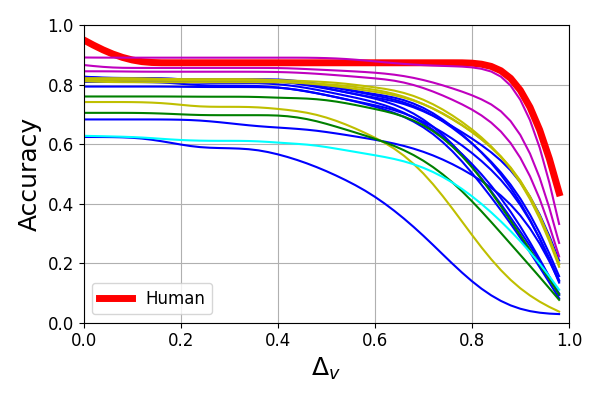}
        \caption{Estimated curves $s_a$}
        \label{fig:brightness-vcr-acc-comparison}    
    \end{subfigure}
    \\
   \begin{subfigure}{0.32\textwidth}
        \includegraphics[width=\textwidth]{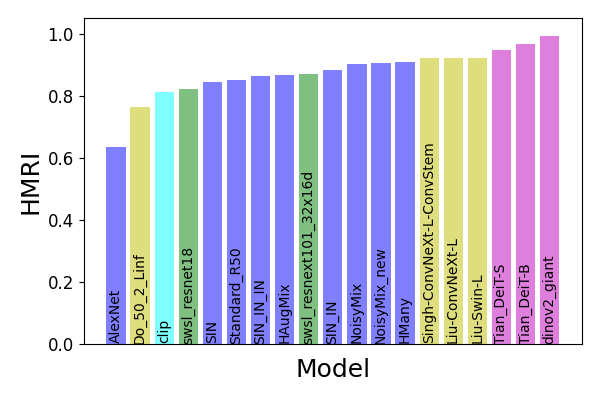}
        \caption{\scriptsize \textit{HMRI} for $\robustnesssymbol_p$}
        \label{fig:brightness-HMRI-pred}
    \end{subfigure} &
    \begin{subfigure}{0.32\textwidth}
        \includegraphics[width=\textwidth]{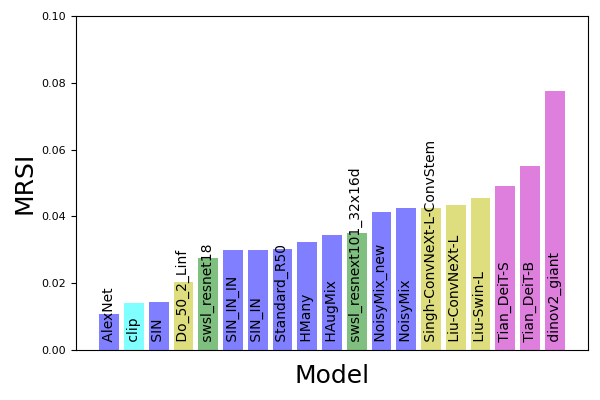}
        \caption{\scriptsize \textit{MRSI} for $\robustnesssymbol_p$}
        \label{fig:brightness-MRSI-pred}
    \end{subfigure}&
\begin{subfigure}{0.32\textwidth}
        \includegraphics[width=\textwidth]{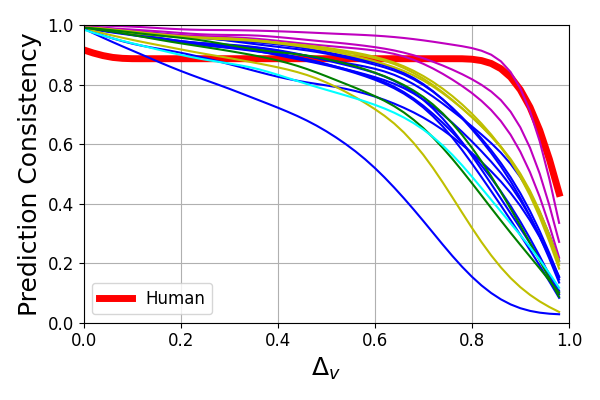}
        \caption{Estimated curves $s_p$}
        \label{fig:brightness-vcr-pred-sim-comparison}    
    \end{subfigure}
    
    \\
    \end{tabular}
    \caption{VCR evaluation results for Brightness.}
    \label{fig:hmri-mrsi-bar-plot_13}
\end{figure}

\begin{figure}[h]
    \centering
    \begin{tabular}{p{0.32\textwidth} p{0.32\textwidth} p{0.32\textwidth}}
    
      \begin{subfigure}{0.32\textwidth}
        \includegraphics[width=\textwidth]{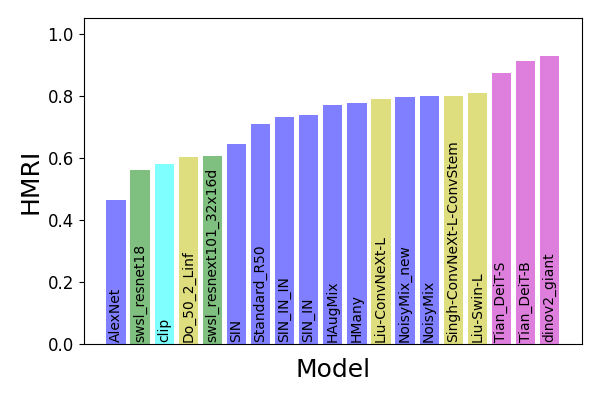}
        \caption{\scriptsize \textit{HMRI} for $\robustnesssymbol_a$ }
        \label{fig:frost-HMRI-acc}
    \end{subfigure}&
    \begin{subfigure}{0.32\textwidth}
        \includegraphics[width=\textwidth]{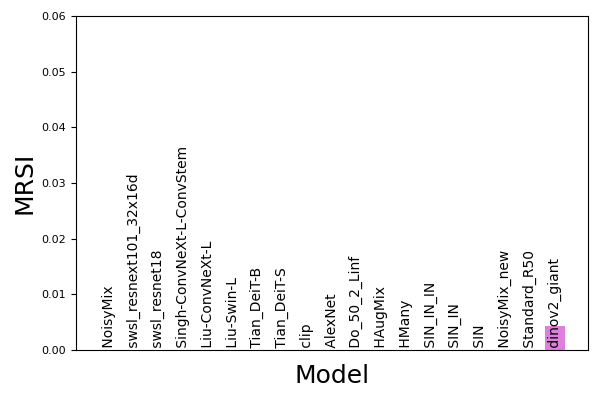}
        \caption{\scriptsize \textit{MRSI} for $\robustnesssymbol_a$}
        \label{fig:frost-MRSI-acc}
    \end{subfigure} &
    \begin{subfigure}{0.32\textwidth}
        \includegraphics[width=\textwidth]{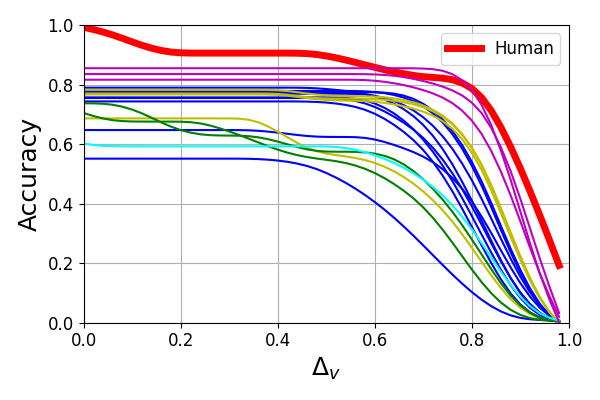}
        \caption{Estimated curves $s_a$}
        \label{fig:frost-vcr-acc-comparison}    
    \end{subfigure}
    \\
   \begin{subfigure}{0.32\textwidth}
        \includegraphics[width=\textwidth]{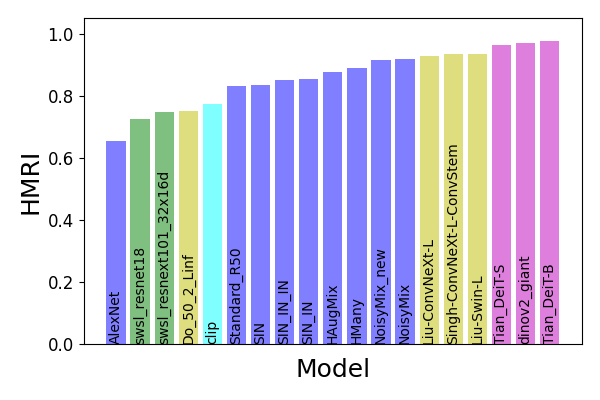}
        \caption{\scriptsize \textit{HMRI} for $\robustnesssymbol_p$}
        \label{fig:frost-HMRI-pred}
    \end{subfigure} &
    \begin{subfigure}{0.32\textwidth}
        \includegraphics[width=\textwidth]{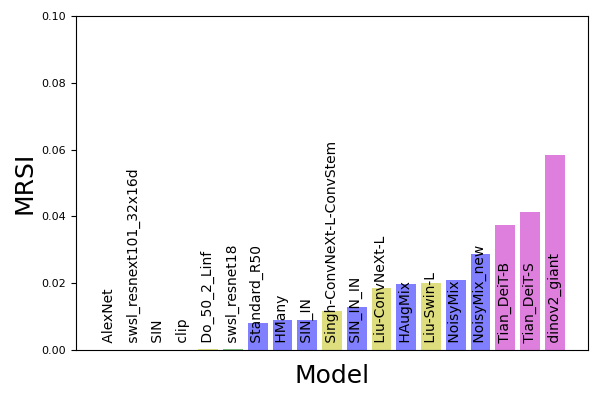}
        \caption{\scriptsize \textit{MRSI} for $\robustnesssymbol_p$}
        \label{fig:frost-MRSI-pred}
    \end{subfigure}&
\begin{subfigure}{0.32\textwidth}
        \includegraphics[width=\textwidth]{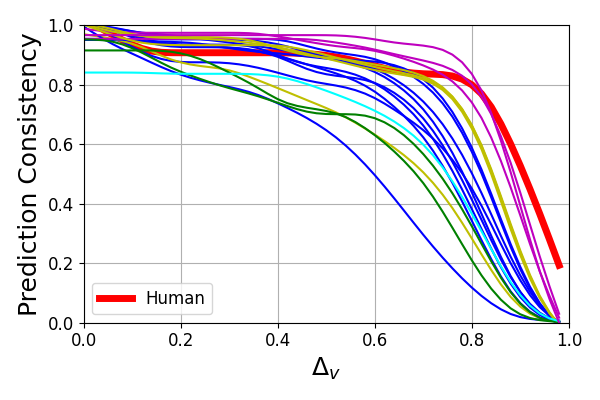}
        \caption{Estimated curves $s_p$}
        \label{fig:frost-vcr-pred-sim-comparison}    
    \end{subfigure}
    
    \\
    \end{tabular}
    \caption{VCR evaluation results for Frost.}
    \label{fig:hmri-mrsi-bar-plot_14}
\end{figure}

\end{document}